\documentclass[10pt,letterpaper,fignum,man,floatsintext,apacite]{apa7}
\usepackage{amsmath}
\usepackage{graphicx}
\usepackage{array}
\usepackage{textcomp}
\usepackage{setspace}
\usepackage{url}
\usepackage{cleveref}
\usepackage{caption}
\usepackage{csquotes}
\usepackage{longtable}
\usepackage{alltt}
\usepackage{tabularx}
\usepackage[sort]{natbib}

\newcommand{\cat}[1]{\textsc{#1}}
\newcommand{\oneargument}[2]{{\cat{#1}} $\rightarrow$ {\cat{#2}}}

\newcommand{\twoargument}[3]{\{{\cat{#1}}, {\cat{#2}}\} $\rightarrow$ {\cat{#3}}}

\newcommand{\threeargument}[4]{\{{\cat{#1}}, {\cat{#2}}, {\cat{#3}}\} $\rightarrow$ {\cat{#4}}}
\newcommand{\phen}[1]{#1}

\definecolor{human}{RGB}{44, 34, 254}
\definecolor{gpt4}{RGB}{255, 0, 0}
\definecolor{gpt4alt}{RGB}{255, 191, 202}
\definecolor{gpt353}{RGB}{2, 127, 0}
\definecolor{gpt353alt}{RGB}{148, 209, 165}
\definecolor{gpt35chat}{RGB}{255,165,0}

\newcommand{\gptthreefivethree}[1]{\textcolor{gpt353}{#1}}
\newcommand{\gptthreefivechat}[1]{\textcolor{gpt35chat}{#1}}
\newcommand{\gptfour}[1]{\textcolor{gpt4}{#1}}
\newcommand{\human}[1]{\textcolor{human}{#1}}
\newcommand{\gptthreefivethreescm}[1]{\textcolor{gpt353alt}{#1}}
\newcommand{\gptfourscm}[1]{\textcolor{gpt4alt}{#1}}

\title{Inductive reasoning in humans and large language models}
\shorttitle{Inductive reasoning in humans and LLMs}

\authorsnames{Simon J. Han, Keith Ransom, Andrew Perfors, Charles Kemp}
\authorsaffiliations{{Melbourne School of Psychological Sciences, University of Melbourne}}

\abstract{

The impressive recent performance of large language models has led many to wonder to what extent they can serve as models of general intelligence or are similar to human cognition. We address this issue by applying GPT-3.5 and GPT-4 to a classic problem in human inductive reasoning known as property induction. Over two experiments, we elicit human judgments on a range of property induction tasks spanning multiple domains. Although GPT-3.5 struggles to capture many aspects of human behaviour, GPT-4 is much more successful: for the most part, its performance qualitatively matches that of humans, and the only notable exception is its failure to capture the phenomenon of premise non-monotonicity. Our work demonstrates that property induction allows for interesting comparisons between human and machine intelligence and provides two large datasets that can serve as benchmarks for future work in this vein. 

}

\keywords{reasoning, property induction, category-based induction, non-monotonicity, neural networks, GPT-3.5, GPT-4, AI, large language models, representation}

\authornote{\addORCIDlink{Simon Jerome Han}{0000-0003-1157-7748}\\ 
\addORCIDlink{Keith Ransom}{0000-0001-5423-6455}\\
\addORCIDlink{Andrew Perfors}{0000-0002-6976-0732}\\
\addORCIDlink{Charles Kemp}{0000-0001-9683-8737}\\

Code and data are available at \protect\url{https://github.com/S-J-HAN/InductiveReasoningInLargeLanguageModels}

This work was supported in part by the Complex Human Data Hub at the University of Melbourne and by ARC FT190100200.

Correspondence concerning this article should be addressed to Jerome Han.  E-mail: \texttt{simon.jerome.han@gmail.com}

}

\begin{document}
\maketitle

In late 2022 and early 2023, large language models (LLMs) exploded into the public arena and captured the imagination of academic researchers and the general public alike. 
Systems such as ChatGPT and GPT-4 are so adept at engaging in natural conversations on a broad range of topics that even sober teams of researchers have concluded that these models show ``sparks of artificial general intelligence''  \citep{bubeck23}. As a result, there is currently intense interest in the scope and limitations of these models and the ways in which they may transform society.

LLMs had been extensively studied even before their recent surge in popularity, and for several years there has been an active research area that aims to carefully evaluate how their abilities compare with those of humans.  Many families of tasks are used in this literature \citep{changb23}, including some that specifically target linguistic abilities \citep{hu20} and others that target commonsense knowledge and logical reasoning \citep{Rae2021gopher}. Here we propose that the set of existing tasks can be usefully supplemented by drawing on the extensive psychological literature on inductive reasoning.  To support this general claim we explore the extent to which two generations of the GPT model (GPT-3.5 and GPT-4) are able to account for core phenomena in human property induction (also known as category-based or categorical induction).

Inductive reasoning is a fundamental cognitive challenge that requires arriving at plausible conclusions in the face of uncertainty \citep{hollandhnt86,slomanl05}. An inference is deductive if the conclusion follows with certainty given the available evidence, but inductive if the conclusion is plausible but not guaranteed. Because most everyday reasoning problems involve sparse, noisy, or uncertain data, most of these problems require inductive rather than deductive reasoning \citep{chaterohh11}. Within the AI literature, work on inductive reasoning falls under many different headings including non-monotonic reasoning~\citep{brewka1997}, commonsense reasoning~\citep{davis2015} and natural language inference~\citep{storks2019}. Psychologists have also studied many varieties of induction, including generalization \citep{shepard87}, categorization \citep{pothosw11}, and analogical reasoning \citep{vosniadouo89}. Here we focus on property induction because this task is easily formulated using simple linguistic stimuli, and because the literature on this task is relatively rich. 

\begin{figure}[t!]
\begin{center}
\includegraphics[width=0.9\textwidth]{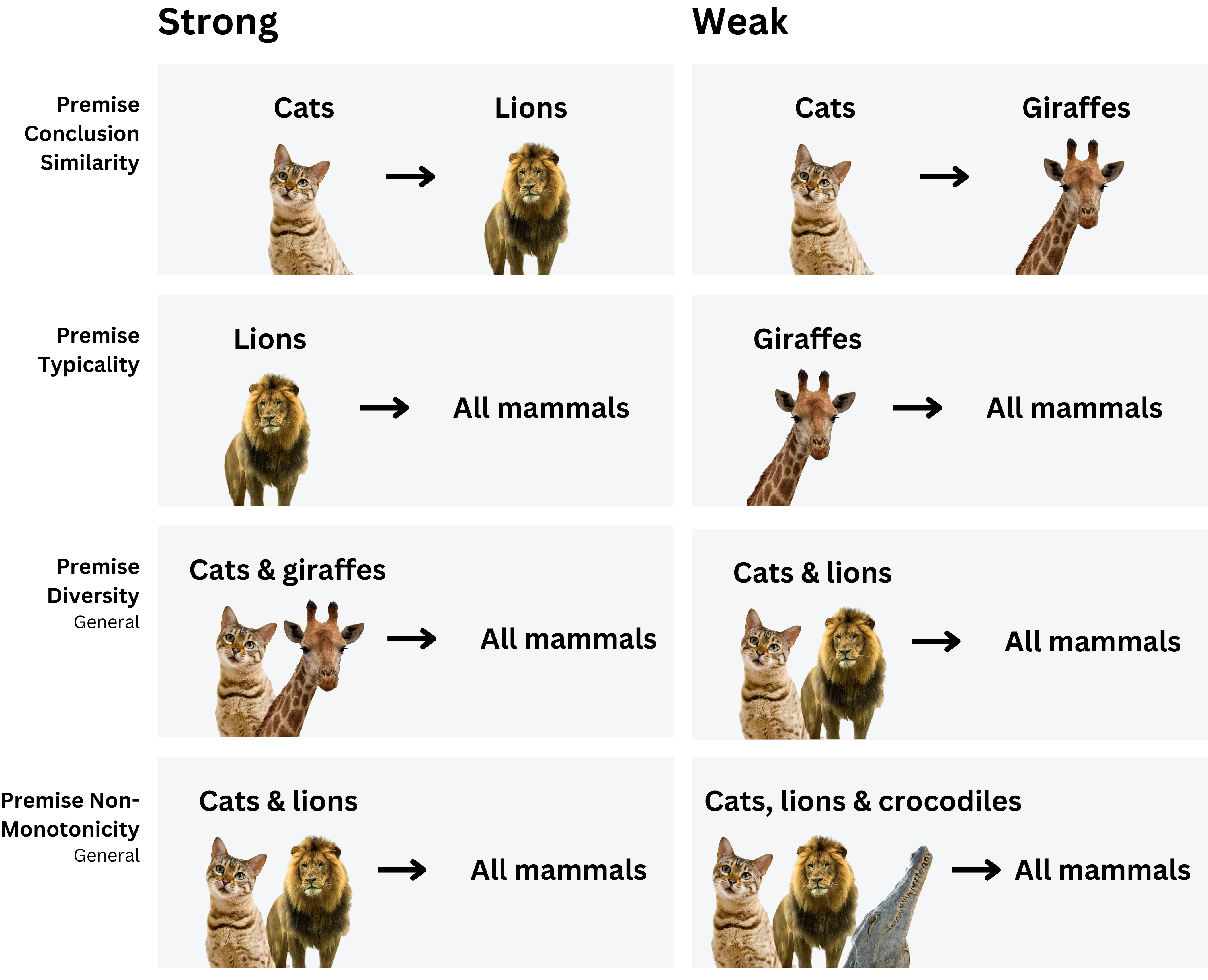}
\end{center}
\vspace{-1.5mm}
\caption{\textbf{Schematic illustration of selected property induction phenomena.} The column on the left depicts arguments that people perceive as stronger than the corresponding column on the right. For instance, people who are told that \textsc{Cats} have some property are more willing to conclude that similar animals like \textsc{Lions} have the property than they are to conclude that dissimilar ones like \textsc{Giraffes} do. This phenomenon, shown in the top row, is known as Premise-Conclusion Similarity. The figure depicts only four of the eleven phenomena we investigate in this paper.
}
\label{fig:demo}
\vspace{-4.5mm}
\end{figure}

In a property induction task, people are given premises that indicate that a property 
is shared by one or more categories and must assess whether the property is shared by a different category  \citep{rips75,slomanl05}. For example, the top left argument in Figure~\ref{fig:demo} might be presented by informing participants that cats have sesamoid bones, then asking them whether lions also have sesamoid bones. Figure~\ref{fig:demo} compares four pairs of arguments, and the distinctions between strong and weak arguments in the top two rows are relatively obvious. In the first row, cats are more similar to lions than to giraffes, and in the second row, lions are more typical mammals than are giraffes. The final two rows show how property induction tasks can elicit more sophisticated kinds of reasoning. In the third row, the argument on the left is stronger even though giraffes are less typical mammals than lions (see row 2). This difference in typicality between lions and giraffes is dominated by the fact that cats and giraffes form a more diverse set of premises than cats and lions. The final row suggests that adding a premise to an argument can sometimes make an argument weaker. Under many circumstances, observing an additional species with sesamoid bones should provide increased confidence that all mammals have sesamoid bones, so examples of Non-Monotonicity are especially interesting.

Property induction is an appealingly simple task that has been used to study the reasoning of children \citep{carey85} and adults from a broad range of cultural backgrounds \citep{lopezacms97}. Despite this apparent simplicity, the task yields a rich range of phenomena that draw on many kinds of knowledge (see \citet{Hayes2018} for a review). This knowledge includes not just similarity \citep{oshersonswls90}, but also causal relationships \citep{medincsh05} and assumptions about the process by which the premises were generated \citep{ransompn16}.
This range of inductive phenomena --- from simple similarity-based effects to theory-based effects that draw on richer kinds of knowledge --- corresponds to a sequence of increasingly difficult challenges for LLMs and other computational models \citep{sloman93,rogersm04,kempt09}. As such, property induction tasks could potentially lead to benchmarks that help to drive continued progress in computer science and AI. Some of the benchmarks currently used to evaluate LLMs focus on inductive problems and are directly inspired by psychological research \citep{siqa,mewl}, but to our knowledge none of these benchmarks considers the task of property induction. Here we introduce two property induction data sets that are relatively large by the standards of psychological research, and thus represent an initial step towards a comprehensive property induction benchmark.

For psychologists, property induction is an important tool for assessing LLMs and predecessors such as LSA \citep{landauerd97} as computational accounts of the acquisition, use, and representation of semantic knowledge. Recent work has evaluated the extent to which LLMs account for human similarity ratings, typicality ratings, and response times \citep{bhatiar21,lakem21} in semantic verification, and several groups have evaluated LLMs on inductive phenomena inspired by the psychological literature, including analogical reasoning~\citep{webbhl22}, pragmatic reasoning~\citep{lipkin2023evaluating}, causal reasoning~\citep{kiciman2023causal} and social reasoning~\citep{shapira2023clever,ullman2023}. Closest to the current paper is a study by  \citet{misraer21}, who focus on typicality and include property induction as one of the tasks that they consider. Typicality is among the phenomena considered here, but we investigate many others as well.  

The next section introduces the inductive phenomena that we analyze, along with a theoretical account of these phenomena known as the similarity-coverage Model \citep{oshersonswls90}. We then present two new datasets that we collected to characterize inductive reasoning in humans, and use them to evaluate the performance of both GPT-3.5 and GPT-4. To preview our results, we find that GPT-4 accounts well for all of the phenomena that we investigate except Non-Monotonicity. 

Because the literature on LLMs is moving so fast, it seems useful to document the period during which this work was carried out. At the time we developed the study,  GPT-3 was the most advanced version available, and we chose to focus on a set of phenomena that included some that were within the reach of GPT-3 but others that seemed more challenging. GPT-4 became available shortly before we submitted this work for publication, and we found that it performed relatively well when we included it in our evaluation. Had GPT-4 been available when we designed the study, we would have aimed to include additional inductive phenomena that seemed clearly out of reach for GPT-3 but that may be feasible for GPT-4. The closing sections of the paper discuss some of these phenomena and suggest additional directions that future work on LLMs and inductive reasoning can pursue.

\begin{table}[t]
\begin{center} 
\caption{\textbf{Eleven property induction phenomena} introduced by \citet{oshersonswls90} and investigated in this paper. The second column is based on the levels occupied by premises and conclusion in a category hierarchy. For specific arguments, premises and conclusion lie at the same level, but for general arguments the conclusion lies at a higher level than the premises. Premises are indicated in the brackets and the conclusion is on the right of the arrow.} 
\label{tbl:phenomena}
\vspace{-1mm}
\resizebox{\textwidth}{!}{
\begin{tabular}{llll} 
\hline 
\textbf{Phenomenon} & \textbf{Type} &  \textbf{Stronger argument} & \textbf{Weaker argument} \\
\hline

Similarity & Specific & 
\twoargument{robin}{bluejay}{sparrow} & 
\twoargument{robin}{bluejay}{goose} \\ 

Typicality & General & 
\oneargument{robin}{bird} & 
\oneargument{penguin}{bird} \\

Specificity & General & 
\twoargument{bluejay}{falcon}{bird} & 
\twoargument{bluejay}{falcon}{animal} \\

Monotonicity & General & 
\threeargument{sparrow}{eagle}{hawk}{bird} & 
\twoargument{sparrow}{eagle}{bird} \\

Monotonicity & Specific & 
\threeargument{pig}{wolf}{fox}{gorilla} & 
\twoargument{pig}{wolf}{gorilla} \\

Diversity & General & 
\twoargument{hippo}{hamster}{mammal} &
\twoargument{hippo}{rhino}{mammal} \\

Diversity & Specific & 
\twoargument{lion}{giraffe}{rabbit} &
\twoargument{lion}{tiger}{rabbit} \\

Non-Monotonicity & General &
\twoargument{crow}{peacock}{bird} &
\threeargument{crow}{peacock}{rabbit}{bird} \\

Non-Monotonicity & Specific &
\oneargument{fly}{bee} & 
\twoargument{fly}{orangutan}{bee} \\

Asymmetry & Specific &
\oneargument{mouse}{bat} & 
\oneargument{bat}{mouse} \\

Inclusion Fallacy & Both & 
\oneargument{robin}{bird} & 
\oneargument{robin}{ostrich} \\
\hline
\end{tabular}}
\end{center}
\vspace{-4.5mm}
\end{table}

\subsection{Inductive Phenomena}
\label{sec:inductivephenomena}

We follow a long tradition of studies that examine inductive reasoning by focusing on property induction with semantically ``blank" or unfamiliar properties. In a typical property induction task, participants are asked to rate the strength of inductive arguments like ``\cat{robins} have property P, therefore \cat{birds} have property P.'' We will use the notation \oneargument{robin}{bird} to indicate an argument that involves generalizing a property from a premise (e.g., \cat{robin}) to a conclusion (e.g., \cat{birds}). Arguments can also have multiple premises, indicated by putting them in brackets on the left.

Although this task may seem simple, it gives rise to numerous phenomena that are indicative of the complex ways that humans reason inductively. \citet{oshersonswls90} presented thirteen such phenomena, eleven of which are shown in Table~\ref{tbl:phenomena}. All eleven involve comparing a stronger argument with a weaker argument; the two phenomena not included in the table or analyzed here are omitted because they are not formulated in terms of a similar comparison.

Some of the phenomena directly capture effects of similarity or typicality. For instance, Premise-Conclusion {\bf Similarity} reflects the finding that people are more likely to generalize a property from one concept to another when the concepts are more similar. Premise {\bf Typicality} is the finding that arguments are stronger if the premises are more typical of the conclusions. A slightly less reliable phenomenon, Premise-Conclusion {\bf Asymmetry}, reflects the fact that an argument that generalizes from a typical category member to a less typical one (e.g. \oneargument{mouse}{bat}) is often rated as stronger than the reverse argument (e.g. \oneargument{bat}{mouse}); this is probably because atypical categories are more likely to have atypical properties.

Other phenomena relate to the hierarchical organization of categories. Conclusion {\bf Specificity} reflects the intuition that greater inductive leaps are required to support broader generalizations; arguments are thus stronger if the conclusion category is more specific. The {\bf Inclusion Fallacy} relates to the observation that a general argument that projects from a category to its enclosing class (e.g. \oneargument{robin}{bird}) is often considered stronger than a more specific argument (e.g. \oneargument{robin}{ostrich}) even though the latter is logically entailed by the former. We evaluate the inclusion fallacy for completeness, but because it is normally viewed as a fallacy it may not necessarily be appropriate as a target for AI models like GPT-3.5 and GPT-4.

In addition to these relatively straightforward phenomena, there are also those which appear to reflect more sophisticated or theory-based reasoning about underlying mechanisms. Premise {\bf Diversity} refers to the fact that arguments are often considered stronger if their premises are less similar to one another. This captures the general intuition, based on an understanding of statistical sampling, that diverse evidence is more compelling than narrow evidence. A similar mechanism may underlie systematic violations of Premise \textbf{Monotonicity}, in which additional positive premises increase the strength of an argument. Monotonicity often holds if all premises are drawn from the same superordinate category, but adding premises from a different superordinate category can lead to the opposite pattern of reasoning, known as
 Premise {\bf Non-Monotonicity}. For example, the inclusion of \cat{orangutan} in the argument \twoargument{fly}{orangutan}{bee} means that the context of the argument (the smallest category which includes the premise and inclusion categories) changes from \cat{insect} to \cat{animal}. This suggests that the property in question is not insect-specific, and thus reduces the chance that bees share it. These violations of premise monotonicity have been shown to be influenced by the reasoner's theoretical assumptions about how the premises were generated \citep{ransompn16,voorspoelsetal15,Hayes19}.

\subsection{Similarity-Coverage Model (SCM)}

In addition to characterizing the inductive phenomena just described, \citep{oshersonswls90} presented a theory known as the similarity-coverage model (SCM) that is able to account for all of them. It will be used as part of our evaluation of GPT-3.5 and GPT-4.

The SCM builds on the fact that several inductive phenomena can be derived purely from similarity between categories. For example, \oneargument{robin}{sparrow} is stronger than  \oneargument{robin}{goose} because robins are more similar to sparrows than geese. Similarly, \oneargument{robin}{bird} is stronger than \oneargument{penguin}{bird} because robins are more similar to the prototypical bird than penguins are. In both cases, the probability that the premise and conclusion categories share a property increases solely based on the similarity of the two sets of categories.

Although similarity-based accounts of property induction are simple and intuitive, they fail to account for more complex phenomena such as non-monotonicity and diversity. The SCM captures them by incorporating a notion called \emph{coverage}, which denotes the degree to which the premise categories are similar to members of the lowest level category class that encapsulates each of the premise and conclusion categories. \citet{oshersonswls90} demonstrate that a weighted combination of coverage and premise-conclusion similarity captures all eleven of the phenomena in Table~\ref{tbl:phenomena}.

\subsection{Overview}

This paper is divided into two experiments, each focused on comparing human inductive judgments with equivalent judgments derived from GPT-3.5 and GPT-4. The design of each experiment was inspired by the experiments of \citet{oshersonswls90}. In Experiment 1, the task was to pick the stronger argument out of a pair of property induction arguments, and in Experiment 2 it was to rate the strength of an argument in isolation. 

Both experiments included stimuli from three category domains: \textsf{Mammals}, \textsf{Birds}, and \textsf{Vehicles}, sourced from the Leuven Natural Concepts Dataset (LNCD) reported in \citet{dedeyne2008}. Psychological studies of induction often use arguments about mammals, and examples involving birds are common in the AI literature on defeasible reasoning. Comparing performance across three separate domains ensures that the LLMs are required to reason about domains with different levels of prominence in their training data.
We selected 24 categories from each of the three domains by first excluding those that were near-duplicates or unfamiliar, and then randomly removing some until the category sets for each domain were of the same size. The full list of categories in each domain is shown in Table~\ref{tbl:allcategories}.

\section{Experiment 1: Argument pairs}

In this experiment, agents were presented with a series of property induction argument pairs and were then asked to pick the stronger argument of each pair.

\subsection{Generating argument pairs}

With three domains, 24 categories in each domain and the 11 phenomena in Table~\ref{tbl:phenomena}, evaluating all possible combinations of arguments is extremely prohibitive. We therefore generated a subset of argument pairs for each phenomenon and domain, ensuring as far as possible that they were sufficient to capture the range of variation. As a first step, for each phenomenon in each domain we sampled thousands of candidate argument pairs by replacing each slot in the argument template with a randomly sampled category. We then selected the candidate pairs that were most appropriate for each of the phenomena in question. For instance, Similarity, Typicality, Asymmetry, and the Inclusion Fallacy are all stronger if one of the arguments contains a highly typical or similar premise category and the other contains a highly atypical or dissimilar premise category. We thus restricted the candidate pool for those phenomena to the categories (and category pairs) that had typicality (and similarity) ratings of 0.75SD above or below their domain's mean typicality (or similarity) rating as given by the LNCD norms \citep{dedeyne2008}. To control for similarity and typicality effects in Diversity and Monotonicity argument pairs, we ensured that the unique premise category in each pair was no more typical or similar in the stronger argument than it was in the weaker argument. To construct premise categories for Non-Monotonicity, we sampled from three supplementary domains: reptiles for \textsf{Mammals}, insects for \textsf{Birds} and tools for \textsf{Vehicles}.

 \begin{table}[t!]
    \centering
        \caption{\small{The 24 categories used in each domain. The same categories were used in both experiments except that in Experiment 2, four  \textsf{Mammals} (\textit{beaver, giraffe, lion,} and \textit{rabbit}) were replaced with \textit{monkey, hippo, fox} and \textit{wolf} in order to facilitate comparison to \citet{oshersonswls90}, which used the latter.}}
        \vspace{2mm}        
    \begin{tabular}{|p{0.1\linewidth} | p{0.8\linewidth}|}
        \hline
        \textsf{Mammals} & bat, beaver, camel, cat, cow, deer, dog, donkey, elephant, giraffe, hamster, hedgehog, horse, kangaroo, lion, llama, mouse, pig, rabbit, rhino, sheep, squirrel, tiger, zebra \\
        \hline
        \textsf{Birds} & blackbird, canary, chicken, crow, dove, duck, eagle, falcon, heron, magpie, ostrich, owl, parrot, peacock, penguin, robin, rooster, seagull, sparrow, stork, swan, swallow, turkey, vulture\\
        \hline
        \textsf{Vehicles} & airplane, bicycle, boat, bus, car, caravan, carriage, cart, helicopter, hovercraft, jeep, moped, motorbike, rocket, skateboard, sled, submarine, taxi, tractor, tram, train, truck, van, zeppelin\\
        \hline
    \end{tabular}
    \vspace{-2mm}
    \label{tbl:allcategories}
\end{table}

Once we had candidate pools, our goal was to select the 24 pairs in each phenomenon-domain split that were most likely to capture that phenomenon. We achieved this by using an SCM model based on LNCD similarity ratings to calculate, for each argument pair, a measure of the disparity between the strength of the stronger and weaker arguments in that pair. We then selected the 24 pairs with the highest SCM disparity.
\footnote{Because the LNCD only includes ratings between pairs of categories within the same domain, we could not obtain SCM scores for  Specificity and Non-Monotonicity argument pairs. However, it was straightforward to derive the stronger and weaker arguments for each of our constructed pairs: for Specificity, the stronger argument is always the argument with the more specific conclusion, and for Non-Monotonicity the stronger argument is always the one with a more specific category superset. For these phenomena, we simply took a random sample of 24 pairs in order to generate the final split.} This resulted in 792 unique argument pairs in total across the whole experiment (11 phenomena $\times$ 3 domains $\times$ 24 argument pairs).

\subsection{Presenting argument pairs to humans}

\subsubsection{Participants}

We recruited 120 people via Amazon Mechanical Turk who were each paid \$1.50USD for the 5-8 minute study. All passed a screening for English language competency prior to participation and indicated informed consent via an online consent form. Both experiments were approved by the Human Research Ethics Subcommittee of the University of Adelaide. 

\subsubsection{Stimuli}

Each participant was shown a different set of 66 argument pairs that were randomly sampled from the set of 792 described above. We ensured that everyone saw stimuli from all three domains and all 11 phenomena (presented in a different random order for each person). In addition to the 66 experimental trials, there were four attention check trials, resulting in 70 trials in total. The attention check trials, which occurred after every 16 trials, were the same for all participants. They were designed to look like a standard argument pair but had a relatively unambiguous answer (e.g., \oneargument{Robins}{Gorillas} vs \oneargument{Robins}{Sparrows}); we reasoned that responding incorrectly to them would indicate inattentiveness or a failure to understand the task. Participants who did not answer at least three of the four check questions correctly were excluded ($N=10$), resulting in $110$ people in the full dataset. There were a mean number of $9.17$ ratings obtained for each of the 33 phenomenon-domain splits (min: 8; max: 10).

\begin{figure}[t!]
\begin{center}
\includegraphics[width=\textwidth]{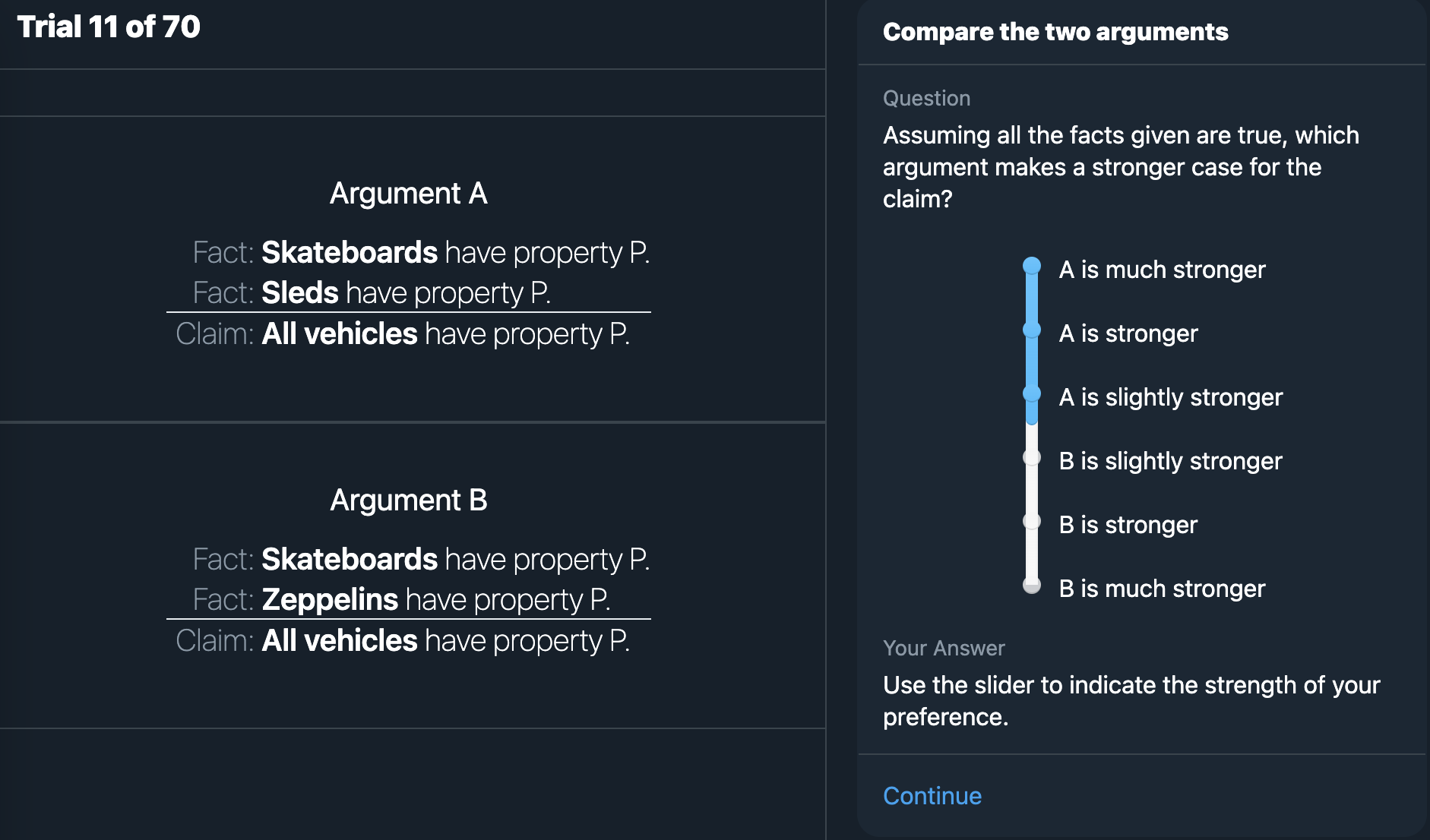}
\end{center}
\vspace{-1.5mm}
\caption{%
\textbf{Experiment 1 screenshot.} On each trial, participants were shown two arguments (A and B) on the left, and then asked to rate which was stronger using the 6-point slider on the right (the slider was always initialized to the midpoint but people were required to move it to one of the six options before they could continue to the next trial). All participants saw a different random subset of the 792 possible argument pairs, constrained so that each person saw examples from all three domains and all 11 phenomena. This screenshot contains an argument pair from the domain of \textsf{Vehicles} and the phenomenon of Premise Diversity (which predicts that Argument B is stronger). }
\label{fig:expt1screenshot}
\vspace{-3mm}
\end{figure}

\subsubsection{Procedure}

At the start of every session, participants were shown the following instructions:
\vspace{-2mm}
\begin{displayquote}
{\small
\texttt{We are interested in how people evaluate arguments. On each trial there will be two arguments labeled `A' and `B.' Each will contain one, two, or three statements separated from a claim by a line. Assume that the statements above the line are facts, and choose the argument whose facts provide a better reason for believing the claim. These are subjective judgments; there are no right or wrong answers.}}
\end{displayquote}

On each of the 70 trials, people were shown two argument pairs labeled \texttt{Argument A} and \texttt{Argument B}; whether the stronger or weaker argument of each pair was labeled as \texttt{Argument A} or \texttt{Argument B} was randomly assigned. Each argument pair was presented on its own page, and on the right of the pair people were asked ``Assuming all the facts given are true, which argument makes a stronger case for the claim?'' They then selected their response using a 6-point Likert scale ranging from 1 (Argument A is much stronger) to 6 (Argument B is much stronger). A screenshot from an example trial is shown in Figure~\ref{fig:expt1screenshot}.

\begin{table}[t]
    \centering
        \caption{\small{Best-performing GPT-4 prompt for Experiment 1. It was identified by evaluating multiple prompts varying the  components of System, Context, Argument, Question, and Options. The column on the right indicates which variant of that component it corresponds to. For System message S3, \texttt{X} is replaced by \texttt{living things} for the \textsf{Mammals} and \textsf{Birds} domains, and \texttt{objects} for the \textsf{Vehicles} domain.}}
        \vspace{1mm}
    \begin{tabular}{|p{0.03\linewidth} | p{0.93\linewidth}|}
    \hline
    S3 & \small{\texttt{You are an expert on X and the types of real world properties that they have. The questions you'll see don't have right or wrong answers, and you are willing to use your best judgment and commit to a concrete, specific response even in cases where you can't be sure that you are correct.
    }}\\
\vspace{1mm} C1 &  \vspace{1mm} \small{\texttt{We are interested in how people evaluate arguments. On each trial there will be two arguments labeled `A' and `B.' Each will contain one, two, or three statements separated from a claim by a line. Assume that the statements above the line are facts, and choose the argument whose facts provide a better reason for believing the claim. These are subjective judgments; there are no right or wrong answers.}} \\
\vspace{1mm} A1 & \vspace{1mm} \small{\texttt{Argument A: Fact - Dogs have property P.}}\\
& \hspace{21mm} \small{\texttt{Claim - All mammals have property P.}}\\
& \small{\texttt{Argument B: Fact - Hedgehogs have property P.}}\\
& \hspace{21mm} \small{\texttt{Claim - All mammals have property P.}}\\
\vspace{1mm} Q3 & \vspace{1mm} \small{\texttt{Question: Assuming all the facts given are true, which argument makes a stronger case for the claim? To get the best answer, first write down your reasoning. Then, based on this,}}\\
\vspace{1mm} O1 & \vspace{1mm} \small{\texttt{Indicate the strength of your preference by providing one of the following options:}}\\
& \small{\texttt{A - Argument A is much stronger}}\\
& \small{\texttt{B - Argument A is stronger}}\\
& \small{\texttt{C - Argument A is slightly stronger}}\\
& \small{\texttt{D - Argument B is slightly stronger}}\\
& \small{\texttt{E - Argument B is stronger}}\\
& \small{\texttt{F - Argument B is much stronger}}\\
\hline
    \end{tabular}
    \label{tbl:expt1bestprompt}
\end{table}

\subsection{Presenting argument pairs to GPT-4}

To obtain analogous judgments from GPT-4, we used the \texttt{gpt-4-0314} engine within the Chat API of OpenAI. We set temperature $t = 0$ for all API requests alongside a maximum response length of 400 tokens.  While higher temperature values in theory might have allowed us to obtain a response distribution from GPT-4 for every argument pair and prompt design, in practice we found that final response ratings did not vary much even for $t = 1$.  We therefore presented the 792 argument pairs to GPT-4 only once for each prompt design.

We designed the prompts for GPT-4 with two aims in mind. First, we wanted them to be as similar as possible to the instructions our human participants saw. Second, given the degree to which the performance of GPT-4 varies based on the prompt, we wanted to make sure that our results could not obviously be improved by better prompts. To accomplish this, we designed a logical space of prompts and selected a subset of the most promising to evaluate. The logic of our design was informed by the observation that a prompt could be constructed by combining task agnostic instructions (the \textbf{System} message) with four consecutive components: the explanation of the task (\textbf{Context}), the structure of the arguments being rated (\textbf{Arguments}), the judgment elicitation (\textbf{Question}), and the specification of the answer format (\textbf{Options}). There were three variants of the System message (S1-S3), four of Context (C1-C4), three of Arguments (A1-A3), four of Question (Q1-Q4), and two of Options (O1-O2); all are shown in \ref{sec:exp1prompts}.

Because running all 288 possible prompts on all 792 arguments was prohibitively expensive, we identified a subset of prompts that we thought would maximize the ability of GPT-4 to respond well. This included a baseline prompt designed to be as similar as possible to the human experiment (S1-C1-A1-Q1-O1), one that is similar to the baseline except with a more specific system message (S3-C1-A1-Q1-O1), one that also adds a chain-of-thought direction (S3-C1-A1-Q3-O1), and one that elicited domain-specific reasoning in particular (S3-C1-A1-Q4-O1). For each of the 33 phenomena-domain splits, we then evaluated the performance of each prompt using the sign test described in Section \ref{sec:exp1result}. The best-performing prompt was S3-C1-A1-Q3-O1, shown in Table~\ref{tbl:expt1bestprompt}, and all of thethe results reported in the main text are based on this prompt. 

\subsection{Presenting argument pairs to GPT-3.5}

In addition to GPT-4, we also presented our set of argument pairs to GPT-3.5. This was achieved using the \texttt{text-davinci-003} engine. Although the GPT-3.5 series of models includes multiple variants, we chose \texttt{text-davinci-003} because it is the best performing model that uses the OpenAI Completions API. Where possible, results for other GPT variants are reported in  \ref{sec:otheropenaimodels}. Like our experiments with GPT-4, we set $t = 0$ for all API requests alongside a maximum response length of 400 tokens. Because we use the Completions API and not the Chat API for GPT-3.5, we adapted our baseline prompt for GPT-4 to GPT-3.5 by including both the \texttt{system message} and the \texttt{user message} in a single completions prompt.

Because the GPT-3.5 Completions API includes token probabilities in its responses, we assess GPT-3.5 differently from GPT-4. Instead of taking the single most likely completion generated by GPT-3.5, we generate a score for each argument pair using the sum of probabilities for each point on the provided Likert scale, weighted by rank, at the token position at which the provided answer occurs. Thus, our results for GPT-3.5 are continuous rather than discrete, and our ability to access token probabilities for GPT-3.5 affords it a methodological advantage in precision over GPT-4.

\subsection{Results}
\label{sec:exp1result}

\begin{figure}[t!]
\begin{center}
\includegraphics[width=\textwidth]{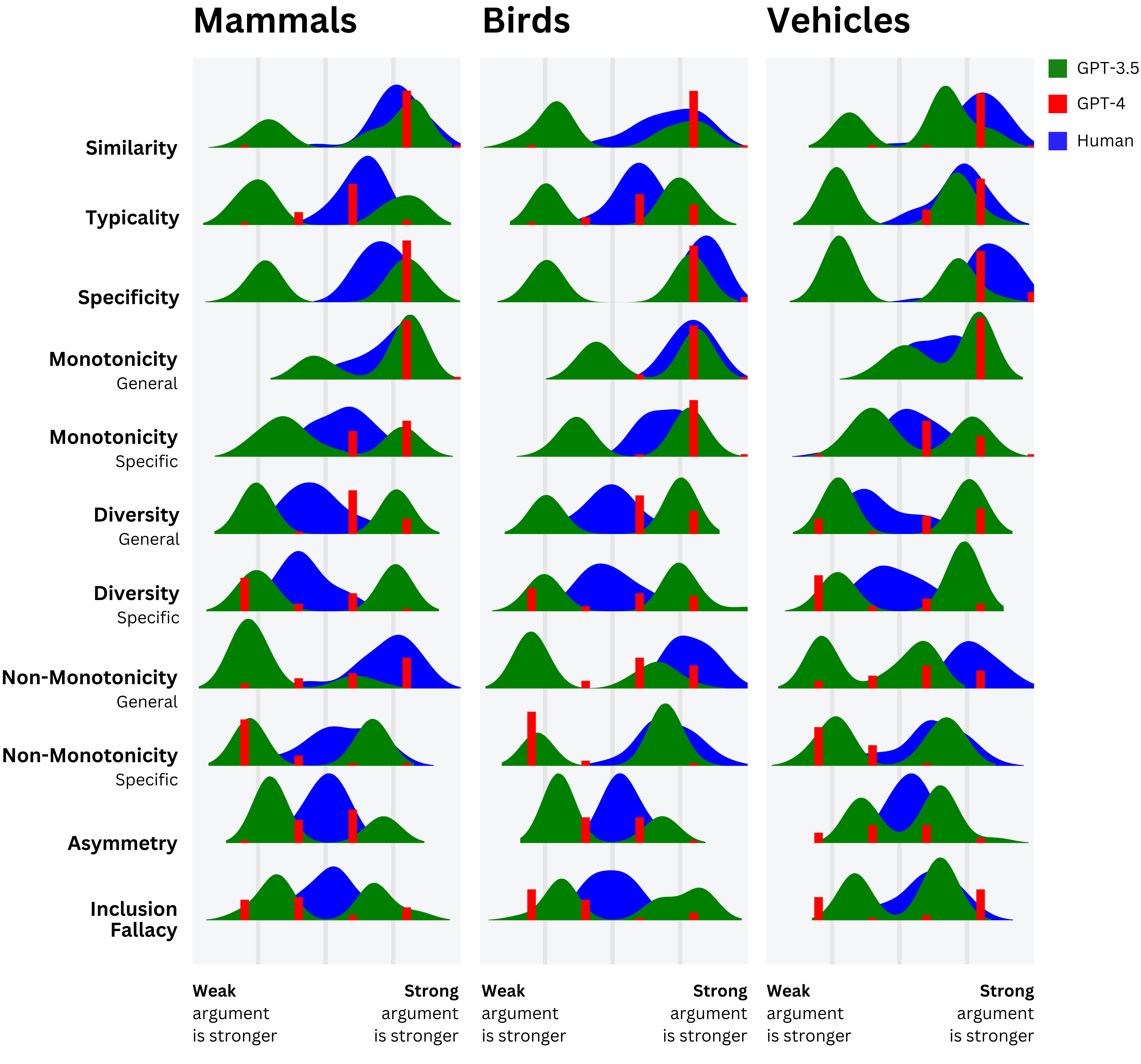}
\end{center}
\vspace{-1.5mm}
\caption{%
\small{\textbf{Inductive reasoning phenomena exhibited by GPT-3.5, GPT-4 and human reasoners.} Responses reflect which arguments were rated stronger (by convention this is the argument on the right) for each of the 3 domains (columns) and 11 phenomena (rows). Distributions are continuous for GPT-3.5 and humans and discrete for GPT-4 because of API constraints (multiple bars occur when it gave different answers for specific stimuli within each domain-phenomenon cell). GPT-3.5 performs poorly in general. While GPT-4 captures phenomena involving Similarity, Specificity, Typicality and Monotonicity, it makes the opposite predictions as humans for Non-Monotonicity (Specific).
}}
\label{fig:e1-main-results}
\vspace{-3mm}
\end{figure}

Figure~\ref{fig:e1-main-results} compares the performance of humans, GPT-3.5, and GPT-4 across all three domains on the 11 inductive reasoning phenomena we investigated. It is immediately apparent that each kind of agent exhibits a qualitatively different pattern of response. Human ratings (dark blue) are almost always unimodal and usually located to one side of the rating scale but closer to the center than either end. Conversely, GPT-3.5 ratings (green) are much more often bimodal and extreme. This reflects the fact that GPT-3.5 preferred to select options A and F rather than any of the other four choices. Direct visual comparisons between GPT-4 (red) and humans (blue) are more difficult because of the discrete nature of the GPT-4 responses, but it is clear that the peaks of the red and blue distributions do not always coincide.

\begin{table}[t!]
\centering
\caption{\small{Quantitative evaluation of GPT-3.5, GPT-4 and Humans on the 11 phenomena across all three domains. The number represents the p-value on a sign test, with significant p-values indicating a preference for one argument over the other. Most of the time this preference is in the theoretically predicted direction (*), but when the weaker argument is endorsed significantly more ($\circ$) it is in the opposite. Results for GPT-3.5's chat variant are reported in Appendix Table~\ref{tbl:expt1stats_appendix}.}}
\vspace{1mm}
\begin{tabular}{|p{0.2\linewidth}| p{0.2\linewidth} | p{0.16\linewidth} p{0.16\linewidth} p{0.16\linewidth} |}
\hline
\textbf{Phenomenon} & \textbf{Domain} & \gptthreefivethree{\textbf{GPT-3.5}} & \gptfour{\textbf{GPT-4}} & \human{\textbf{Humans}}\\ \hline
Similarity & Mammals & 0.17 & \gptfour{\textbf{$<$0.001 *}} & \human{\textbf{$<$0.001 *}} \\
  & Birds & 0.69 & \gptfour{\textbf{$<$0.001 *}} & \human{\textbf{$<$0.001 *}} \\
  & Vehicles & 0.11 & \gptfour{\textbf{$<$0.001 *}} & \human{\textbf{$<$0.001 *}} \\
\hline
Typicality & Mammals & 0.54 & \gptfour{\textbf{$<$0.02 *}} & \human{\textbf{$<$0.001 *}} \\
  & Birds & 0.54 & \gptfour{\textbf{$<$0.001 *}} & \human{\textbf{$<$0.001 *}} \\
  & Vehicles & 1.0 & \gptfour{\textbf{$<$0.001 *}} & \human{\textbf{$<$0.001 *}} \\
\hline
Specificity & Mammals & 0.84 & \gptfour{\textbf{$<$0.001 *}} & \human{\textbf{$<$0.001 *}} \\
  & Birds & 0.84 & \gptfour{\textbf{$<$0.001 *}} & \human{\textbf{$<$0.001 *}} \\
  & Vehicles & 0.54 & \gptfour{\textbf{$<$0.001 *}} & \human{\textbf{$<$0.001 *}} \\
\hline
Monotonicity & Mammals & \gptthreefivethree{\textbf{$<$0.02 *}} & \gptfour{\textbf{$<$0.001 *}} & \human{\textbf{$<$0.001 *}} \\
(General) & Birds & 0.54 & \gptfour{\textbf{$<$0.001 *}} & \human{\textbf{$<$0.001 *}} \\
  & Vehicles & \gptthreefivethree{\textbf{$<$0.001 *}} & \gptfour{\textbf{$<$0.001 *}} & \human{\textbf{$<$0.001 *}} \\
\hline
Monotonicity & Mammals & 0.31 & \gptfour{\textbf{$<$0.001 *}} & 0.06 \\
(Specific) & Birds & 0.84 & \gptfour{\textbf{$<$0.001 *}} & \human{\textbf{$<$0.001 *}} \\
  & Vehicles & 0.84 & \gptfour{\textbf{$<$0.001 *}} & 0.29 \\
\hline
Diversity & Mammals & 0.84 & \gptfour{\textbf{$<$0.001 *}} & 0.06 \\
(General) & Birds & 0.54 & \gptfour{\textbf{$<$0.001 *}} & 1.0 \\
  & Vehicles & 1.0 & 0.06 & \human{\textbf{$<$0.03 $\circ$}} \\
\hline
Diversity & Mammals & 1.0 & 0.15 & \human{\textbf{$<$0.01 $\circ$}} \\
(Specific) & Birds & 0.54 & 0.84 & 0.68 \\
  & Vehicles & 0.31 & 0.15 & 0.68 \\
\hline
Nonmonotonicity & Mammals & \gptthreefivethree{\textbf{$<$0.001 $\circ$}} & \gptfour{\textbf{$<$0.02 *}} & \human{\textbf{$<$0.001 *}} \\
(General) & Birds & 0.31 & \gptfour{\textbf{$<$0.001 *}} & \human{\textbf{$<$0.001 *}} \\
  & Vehicles & 0.84 & 0.15 & \human{\textbf{$<$0.01 *}} \\
\hline
Nonmonotonicity & Mammals & 1.0 & \gptfour{\textbf{$<$0.001 $\circ$}} & 0.15 \\
(Specific) & Birds & 0.15 & \gptfour{\textbf{$<$0.001 $\circ$}} & \human{\textbf{$<$0.001 *}} \\
  & Vehicles & 1.0 & \gptfour{\textbf{$<$0.001 $\circ$}} & \human{\textbf{$<$0.01 *}} \\
\hline
Asymmetry & Mammals & 0.06 & 0.68 & 0.4 \\
  & Birds & 0.06 & 1.0 & \human{\textbf{$<$0.001 *}} \\
  & Vehicles & 0.54 & 0.82 & 0.05 \\
\hline
Inclusion & Mammals & 0.84 & 0.06 & 0.06 \\
Fallacy & Birds & 0.84 & \gptfour{\textbf{$<$0.001 $\circ$}} & 1.0 \\
  & Vehicles & 0.54 & 0.54 & \human{\textbf{$<$0.001 *}} \\
\hline
\end{tabular}
\label{tbl:expt1stats}
\vspace{-3mm}
\end{table}

While Figure~\ref{fig:e1-main-results} illustrates performance visually, Table~\ref{tbl:expt1stats} characterizes it statistically. Since our theories primarily make predictions about \textit{which} argument (if any) is endorsed more rather than \textit{by how much}, we recode each response as a (-) or a (+), where (+) indicates that the theoretically-predicted stronger argument was actually rated as stronger and (-) indicates the opposite; the few ratings that were exactly even were discarded. For each of the three agents (humans, GPT-3.5, and GPT-4) and each of 33 phenomenon/domain splits, this yields a vector of +/- values; we can then use a one-sample sign test to compare them against the null hypothesis that there will be equal numbers of each. The numbers in each cell of Table~\ref{tbl:expt1stats} correspond to the p-value on the sign test for that agent, domain, and phenomenon; significant p-values indicate a significant preference for one argument or the other. Since the sign test does not capture directionality, some items are significant but in the opposite of the predicted direction, such that the agent favors the theoretically \textit{weaker} argument; these are indicated in Table~\ref{tbl:expt1stats} with the $\circ$ symbol.

There is a substantial difference across phenomena in the ability of the large language models to capture human behaviour. For some phenomena -- Specificity, Similarity, Typicality, and Monotonicity -- both humans and GPT-4 consistently endorsed the stronger argument more, in accordance with previous findings. For GPT-3.5, performance was poor; it was also usually bimodal, reflecting the fact that it often appeared to react based on superficial features of the prompt like the order the arguments appeared in.\footnote{In earlier work \citep{hanrpk22} 
we reported results suggesting that GPT-3.5 does capture phenomena including Similarity, Typicality, and Specificity, but in that evaluation we controlled for argument order by including prompts with both possible orders. The difference between our previous and current results exposes how heavily GPT-3.5 is influenced by superficial features of the prompt.}  Consistent with this, as shown in  ~\ref{sec:exp1phenomena}, the reasoning by GPT-4 anecdotally appears more appropriate than that presented by GPT-3.5, referencing relevant concepts (e.g., stating that ``robins are more representative of the typical bird'' when explaining its answer for  Typicality).

There are another set of phenomena -- Diversity, Asymmetry, and Inclusion Fallacy -- where people in our experiments did not robustly endorse the argument that is theoretically predicted to be stronger. For the most part, GPT-3.5 and GPT-4 did not do so either, with the exception that GPT-4 (unlike our participants) showed the Diversity effect for General arguments. We consider reasons for this puzzling pattern of results in the Discussion.

Perhaps most interestingly, neither GPT-3.5 nor GPT-4 captured human behaviour on the Non-Monotonicity phenomena. For General arguments, GPT-3.5 leans towards endorsing the \textit{opposite} argument from the one that is theoretically predicted and endorsed by our participants. GPT-4 performs fairly well for General arguments but shows a strong preference in the opposite direction from people for the Specific arguments. Indeed, GPT-4's stated reasoning often reflects the erroneous idea that additional examples should make a conclusion stronger: it makes an argument \textit{for} Monotonicity and \textit{against} Non-Monotonicity (\ref{sec:exp1phenomena}). We return to Non-Monotonicity and GPT-4's failure to capture this phenomenon in the Discussion.

\section{Experiment 2: Individual arguments}

Considering inductive phenomena in isolation is a useful starting point, but this approach is limited because multiple phenomena are relevant to some inferences, and these phenomena sometimes conflict. For example, from the perspective of Diversity \twoargument{flamingo}{albatross}{bird} is relatively strong because the premise categories are so different from each other. However, it is weak from the perspective of Typicality since the premise categories are atypical of birds.

In Experiment 2, we therefore moved beyond the individual phenomena in Table~\ref{tbl:phenomena} by assessing the ability of GPT-3.5 and GPT-4 to rate the inductive strength of relatively large sets of arguments. \citet{oshersonswls90} studied this in humans by asking participants to rank two sets of arguments involving mammals. One set included 36 two-premise Specific arguments such as \twoargument{cow}{chimp}{horse}, where the conclusion in all cases was \cat{horse}. The second included 45 three-premise \textbf{General} arguments such as \threeargument{horse}{cow}{mouse}{all mammals}, where the conclusion category was always \cat{all mammals}. The  \citet{oshersonswls90} dataset has been used in several subsequent studies \citep{kempt09,hanrpk22} but is limited because it includes a relatively small set of arguments from a single domain. As a result, our first step in Experiment 2 was to obtain judgments from humans on a much larger and more varied dataset of arguments. We then compared human judgments with ratings of argument strength elicited from GPT-3.5 and GPT-4 as described below.

Experiment 2 used a design in which agents were presented with a series of individual arguments and asked to rate each argument's strength in isolation. The experiment therefore departs from the approach of \citet{oshersonswls90}, who asked participants to rank argument sets, but follows the approach of \citet{glick11}, who also collected strength ratings for individual arguments. 

\subsection{Generating individual arguments}

For each of our three domains, we generated 100 two-premise general arguments (whose conclusion was \textsc{all mammals}, \textsc{all birds}, or \textsc{all vehicles}) and 100 two-premise specific arguments (whose conclusion was one of the 24 categories in that domain). We did this by first randomly sampling 10,000 arguments for each of the six domain-conclusion splits (3 domains $\times$ 2 conclusions, general or specific). After removing duplicates, we ranked the strength of each argument using the SCM. In order to ensure that our set of arguments contained the full range of strengths, we divided the ranked set of arguments into 25 equal width bins, then sampled four from each bin to construct the full set of 100 arguments. Sampling was random subject to the constraint that no premise set could appear more than three times and no category could appear more than 15 times.

For each set of 100 two-premise arguments we then constructed a corresponding set of single-premise arguments consisting of all of the premise-conclusion mappings from the two-premise set. Thus, for instance, the two-premise argument \twoargument{canary}{seagull}{stork} would correspond to the one-premise arguments \oneargument{canary}{stork} and \oneargument{seagull}{stork}.

The precise number of single premise arguments varied depending on the degree of overlap in the categories that were selected, but ranged from $24$ for general arguments to $169$ for specific mammal and bird arguments. In total, across all six domain-conclusion splits as well as both one-premise and two-premise  sets, there were $1168$ arguments.

\subsection{Presenting individual arguments to humans}

\subsubsection{Participants}

We recruited $610$ people via Amazon Mechanical Turk who were each paid \$1.00USD for the five minute study. All participants passed a screening for English language competency prior to participation and indicated informed consent via an online consent form. 

\subsubsection{Stimuli}

Each participant was randomly assigned to one of the six domain-conclusion splits and only rated arguments from that split. The arguments were presented in two blocks, one corresponding to one-premise arguments and one corresponding to two-premise arguments. Each participant saw one of 10 versions of each block, each based on a stratified sample of the 100 possible two-premise arguments for that domain and conclusion type.\footnote{Argument stratification was achieved using human similarity based SCM scores in order to ensure that each block contained arguments with a wide variety of strengths. There were 11 rather than 10 versions of each block in the General Birds split because two of the arguments in that split shared the same SCM score. Consequently some General Bird arguments have 11 rather than 10 participant ratings. All other aspects of this split are identical to the other splits.} Each participant completed 42 trials consisting of 10 trials for the two-premise block and 32 trials for the one-premise block. The 32 trials of the one-premise block were composed of one-premise arguments that corresponded with the two-premise block's arguments alongside a random sample of additional one-premise arguments to ensure that all participants saw the same number of arguments in the one-premise block. In addition to the 42 trials, participants were also shown two trial arguments at the start of each block as well as 4 attention check arguments every 8 trials. The attention check arguments were handcrafted to have an unambiguous answer (for example: \oneargument{All animals}{All mammals}).

\begin{figure}[t!]
\begin{center}
\includegraphics[width=\textwidth]{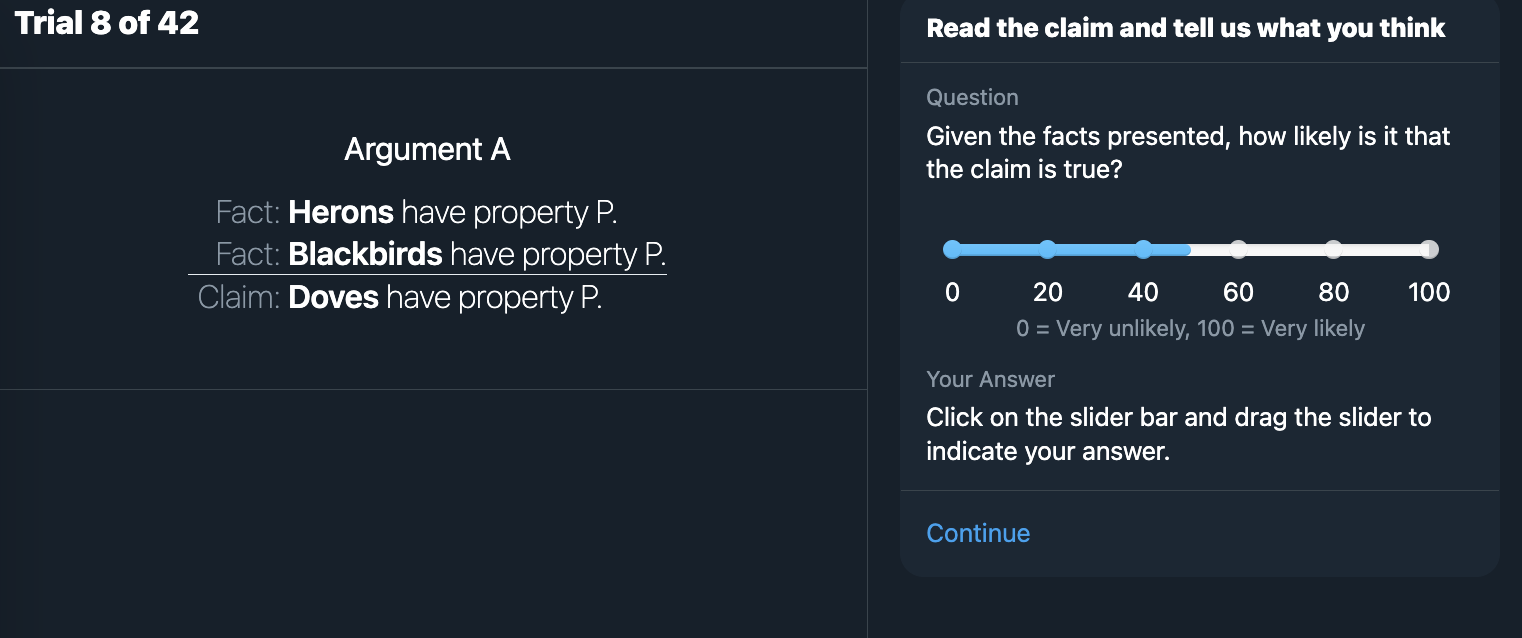}
\end{center}
\vspace{-1mm}
\caption{%
\textbf{Experiment 2 screenshot.} On each trial, participants were shown one argument on the left, and then asked to rate it using the slider on the right (the slider was always initialized to the midpoint but people were required to move it before they could continue to the next trial). All participants saw 42 arguments from one of the six domain-conclusion splits in two blocks, one for one-premise arguments and one for two-premise arguments. This screenshot contains a two-premise  argument from the domain of \textsf{Birds} with the specific conclusion category \textsc{Dove}. }
\label{fig:expt2screenshot}
\vspace{-3mm}
\end{figure}

\subsubsection{Procedure}

At the start of every session, participants were shown the following instructions:
\vspace{-2mm}
\begin{displayquote}
{\small \texttt{We're going to show you a series of claims relating to living things and the properties they share. Rather than mention any specific property (e.g. "Hyenas have sesamoid bones") we'll refer to an unspecified property (e.g. "Hyenas have property P"). Each claim may be true or false, and to help you decide which, we'll provide you with facts about whether or not other living things have the same property (e.g. "Lions have property P", and "Zebras have property P").}}
\end{displayquote}

People were then shown separate instructions for the one-premise and two-premise blocks. For each block, participants were told how many supporting facts to expect (one or two) and informed that their job was to rate how likely the claim was. They then saw two sample trials using fruit, and upon completing them began the main sequence of trials in that block.

On each trial, people were shown a single argument labeled \texttt{Argument A} on the left of the screen. On the right, they were asked ``Given the facts presented, how likely is it that the claim is true?'' They then selected their response using a slider with a scale ranging from 0 (very unlikely) to 100 (very likely). A screenshot from an example trial is shown in Figure~\ref{fig:expt2screenshot}.

\subsubsection{Replicating \citeauthor{oshersonswls90}}

To ensure that our study was broadly consistent with the original study by \citet{oshersonswls90}, we ran a version of our procedure that used the 36 specific arguments used by \citeauthor{oshersonswls90} rather than our LNCD based arguments. This experiment involved 40 participants who each rated one block of 10 multi-premise and one block of 10 single-premise Osherson arguments. In all other respects, its procedure was identical to what is described above. We found that across the 36 arguments, these rankings had a Spearman correlation of 0.65 (\textit{p} $<$ 0.001) with the original rankings obtained by \citeauthor{oshersonswls90}.

\subsection{Presenting individual arguments to GPT-4}

Our method for presenting individual arguments to GPT-4 is very similar to that of our first experiment. We again used the \texttt{gpt-4-0314} engine within the Chat API of OpenAI with temperature $t = 0$ and a maximum response length of 400 tokens. For each prompt design, we presented the $600$ two premise arguments and $568$ single premise arguments to GPT-4 only once.

The design of possible prompts was also very similar. The task agnostic instructions (the \textbf{System} message) were identical, as was the structure of the arguments being rated (\textbf{Arguments}). The explanation of the task (\textbf{Context}), judgment elicitation (\textbf{Question}), and the specification of the answer format (\textbf{Options}) were straightforwardly adapted to correspond to this task. The only major change was the addition of a new component (\textbf{Trials}) corresponding to the presence or absence of the two practice trials involving fruit; we added this because the participants in our experiment saw these. All variants are shown in~\ref{sec:exp2prompts} using the example of one-premise arguments; prompts for the two-premise arguments were exactly analogous.

As before, for resource reasons we identified a subset of prompts to explore rather than systematically testing hundreds. These corresponded to the four in Experiment 1 plus an additional one (S3-C1-A1-Q1-O1-T1) which included the practice trials, which we sequentially fed to GPT-4 in order to include its own answers to the trials in the final prompt. Performance was evaluated based on the correlation obtained between GPT-4 judgments and human judgments, described more fully in Section \ref{sec:exp2result}. The best-performing prompt was the one with the practice trials, shown in Table~\ref{tbl:expt2bestprompt}.

\begin{table}[t]
    \centering
        \caption{\small{\textbf{Best-performing GPT-4 prompt for Experiment 2.} It was identified by evaluating multiple prompts varying the  components of System, Context, Trials, Argument, Question, and Options. The column on the left indicates which variant of that component it corresponds to. 
        }}
        \vspace{1mm}
    \begin{tabular}{|p{0.05\linewidth} | p{0.9\linewidth}|}
    \hline
    S3 & \small{\texttt{You are an expert on X and the types of real world properties that they have. The questions you'll see don't have right or wrong answers, and you are willing to use your best judgment and commit to a concrete, specific response even in cases where you can't be sure that you are correct.}}\\
\vspace{1mm} C1 &  \vspace{1mm} \small{\texttt{We're going to show you a series of claims relating to living things and the properties they share. Rather than mention any specific property (e.g. "Hyenas have sesamoid bones") we'll refer to an unspecified property (e.g. "Hyenas have property P"). Each claim may be true or false, and to help you decide which, we'll provide you with facts about whether or not other living things have the same property (e.g. "Lions have property P", and "Zebras have property P").}} \\
\vspace{1mm} T1 &  \vspace{1mm} \small{\texttt{This section contains a series of claims that include only one supporting fact. Before we start, we'll give you two examples as practice.}}\\ 
 & \hspace{8mm}\textit{\small{\texttt{[insert two examples following the same format as the main trials]}}}\\
 & \small{\texttt{Now that you've practiced you're ready to continue on to the main trials for this section.}}\\
\vspace{1mm} A1 & \vspace{1mm} \small{\texttt{Argument A: Fact - Dogs have property P.}}\\
& \hspace{21mm} \small{\texttt{Claim - All mammals have property P.}}\\
\vspace{1mm} Q1 & \vspace{1mm} \small{\texttt{Question: Given the facts presented, how likely is it that the claim is true?}}\\
\vspace{1mm} O1 & \vspace{1mm} \small{\texttt{Indicate your answer by providing a number between 0 and 100, where 0 means that the claim is very unlikely and 100 means that the claim is very likely.}}\\
\hline
    \end{tabular}
    \label{tbl:expt2bestprompt}
\end{table}

\subsection{Presenting individual arguments to GPT-3.5}

Like our first experiment, we used the \texttt{text-davinci-003} engine within the OpenAI Completions API to elicit argument ratings from GPT-3.5, setting $t = 0$ alongside a maximum response length of 100 tokens. We again relied on GPT-3.5 token probabilities and presented our top performing prompt for GPT-4 to GPT-3.5, concatenating the system and user messages into a single prompt. To convert GPT-3.5 responses to argument ratings, we took the sum of the top five token completion probabilities, each multiplied by their numeric value. Again, this affords GPT-3.5 a methodological advantage in precision over GPT-4.

\subsection{Extracting similarity judgments from GPT-4}

Supplementing our analysis of argument ratings, we also extracted similarity judgments from GPT-4 and GPT-3.5. We used the same GPT-4 engine and parameters as our previous experiments, and set the response scale as 0-20 to match the original study by \citet{dedeyne2008}. We then presented GPT-4 with every category pair in the LNCD using the \texttt{user message} below. In each message, \texttt{X} was replaced by the category pair's domain name (e.g., \texttt{animals}) while \texttt{C1} and \texttt{C2} were replaced by the category names (e.g., \texttt{rabbits} and \texttt{hippos}).

\vspace{-2mm}
\begin{displayquote}
{\small \texttt{You are an expert on X and the various properties that they have. With these properties in mind, we will ask you to rate the similarity of two X on a scale of 0 to 20, where 0 means that the X have no similarity and 20 means that they are identical.\newline \newline Question: With their respective properties in mind, how similar are C1 and C2 on a scale of 0 to 20? Answer with a single number. 
Answer:
}}
\end{displayquote}

\subsection{Extracting similarity judgments from GPT-3.5}

To extract similarity judgments from GPT-3.5, we followed two approaches. The first was to present GPT-3.5 with the same prompt as GPT-4. As we did when extracting argument ratings, we calculated GPT's similarity rating as the sum of the top five token completion probabilities multiplied by the numeric value of each token; each rating was then converted to a ranking.

In the second approach, we passed each category name to the OpenAI Embeddings API using the \texttt{text-embedding-ada-002} endpoint, and then extracted an embeddings-based similarity score by calculating the cosine similarity between each LNCD category pair. Although the \texttt{text-embedding-ada-002} model is different from the \texttt{text-davinci-003} model that we used elsewhere, we used it because it is the best performing GPT-3 based model for embeddings-based applications \citep{gpt3embeddings}. We thus view its similarity ratings as a noisy indication of what is possible when similarity is measured using internal representations rather than completions. 

\subsection{Results}
\label{sec:exp2result}

\begin{figure}[t!]
\begin{center}
\includegraphics[width=\textwidth]{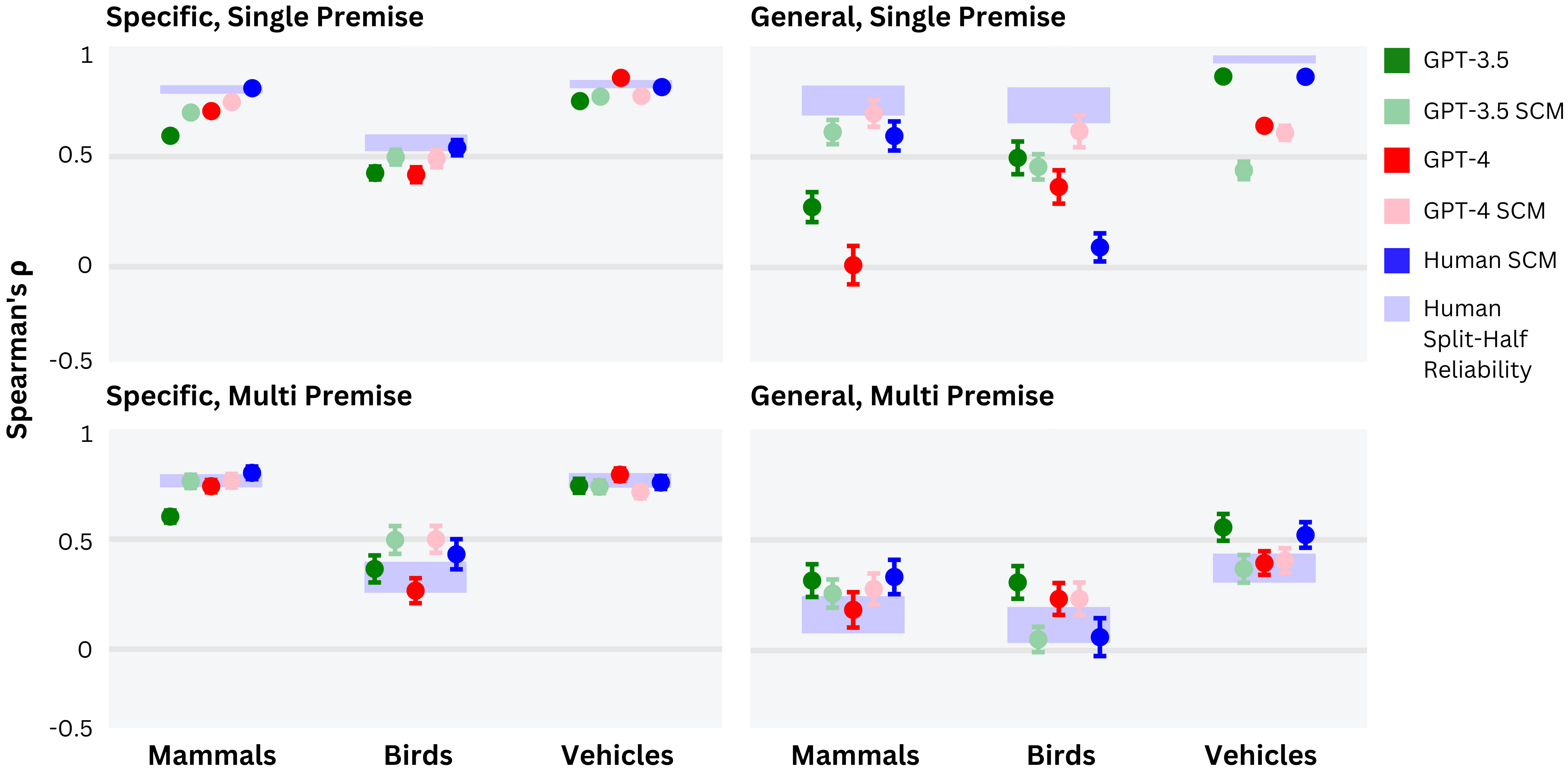}
\end{center}
\vspace{-1mm}
\caption{Correlations (Spearman's $\rho$) between human argument rankings and rankings of five different models.  Human SCM (a special-purpose model that applies the similarity-coverage model to human similarity ratings) performs best overall, but its performance is nearly matched by GPT-4 on Specific arguments and GPT-3.5 on General ones. The remaining two models apply the SCM to similarity ratings derived from GPT-4 and GPT-3.5. GPT-3.5 SCM improves on GPT-3.5 for Specific arguments, and GPT-4 SCM improves on GPT-4 for General arguments. The width of each split-half bar spans one standard error on either side of the mean. Results for GPT-3 and other GPT-3.5 models are shown in Appendix Table~\ref{fig:argrating_appendix}.} 
\label{fig:argrating}
\vspace{-4mm}
\end{figure}

Split-half reliabilities across 100 splits for each argument set are shown in Figure~\ref{fig:argrating}. The split-half reliabilities for multi-premise general arguments are relatively low, indicating a high level of variability across participants, and as a result we will not attempt to interpret differences in model performance for these arguments. In contrast, the split-half-reliabilities for single-premise general arguments and for specific arguments seem high enough to use these data for evaluating alternative models.

As a benchmark for comparison, we computed ratings of argument strength for a \textbf{Human SCM} model that applies the similarity-coverage model to the human similarity ratings collected by \citet{dedeyne2008}. This human SCM model is a special-purpose model from the literature that was specifically designed to capture judgments about the kinds of arguments included in our experiment. A general-purpose system such as GPT-4 has therefore performed relatively well if it accounts for our data as well as the human SCM model.

Figure~\ref{fig:argrating} shows the correlations between the five different models and our human ratings. We additionally perform quantitative comparisons between model pairs based on 1000 bootstrap samples for each model, and report those in Table~\ref{tbl:expt2stats}. 

For \textbf{Specific} arguments, Figure~\ref{fig:argrating} suggests that GPT-4 accounts for human ratings nearly as well as Human SCM. Table~\ref{tbl:expt2stats} shows that Human SCM outperforms GPT-4 for all four sets of Specific arguments about Mammals and Birds, but GPT-4 outperforms Human SCM for the two sets about Vehicles. Relative to GPT-4, GPT-3.5 performs worse for Specific arguments. That said, across all six sets, the differences between the models are relatively small.

\begin{table}[t!]
\centering 
\caption{\small{Comparisons between pairs of models for Experiment 2. Each comparison is based on 1000 bootstrap samples per model, and the entries in column ``M1 vs M2'' indicate the proportion of samples for which M1 correlates better with human ratings than does M2. Proportions greater than 0.95 indicate strong evidence in favor of M1, and smaller than 0.05 indicate strong evidence in favor of M2 (these are bold and colored according to the favored model). Human SCM, GPT-4 and GPT-3.5 are abbreviated as H, 4 and 3.}}
\vspace{1mm}
\resizebox{\textwidth}{!}{
\vspace{-2mm}
\begin{tabular}{|c|c|c|ccccc|}
\hline
\textbf{Conc}. & \textbf{Prems} & \textbf{Domain} &  \human{\textbf{H}} vs \gptfour{\textbf{4}}   & \human{\textbf{H}} vs \gptthreefivethree{\textbf{3}}  & \gptfour{\textbf{4}} vs \gptthreefivethree{\textbf{3}} & \gptfour{\textbf{4}} vs \gptfourscm{\textbf{4 SCM}} & \gptthreefivethree{\textbf{3}} vs \gptthreefivethreescm{\textbf{3 SCM}}\\
\hline
Specific & Single & Mammals & \human{\textbf{1.0}} & \human{\textbf{1.0}} & \gptfour{\textbf{1.0}} & \gptfourscm{\textbf{0.02}} & \gptthreefivethreescm{\textbf{0.0}}\\
& & Birds & \human{\textbf{0.99}} & \human{\textbf{0.99}} & 0.38 & \gptfourscm{\textbf{0.0}} & \gptthreefivethreescm{\textbf{0.0}}\\
& & Vehicles & \gptfour{\textbf{0.0}} & \human{\textbf{1.0}} & \gptfour{\textbf{1.0}} & \gptfour{\textbf{1.0}} & 0.13\\
\hline
Specific & Multi & Mammals & \human{\textbf{0.98}} & \human{\textbf{1.0}} & \gptfour{\textbf{1.0}} & 0.16 & \gptthreefivethreescm{\textbf{0.0}}\\
& & Birds & \human{\textbf{0.97}} & 0.86 & 0.07 & \gptfourscm{\textbf{0.0}} & \gptthreefivethreescm{\textbf{0.0}}\\
& & Vehicles & 0.05 & 0.71 & \gptfour{\textbf{1.0}} & \gptfour{\textbf{0.99}} & 0.57\\
\hline
General & Single & Mammals & \human{\textbf{1.0}} & \human{\textbf{0.99}} & \gptthreefivethree{\textbf{0.0}} & \gptfourscm{\textbf{0.0}} & \gptthreefivethreescm{\textbf{0.0}}\\
& & Birds & \gptfour{\textbf{0.0}} & \gptthreefivethree{\textbf{0.0}} & \gptthreefivethree{\textbf{0.05}} & \gptfourscm{\textbf{0.0}} & 0.69\\
& & Vehicles & \human{\textbf{1.0}} & 0.46 & \gptthreefivethree{\textbf{0.0}} & 0.83 & \gptthreefivethree{\textbf{1.0}}\\
\hline
General & Multi & Mammals & \human{\textbf{0.96}} & 0.60 & \gptthreefivethree{\textbf{0.02}} & 0.07 & 0.84\\
& & Birds & 0.05 & \gptthreefivethree{\textbf{0.0}} & 0.15 & 0.49 & \gptthreefivethree{\textbf{0.99}}\\
& & Vehicles & \human{\textbf{0.99}} & 0.13 & \gptthreefivethree{\textbf{0.0}} & 0.41 & \gptthreefivethree{\textbf{1.0}}\\
 \hline
\end{tabular}}
\label{tbl:expt2stats}
\end{table}

The results for \textbf{General} arguments reveal a more substantial gap between Human SCM and GPT-4. Table~\ref{tbl:expt2stats} shows that Human SCM outperforms GPT-4 on single-premise sets for Mammals and Vehicles, but that GPT-4 is superior for Birds. The difference between the models is  especially large for single-premise arguments about Mammals, where GPT-4 is uncorrelated with human ratings. Interestingly, GPT-3.5 performs better than GPT-4 for General arguments and seems roughly comparable to Human SCM for single-premise General arguments.

We did not anticipate the superiority of GPT-3.5 over GPT-4 for General arguments, but in retrospect this finding seems plausible. An important difference between the two models is that GPT-4 alone was trained using reinforcement learning from human feedback (RLHF), which aims to better align the model's output with human expectations. Among other benefits, RLHF makes the model less likely to ``hallucinate'' responses with no factual basis, but
also impairs the model's reasoning ability in some contexts \citep{gpt4}. Evaluating general arguments may be one of these contexts, because projecting a novel property to a superordinate category such as  \cat{all mammals} requires an inductive leap that goes well beyond the limited evidence provided by information about two or three individual mammals. Compared to GPT-3.5, GPT-4 may be less likely to take this inductive leap because RLHF encourages it to remain within the bound of established facts. As Figure~\ref{fig:argrating_appendix} demonstrates, both the GPT-4 and GPT-3.5-chat-turbo models tend to perform worse than non-RLHF GPT variants for general arguments, and for GPT-3.5-chat-turbo in particular we find that almost a quarter of all responses to general arguments explicitly mention neutrality or suggest that the answer is `difficult to determine' given the provided information.

Although neither GPT-3.5 nor GPT-4 accounts as well as human SCM for both Specific and General arguments, the strong performance of GPT-4 for Specific arguments and GPT-3.5 for General arguments suggests that a single GPT model (perhaps GPT-4 trained with less RLHF) might be able to account for our data as well as Human SCM. Overall, our results suggest that large language models are broadly able to capture inductive inferences about blank properties about as well as the best special-purpose models available in the psychological literature.

\begin{figure}[t!] 
\begin{center}
\includegraphics[width=\textwidth]{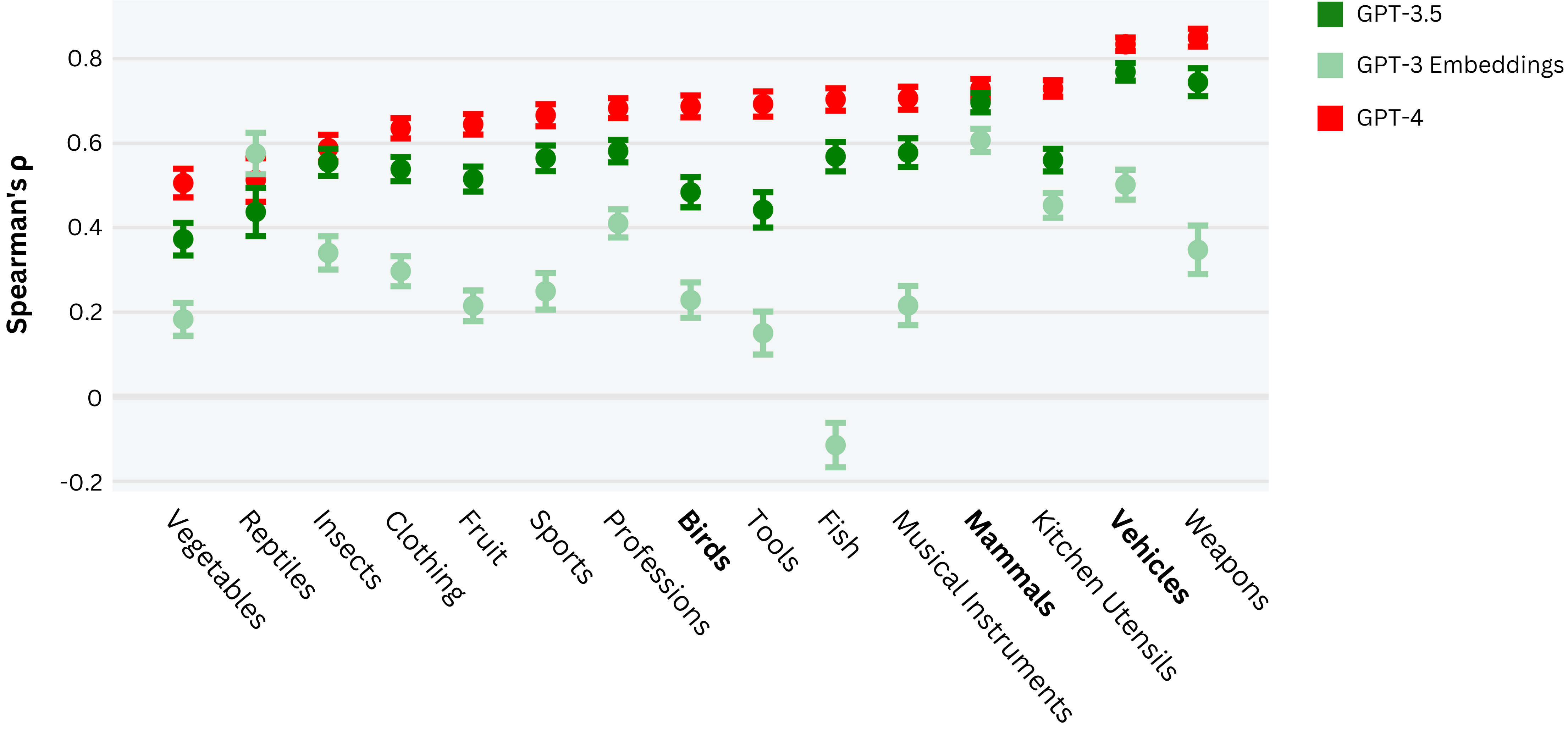}
\end{center}
\caption{Correlations (Spearman's $\rho$) between human similarity ratings and similarity ratings derived from GPT-4, GPT-3.5 and GPT-3 Embeddings. Error bars show standard error of the mean across 1000 bootstrap samples.} 
\vspace{-4mm}
\label{fig:representation}
\end{figure}

\subsection{Disentangling representation and reasoning}

Evaluating an inductive argument requires some kind of reasoning process that operates over representations of the categories in the argument. For connectionist models such as GPT-4 and GPT-3.5 it may be difficult to establish a sharp division between representation and reasoning \citep{rogersm14}, but we can still develop analyses that attempt to separate the two.

\subsubsection{Representation}
To probe the representational abilities of GPT-3.5 and GPT-4 we used the similarity judgments extracted from these models by the procedure described above. Although these similarity judgments are not direct measures of internal representations, we take them as evidence of underlying representations in the same way that human similarity judgments are often taken as evidence of underlying mental representations. Figure~\ref{fig:representation} shows correlations between these similarity judgments and the human similarity judgments reported by \cite{dedeyne2008}. Most relevant to us are results for the three domains considered in our experiments (\textsf{Mammals}, \textsf{Birds}, and \textsf{Vehicles}), but we include results for all of the domains in \citet{dedeyne2008}.

The correlations between GPT-4 and human similarity ratings are relatively high and exceed 0.6 for all three domains in our experiments. The prompt-based correlations achieved by GPT-3.5 are lower across all domains but still exceed 0.4 for all three of the domains in our experiments. In contrast, the embedding-based results are relatively poor, and we therefore use prompt-based GPT-3.5 similarities in our subsequent analyses. 

Our results in Figure~\ref{fig:representation} parallel previous findings that the internal representations of LLMs can be used to make relatively accurate predictions about human similarity judgments \citep{bhatiar21}. All of these findings therefore suggest that the category representations of both GPT-4 and GPT-3.5 should be sufficient to allow human-like ratings of inductive strength. 

\subsubsection{Reasoning}

We now ask how well the GPT models exploit their representational abilities when evaluating inductive arguments. To do so we consider \textbf{GPT-4 SCM} and \textbf{GPT-3.5 SCM}, two hybrid models that apply the SCM (a special-purpose model of property induction) to similarity ratings derived from the GPT models. According to the SCM, the strength of a specific argument with premises $P$ and conclusion $c$ is
\begin{equation}
\text{SCM}_{GPT-N}(P, c) = \alpha \cdot \max_{p \in P} \text{sim}_{GPT-N}(p, c) + \frac{1 - \alpha}{|C|} \sum_{c_i \in C} \max_{p \in P} \text{sim}_{GPT-N}(p, c_i),
\end{equation}
where the premise and conclusion categories are all drawn from a finite  set of categories called domain \(C\).
In \cite{oshersonswls90} \(\ 0 < \alpha < 1\) is left as a free parameter, and here we set \(\alpha\) to 0.5.

For general arguments with premises \(P\) that all belong to a conclusion domain \(C\), the GPT-SCM rates argument strength using:
\begin{equation}
\text{SCM}_{GPT-N}(P, C) = \frac{1}{|C|} \sum_{c_i \in C} \max_{p \in P} \text{sim}_{GPT-N}(p, c_i).
\end{equation}
In both cases, \(\text{sim}_{GPT-N}(c_i, c_j)\) refers to the similarity rating derived from a GPT-N model for categories \(c_i\) and \(c_j\). 

Comparing against a  GPT-SCM model can potentially expose a gap between the representational and reasoning abilities of a GPT model. For example, if GPT-3.5 captures human similarity ratings relatively well but does not use its representations effectively when evaluating inductive arguments, then GPT-3.5 SCM should account for our human data better than GPT-3.5. On the other hand, if GPT-3.5 reasons as well as possible given the representations that contribute to its similarity ratings, then GPT-3.5 SCM should perform no better than GPT-3.5.

Figure~\ref{fig:argrating} and Table~\ref{tbl:expt2stats} show that GPT-4 SCM does better than GPT-4 on three of the six sets of \textbf{Specific} arguments and  worse on two: this suggests that there is no major improvement of the hybrid model relative to GPT-4. By contrast, GPT-3.5 SCM performs better than GPT-3.5 across four of the six specific data sets, and is worse on none of them. Overall, these results suggest that combining GPT similarity ratings with the SCM provides more of a boost for GPT-3.5 than GPT-4, which in turn supports the conclusion that GPT-3.5 is more limited in its reasoning capacities than is GPT-4.

For \textbf{General} arguments, the pattern of results is again reversed. GPT-4 SCM improves relative to GPT-4 on two of the three single-premise argument sets, but GPT-3.5 SCM is comparable to GPT-3.5 (better on one set and worse on another).  These results are consistent with the possibility that GPT-4's ability to handle General arguments is relatively poor and may have been compromised by RLHF. On the other hand, on General arguments GPT-3.5 appears to reason as well as could be expected given its category representations.

Throughout this section we have evaluated the GPT models against human judgments and have implicitly assumed that a model that differs from humans is reasoning relatively poorly. It is possible, however, that the models differ from humans because they are less prone to inferential errors and are therefore reasoning better than our experimental participants. Our primary goal has been to explore the extent to which LLMs make human-like inferences, which means that the normative status of these inferences has not been a major concern. Future work, however, can evaluate both human participants and GPT models against normative accounts of inductive reasoning \citep{heit98,tenenbaumgk06,kempt09} with the aim of identifying ways in which the models may be better and worse than humans.

\section{Discussion} 

We developed two new benchmark datasets that are directly inspired by previous psychological work on property induction but incorporate a much larger set of categories, arguments, and domains. Across two studies, we  compared responses from GPT-3.5 and GPT-4 with the human inferences in these datasets. Experiment 1 focused on 11 qualitative phenomena including Similarity, Typicality, Diversity, and Non-Monotonicity; of them, only Non-Monotonicity (Specific) was observed in our participants but not well captured by either GPT model. Experiment 2 asked models and humans to provide quantitative ratings of argument strength, and found that the best results from GPT-3.5 and GPT-4 were comparable with results obtained by the similarity-coverage model, a special-purpose model of inductive reasoning that incorporates human similarity judgments. Taken together, the experiments therefore suggest that GPT-style models broadly perform well at the tasks that we used, and capture everything that we considered except Non-Monotonicity.

We began this project when GPT-3 was the most advanced GPT model available, and expected at the time that it might capture effects related to Similarity and Typicality but would struggle to capture Diversity and Non-Monotonicity. Recall that Non-Monotonicity describes the phenomenon in which adding premises results in \textit{weaker} arguments: the three-premise argument \threeargument{crow}{peacock}{rabbit}{bird} is considered less strong than the corresponding two-premise argument \twoargument{crow}{peacock}{bird}. One intuitive explanation for why people appear to reason this way is that they implicitly assume that the premises are being generated by a helpful provider of information \citep{ransompn16,voorspoelsetal15,medincsh05}. Given a desired conclusion of \cat{bird}, a helpful provider would be much more likely to offer the premises \cat{crow} and \cat{peacock} alone than to also include \cat{rabbit}; this is because including \cat{rabbit} implicitly changes the context to favor a conclusion like \cat{mammal}. 

The relationship between Non-Monotonic reasoning and assumptions about the data generation process  becomes even more obvious when considering other Non-Monotonic argument pairs like \oneargument{tiger}{mammal} vs \twoargument{tiger}{lion}{mammal}. If the conclusion truly is \textsc{all mammals} it seems a strange coincidence indeed (or evidence of incompetence or unhelpfulness on the part of the provider) for both premises to involve large cats: a helpful provider should offer clues that cover the conclusion category as well as possible. In addition to being intuitively appealing, this chain of thought can be formalized as Bayesian reasoning based on assumptions about how data is sampled \citep{ransompn16}. GPT-4's failure to capture non-monotonicity (specific) therefore suggests that it may fail to reason about how the premises of an inductive argument were generated.

Diversity is a second phenomenon that has been justified on the basis of sampling assumptions \citep{Hayes19}, and is partially captured by GPT-4. Our results for Diversity, however, are puzzling in two respects. First, GPT-4 captured Diversity for arguments with general conclusions (e.g., \cat{mammal}), but did not show Diversity effects when the conclusions were specific (e.g., \cat{rabbit}).   Second, our human participants did \textit{not} show any Diversity effect at all, and it is important to ask why we did not replicate the work of \citet{oshersonswls90} in this respect. 

The design for Experiment 1 followed the work of \citet{oshersonswls90} relatively closely, and in retrospect we see no obvious problems with the design or the stimuli. That said, one difference between our experiment and that of \citet{oshersonswls90} is that our participants not only saw stimuli from more domains (\textsf{Vehicles} and \textsf{Birds} as well as \textsf{Mammals}), but also a wider selection of categories within a single domain than theirs did. It is possible that this experimental context created such a large level of \textit{background} diversity that it dwarfed the much smaller variation within the key Diversity argument pairs we tested. This is consistent with previous work suggesting that Diversity may be more elusive or context-dependent than many of the other phenomena in Experiment 1 \citep{lopezggs92,lopezacms97,proffittcm00}. For example, \citet{lopezggs92} found support for Similarity, Typicality and  Specificity in kindergarteners but no evidence for Diversity.  If Diversity-based reasoning is relatively fragile, it may have little chance of emerging under circumstances when participants must quickly rate a large number of argument pairs from a wide variety of domains. 

\section{Strengths and limitations of our data} 

Compared to existing property induction datasets, our data offer several advantages. First, our data are relatively comprehensive; as far as we know these datasets are the largest set of human property induction judgments ever collected. Second, the arguments in these data sets were generated according to systematic procedures described above, which improves on the common practice of working with a small set of hand-crafted arguments. Finally, our data have been released publicly, which allows future researchers to replicate our results, evaluate future LLMs against our data, and explore aspects of our data that we have not considered ourselves.

Despite these strengths, our data are subject to some important limitations. In addition to our failure to replicate the effect of Diversity, our data are somewhat noisy. For instance, Experiment 2 was inspired by a study of \citet{oshersonswls90} that asked participants to rank a set of arguments, but we felt that GPT-3.5 (the best LLM at the time) would be unable to handle this task. We therefore asked participants to rate the strength of individual arguments, which allowed us to give the same task to humans and LLMs. It also made it straightforward to collect data for a relatively large set of arguments per domain.  Single argument ratings, however, are intrinsically noisy, which means that our data may not provide enough resolution to identify all of the factors that influence inductive reasoning in humans. Our first version of Experiment 2 followed \citet{oshersonswls90} in using general arguments with three premises, but the resulting data were so noisy that we considered them unreliable. We therefore simplified the task and used general arguments with two premises, but as suggested above the split-half reliabilities for multi-premise general arguments are still relatively low.  

Because a ranking task allows for direct comparisons between arguments, ranking and rating are fundamentally different tasks that may expose different aspects of inductive reasoning, and future work should consider both tasks. Of the two, rating is the easier target for LLMs because ranking requires information to be integrated across context windows that include at least two arguments, and requires an element of planning: in order to succeed, one must first find the strongest argument, then the next strongest, and so on. Given that GPT-4 appears to provide a good account of human argument ratings, future studies should explore whether it still performs well when evaluated against humans using a ranking task.

A final limitation of our data is that we worked with a convenience sample recruited through Amazon Mechanical Turk. All of our participants   previously passed a manually assessed qualification task measuring English proficiency and meticulousness, and are thus consistently more reliable than the general MTurk population. However, this does not mean that this participant pool is completely unproblematic. Property induction phenomena are known to vary across cultures \citep{lopezacms97}, developmental stages \citep{lopezggs92}, and levels of expertise \citep{proffittcm00}, and it is not entirely clear which group of people provide the most natural comparison to LLMs such as GPT-4. From one perspective, GPT-4 has been trained on a vast body of information about mammals, birds, and vehicles, and may therefore be best compared with human experts. It seems possible, for example, that our participants know relatively little about birds, and that GPT-4's responses would match the responses of domain experts more closely than they matched the responses of our experimental participants. If so, then evaluating GPT-4 against our human data could potentially underestimate its true abilities.

\subsection{What are the GPT models doing?}

Our analyses characterize the performance of GPT models on property induction tasks in some detail, but we have relatively little  insight into why the models behave as they do. At one level we understand these models: we know that they are trained on vast corpora on the task of predicting the next token, and that GPT-4 goes through a second training phase that relies on reinforcement learning from human feedback. But at another level these models remain opaque and mysterious, and it is unclear how this training gives rise to some of the behaviour we observe. 

At least three approaches can be used to develop a better understanding of large language models. We refer to them here as \emph{LLM psychology}, \emph{LLM neuroscience} and \emph{LLM ecological analysis}. LLM Psychology \citep{binzs23,hagendorff23} studies language models using behavioral experiments, computational analyses and all of the other techniques that psychologists have used to study human cognition and behavior. Our work offers several examples of this. We relied heavily on behavioral experiments to generate comparison data, and used related techniques to explore how GPT models respond to inductive problems as well as how sensitive they are to the way in which they are prompted. Moreover, we also developed computational analyses that build on psychological models such as the similarity-coverage model. 

All of the analyses in this paper are thus examples of \emph{LLM cognitive psychology}. However, future work could draw on ideas from other branches of psychology. For example, \emph{LLM developmental psychology} might aim to characterize the order in which inductive phenomena appear over the course of training a language model, and the resulting developmental sequence could be compared with analogous developmental sequences in humans \citep{lopezggs92}.

While \emph{LLM psychology} focuses on data from behavioral studies,  \emph{LLM neuroscience} goes beyond these studies by directly probing the mechanisms that give rise to behavior. We took a simple step in this direction by extracting and working with embeddings that are likely to approximate the internal representations of GPT-3. Probing the internal mechanisms that emerge in LLMs is an active area of research, but there are already several approaches that could be used to gain a better understanding of how GPT models evaluate inductive arguments \citep{lina21,olsson22context,voita19}. For example, since LLMs incorporate a set of attention weights, analyzing these weights may reveal systematic regularities in what parts of the premise(s) are weighted most highly by a LLM while assessing an argument's conclusion. A major challenge in applying LLM neuroscience to GPT-3.5 and GPT-4 is that these models are not open source, but other LLMs such as LLaMA \citep{llama23} are publicly available and their internal mechanisms can be studied in great detail. 

A third approach, \emph{LLM ecological analysis}, focuses less on the computations carried out by LLMs and more on understanding the structure of the data on which they are trained. These data constitute the ``environment'' of the LLM, and studying a LLM by analyzing its training data is reminiscent of ``rational analysis'' \citep{anderson90} and other research programs \citep{brunswik57,simon70,toddg07} that aim to understand an organism's behavior by characterizing the environment in which the organism is embedded. For studies of property induction, it may be valuable to explore the extent to which premise and conclusion categories co-occur in the training data, and to investigate whether and how LLMs go beyond models of inductive reasoning that rely on co-occurrence alone. A more urgent question is the extent to which the test tasks themselves (or highly similar ones) were available in the training data~\citep{magar2022data,frank23,mitchell23}. Most of the arguments in our data sets were generated for this project rather than drawn from the literature, so it seems unlikely that GPT-4 has previously seen these exact arguments. Even so, the training data for GPT-4 presumably contain numerous discussions of property induction, probably including the highly-cited paper \citep{oshersonswls90} that introduced the inductive phenomena we considered in Experiment 1. It seems possible that these components of the training data are at last partially responsible for the high level of performance achieved by GPT-4. Because the training data for GPT-3.5 and GPT-4 are not publicly available, any systematic work on LLM ecological analysis will need to consider models like LLaMA \citep{llama23} where that data is openly accessible.

\subsection{Other inductive reasoning tasks}

There is room for multiple future studies to explore how LLMs respond to the items in our two datasets, but future work should also explore how LLMs respond to other inductive tasks. Here we focused on inferences about blank properties, and psychological work in this tradition has identified several phenomena that go beyond the ones we examined \citep{sloman93,medincsh05}. For example, \citet{medincsh05} show that causal relationships lead to phenomena including
 causal asymmetry (\oneargument{carrot}{rabbit} is stronger than \oneargument{rabbit}{carrot}), causal violations of similarity (\oneargument{banana}{monkey} is stronger than \oneargument{mouse}{monkey}) and diversity (\twoargument{flea}{butterfly}{sparrow} is stronger than \twoargument{flea}{dog}{sparrow}). The same authors show that arguments in which all premise categories share a salient property can  lead to non-monotonicity (e.g.\ \oneargument{brown bear}{buffalo} is stronger than \threeargument{brown bear}{polar bear}{grizzly bear}{buffalo}).  We gave examples including the ones just described to GPT-3.5 and GPT-4, and the results in \ref{sec:otherpiphenomena} suggest that GPT-4 is sensitive to causal relationships between categories but may often struggle with non-monotonicity.
 
When evaluating an inductive argument, knowledge about the property in question is usually just as important as knowledge about the premise and conclusion categories, and several studies have documented property effects in inductive reasoning \citep{gelmanm86,smithso93,heitr94}. For example, \citet{gelman88} shows that even young children reason differently about biological properties (e.g.\ ``has a spleen inside'') and non-generalizable properties (e.g.\ ``is dirty'').  \citet{heitr94} considered both anatomical properties (e.g.\ ``has sesamoid bones'') and behavioral properties (e.g. ``usually gathers large amounts of food at once''), and found that these properties interacted with anatomical and behavioral similarity between premise and conclusion categories (e.g.\ \oneargument{goldfish}{shark} is stronger than \oneargument{wolf}{shark} for anatomical properties but not for behavioral properties). Although there are numerous empirical studies of property effects in inductive reasoning, to our knowledge there are no psychological models that can capture these effects in a general way; perhaps the closest is a formal account of human plausible reasoning developed by \citet{collinsm89}. LLMs therefore qualify as the first ever models that seem theoretically capable of capturing a broad range of property effects, and future work should compare them with humans using systematic benchmarks that include inductive arguments with a variety of non-blank properties. \ref{sec:otherpiphenomena} includes example responses for a handful of relevant cases, and suggests that GPT-4 will struggle to capture many of the property effects documented in the literature.

The psychological literature on property induction is somewhat distinct from work on generalization \citep{shepard87}, categorization \citep{pothosw11}, identification \citep{kempcl10}, and analogical reasoning \citep{vosniadouo89}, but all of these topics can be viewed as special cases of inductive reasoning \citep{kempj14}. According to one standard definition, an inference is inductive (or ampliative, or defeasible) if it reaches at a conclusion that does not follow with certainty from the available evidence \citep{hollandhnt86,chaterohh11}. From this perspective there are many different inductive problems, which opens up the possibility for a line of work that applies LLMs to inductive problems drawn from many parts of the psychological literature. Existing work along these lines includes studies that evaluate LLMs on commonsense reasoning tasks from benchmarks such as BIG-bench \citep{bigbench}, and a study that explores analogical reasoning in LLMs \citep{webbhl22}.  

\ref{sec:otherphenomena} includes responses of GPT-3.5 and GPT-4 to a range of other inductive problems, and these examples suggest the possibility of developing multiple new benchmarks that explore different aspects of inductive reasoning. Although the existing literature on LLMs explores a wide range of reasoning problems, our impression is that deductive tasks are currently slightly more prominent than inductive tasks. For example, 59 of the 204 BIG-bench are tagged as ``logical reasoning'' tasks, and 50 are tagged as ``commonsense reasoning''~\citep{bigbench}. \citet[p 553]{chaterohh11}, however, suggest that ``many, and perhaps even almost all, inferences outside mathematics involves uncertain inductive inference'' (p 553). Although inductive and deductive reasoning are both important, future work on LLMs should perhaps prioritize inductive reasoning because the majority of reasoning problems encountered in everyday situations are inductive rather than deductive.

\section{Conclusion} 

We compared inductive inferences in reasoning in humans and large language models, and found that GPT-4 provides a relatively good account of property induction in humans. At the time we began this project GPT-3 was the most advanced model available to us, and we correctly anticipated that this model would struggle to account for many aspects of our data. GPT-4, however, performs substantially better, which motivates future work on property effects and other inductive phenomena that may be more challenging to capture than most of the phenomena considered here. 

Our work draws on AI and psychology and holds lessons for both fields. For AI, our work suggests that evaluations of LLMs can draw on psychological work on inductive reasoning. Moreover, our datasets represent a step towards a large scale evaluative benchmark that could be considered alongside other popular benchmarks. For psychology, our work suggests that comprehensive benchmarks similar to those used in evaluating LLMs can also valuable for understanding how humans learn and reason.  In comparing LLMs and people we realized that we do not have a convincing theoretical account on \textit{either} side of the comparison: not only do we lack an understanding of how GPT-4 succeeds on these tasks, but we also lack a detailed picture of how humans perform as well. Psychological models such as the similarity-coverage model are useful starting points, but they only account partially for our data, and it is unclear whether they can be extended to handle phenomena such as property effects. We hope that large language models point the way towards psychological models that come closer to capturing the rich intricacy of human inductive reasoning.

\section{Author Contributions}

S.J.H, K.R and C.K. wrote code for the project. S.J.H, A.P and C.K wrote the paper. All authors discussed the models and analyses and commented on the manuscript.  

\bibliography{references}

\begin{thebibliography}{}

\bibitem [\protect \citeauthoryear {%
Anderson%
}{%
Anderson%
}{%
{\protect \APACyear {1990}}%
}]{%
anderson90}
\APACinsertmetastar {%
anderson90}%
\begin{APACrefauthors}%
Anderson, J\BPBI R.%
\end{APACrefauthors}%
\unskip\
\newblock
\APACrefYear{1990}.
\newblock
\APACrefbtitle {The Adaptive Character of Thought} {The adaptive character of
  thought}.
\newblock
\APACaddressPublisher{Hillsdale, NJ}{Lawrence Erlbaum Associates}.
\PrintBackRefs{\CurrentBib}

\bibitem [\protect \citeauthoryear {%
Bhatia%
\ \BBA {} Richie%
}{%
Bhatia%
\ \BBA {} Richie%
}{%
{\protect \APACyear {2021}}%
}]{%
bhatiar21}
\APACinsertmetastar {%
bhatiar21}%
\begin{APACrefauthors}%
Bhatia, S.%
\BCBT {}\ \BBA {} Richie, R.%
\end{APACrefauthors}%
\unskip\
\newblock
\APACrefYearMonthDay{2021}{}{}.
\newblock
{\BBOQ}\APACrefatitle {Transformer Networks of Human Conceptual Knowledge}
  {Transformer networks of human conceptual knowledge}.{\BBCQ}
\newblock
\APACjournalVolNumPages{Psychological Review}{}{}{}.
\newblock
\begin{APACrefDOI} \doi{10.1037/rev0000319} \end{APACrefDOI}
\PrintBackRefs{\CurrentBib}

\bibitem [\protect \citeauthoryear {%
Binz%
\ \BBA {} Schulz%
}{%
Binz%
\ \BBA {} Schulz%
}{%
{\protect \APACyear {2023}}%
}]{%
binzs23}
\APACinsertmetastar {%
binzs23}%
\begin{APACrefauthors}%
Binz, M.%
\BCBT {}\ \BBA {} Schulz, E.%
\end{APACrefauthors}%
\unskip\
\newblock
\APACrefYearMonthDay{2023}{}{}.
\newblock
{\BBOQ}\APACrefatitle {Using cognitive psychology to understand {GPT-3}} {Using
  cognitive psychology to understand {GPT-3}}.{\BBCQ}
\newblock
\APACjournalVolNumPages{Proceedings of the National Academy of
  Sciences}{120}{6}{e2218523120}.
\PrintBackRefs{\CurrentBib}

\bibitem [\protect \citeauthoryear {%
Brewka%
, Dix%
, Konolige%
\BCBL {}\ \protect \BOthers {.}}{%
Brewka%
\ \protect \BOthers {.}}{%
{\protect \APACyear {1997}}%
}]{%
brewka1997}
\APACinsertmetastar {%
brewka1997}%
\begin{APACrefauthors}%
Brewka, G.%
, Dix, J.%
, Konolige, K.%
\BCBL {}\ \BOthersPeriod {.}\end{APACrefauthors}%
\unskip\
\newblock
\APACrefYear{1997}.
\newblock
\APACrefbtitle {Nonmonotonic reasoning: an overview} {Nonmonotonic reasoning:
  an overview}\ (\BVOL~73).
\newblock
\APACaddressPublisher{}{CSLI publications Stanford}.
\PrintBackRefs{\CurrentBib}

\bibitem [\protect \citeauthoryear {%
Brunswik%
}{%
Brunswik%
}{%
{\protect \APACyear {1957}}%
}]{%
brunswik57}
\APACinsertmetastar {%
brunswik57}%
\begin{APACrefauthors}%
Brunswik, E.%
\end{APACrefauthors}%
\unskip\
\newblock
\APACrefYearMonthDay{1957}{}{}.
\newblock
{\BBOQ}\APACrefatitle {Scope and aspects of the cognitive problem} {Scope and
  aspects of the cognitive problem}.{\BBCQ}
\newblock
\BIn{} H.~Gruber, K\BPBI R.~Hammond\BCBL {}\ \BBA {} R.~Jessor\ (\BEDS),
  \APACrefbtitle {Contemporary approaches to cognition} {Contemporary
  approaches to cognition}\ (\BPGS\ 5--31).
\newblock
\APACaddressPublisher{}{Harvard University Press}.
\PrintBackRefs{\CurrentBib}

\bibitem [\protect \citeauthoryear {%
Bubeck%
\ \protect \BOthers {.}}{%
Bubeck%
\ \protect \BOthers {.}}{%
{\protect \APACyear {2023}}%
}]{%
bubeck23}
\APACinsertmetastar {%
bubeck23}%
\begin{APACrefauthors}%
Bubeck, S.%
, Chandrasekaran, V.%
, Eldan, R.%
, Gehrke, J.%
, Horvitz, E.%
, Kamar, E.%
\BDBL {}others%
\end{APACrefauthors}%
\unskip\
\newblock
\APACrefYearMonthDay{2023}{}{}.
\newblock
{\BBOQ}\APACrefatitle {Sparks of artificial general intelligence: Early
  experiments with {GPT-4}} {Sparks of artificial general intelligence: Early
  experiments with {GPT-4}}.{\BBCQ}
\newblock
\APACjournalVolNumPages{arXiv preprint arXiv:2303.12712}{}{}{}.
\PrintBackRefs{\CurrentBib}

\bibitem [\protect \citeauthoryear {%
Carey%
}{%
Carey%
}{%
{\protect \APACyear {1985}}%
}]{%
carey85}
\APACinsertmetastar {%
carey85}%
\begin{APACrefauthors}%
Carey, S.%
\end{APACrefauthors}%
\unskip\
\newblock
\APACrefYear{1985}.
\newblock
\APACrefbtitle {Conceptual change in childhood} {Conceptual change in
  childhood}.
\newblock
\APACaddressPublisher{Cambridge, MA}{MIT Press}.
\PrintBackRefs{\CurrentBib}

\bibitem [\protect \citeauthoryear {%
Chang%
\ \BBA {} Bergen%
}{%
Chang%
\ \BBA {} Bergen%
}{%
{\protect \APACyear {2023}}%
}]{%
changb23}
\APACinsertmetastar {%
changb23}%
\begin{APACrefauthors}%
Chang, T\BPBI A.%
\BCBT {}\ \BBA {} Bergen, B\BPBI K.%
\end{APACrefauthors}%
\unskip\
\newblock
\APACrefYearMonthDay{2023}{}{}.
\newblock
{\BBOQ}\APACrefatitle {Language Model Behavior: A Comprehensive Survey}
  {Language model behavior: A comprehensive survey}.{\BBCQ}
\newblock
\APACjournalVolNumPages{arXiv preprint arXiv:2303.11504}{}{}{}.
\PrintBackRefs{\CurrentBib}

\bibitem [\protect \citeauthoryear {%
Chater%
, Oaksford%
, Hahn%
\BCBL {}\ \BBA {} Heit%
}{%
Chater%
\ \protect \BOthers {.}}{%
{\protect \APACyear {2011}}%
}]{%
chaterohh11}
\APACinsertmetastar {%
chaterohh11}%
\begin{APACrefauthors}%
Chater, N.%
, Oaksford, M.%
, Hahn, U.%
\BCBL {}\ \BBA {} Heit, E.%
\end{APACrefauthors}%
\unskip\
\newblock
\APACrefYearMonthDay{2011}{}{}.
\newblock
{\BBOQ}\APACrefatitle {Inductive logic and empirical psychology} {Inductive
  logic and empirical psychology}.{\BBCQ}
\newblock
\BIn{} D\BPBI M.~Gabbay, S.~Hartmann\BCBL {}\ \BBA {} J.~Woods\ (\BEDS),
  \APACrefbtitle {Handbook of the History of Logic. {V}olume 10: Inductive
  Logic} {Handbook of the history of logic. {V}olume 10: Inductive logic}\
  (\BPGS\ 553--624).
\newblock
\APACaddressPublisher{Amsterdam}{North Holland}.
\PrintBackRefs{\CurrentBib}

\bibitem [\protect \citeauthoryear {%
Collins%
\ \BBA {} Michalski%
}{%
Collins%
\ \BBA {} Michalski%
}{%
{\protect \APACyear {1989}}%
}]{%
collinsm89}
\APACinsertmetastar {%
collinsm89}%
\begin{APACrefauthors}%
Collins, A.%
\BCBT {}\ \BBA {} Michalski, R.%
\end{APACrefauthors}%
\unskip\
\newblock
\APACrefYearMonthDay{1989}{}{}.
\newblock
{\BBOQ}\APACrefatitle {The logic of plausible reasoning: A core theory} {The
  logic of plausible reasoning: A core theory}.{\BBCQ}
\newblock
\APACjournalVolNumPages{Cognitive Science}{13}{1}{1--49}.
\PrintBackRefs{\CurrentBib}

\bibitem [\protect \citeauthoryear {%
Davis%
\ \BBA {} Marcus%
}{%
Davis%
\ \BBA {} Marcus%
}{%
{\protect \APACyear {2015}}%
}]{%
davis2015}
\APACinsertmetastar {%
davis2015}%
\begin{APACrefauthors}%
Davis, E.%
\BCBT {}\ \BBA {} Marcus, G.%
\end{APACrefauthors}%
\unskip\
\newblock
\APACrefYearMonthDay{2015}{}{}.
\newblock
{\BBOQ}\APACrefatitle {Commonsense reasoning and commonsense knowledge in
  artificial intelligence} {Commonsense reasoning and commonsense knowledge in
  artificial intelligence}.{\BBCQ}
\newblock
\APACjournalVolNumPages{Communications of the ACM}{58}{9}{92--103}.
\PrintBackRefs{\CurrentBib}

\bibitem [\protect \citeauthoryear {%
De~Deyne%
\ \protect \BOthers {.}}{%
De~Deyne%
\ \protect \BOthers {.}}{%
{\protect \APACyear {2008}}%
}]{%
dedeyne2008}
\APACinsertmetastar {%
dedeyne2008}%
\begin{APACrefauthors}%
De~Deyne, S.%
, Verheyen, S.%
, Ameel, E.%
, Vanpaemel, W.%
, Dry, M\BPBI J.%
, Voorspoels, W.%
\BCBL {}\ \BBA {} Storms, G.%
\end{APACrefauthors}%
\unskip\
\newblock
\APACrefYearMonthDay{2008}{}{}.
\newblock
{\BBOQ}\APACrefatitle {Exemplar by feature applicability matrices and other
  {D}utch normative data for semantic concepts} {Exemplar by feature
  applicability matrices and other {D}utch normative data for semantic
  concepts}.{\BBCQ}
\newblock
\APACjournalVolNumPages{Behavior Research Methods}{40}{4}{1030--1048}.
\PrintBackRefs{\CurrentBib}

\bibitem [\protect \citeauthoryear {%
Frank%
}{%
Frank%
}{%
{\protect \APACyear {2023}}%
}]{%
frank23}
\APACinsertmetastar {%
frank23}%
\begin{APACrefauthors}%
Frank, M\BPBI C.%
\end{APACrefauthors}%
\unskip\
\newblock
\APACrefYearMonthDay{2023}{}{}.
\newblock
{\BBOQ}\APACrefatitle {Baby steps in evaluating the capacities of large
  language models} {Baby steps in evaluating the capacities of large language
  models}.{\BBCQ}
\newblock
\APACjournalVolNumPages{Nature Reviews Psychology}{}{}{1--2}.
\PrintBackRefs{\CurrentBib}

\bibitem [\protect \citeauthoryear {%
Gelman%
}{%
Gelman%
}{%
{\protect \APACyear {2007}}%
}]{%
gelman88}
\APACinsertmetastar {%
gelman88}%
\begin{APACrefauthors}%
Gelman, S\BPBI A.%
\end{APACrefauthors}%
\unskip\
\newblock
\APACrefYearMonthDay{2007}{}{}.
\newblock
{\BBOQ}\APACrefatitle {The development of induction within natural kind and
  artifact categories} {The development of induction within natural kind and
  artifact categories}.{\BBCQ}
\newblock
\APACjournalVolNumPages{Cognitive Psychology}{20}{}{65--95}.
\PrintBackRefs{\CurrentBib}

\bibitem [\protect \citeauthoryear {%
Gelman%
\ \BBA {} Markman%
}{%
Gelman%
\ \BBA {} Markman%
}{%
{\protect \APACyear {1986}}%
}]{%
gelmanm86}
\APACinsertmetastar {%
gelmanm86}%
\begin{APACrefauthors}%
Gelman, S\BPBI A.%
\BCBT {}\ \BBA {} Markman, E\BPBI M.%
\end{APACrefauthors}%
\unskip\
\newblock
\APACrefYearMonthDay{1986}{}{}.
\newblock
{\BBOQ}\APACrefatitle {Categories and induction in young children} {Categories
  and induction in young children}.{\BBCQ}
\newblock
\APACjournalVolNumPages{Cognition}{23}{}{183--209}.
\PrintBackRefs{\CurrentBib}

\bibitem [\protect \citeauthoryear {%
Glick%
}{%
Glick%
}{%
{\protect \APACyear {2011}}%
}]{%
glick11}
\APACinsertmetastar {%
glick11}%
\begin{APACrefauthors}%
Glick, J\BPBI J\BPBI P.%
\end{APACrefauthors}%
\unskip\
\newblock
\APACrefYear{2011}.
\unskip\
\newblock
\APACrefbtitle {Uncovering the organization of semantic structure with
  similarity and inductions} {Uncovering the organization of semantic structure
  with similarity and inductions}\ \APACtypeAddressSchool {\BUPhD}{}{}.
\unskip\
\newblock
\APACaddressSchool {}{Stanford University}.
\PrintBackRefs{\CurrentBib}

\bibitem [\protect \citeauthoryear {%
Hagendorff%
}{%
Hagendorff%
}{%
{\protect \APACyear {2023}}%
}]{%
hagendorff23}
\APACinsertmetastar {%
hagendorff23}%
\begin{APACrefauthors}%
Hagendorff, T.%
\end{APACrefauthors}%
\unskip\
\newblock
\APACrefYearMonthDay{2023}{}{}.
\newblock
{\BBOQ}\APACrefatitle {Machine Psychology: Investigating Emergent Capabilities
  and Behavior in Large Language Models Using Psychological Methods} {Machine
  psychology: Investigating emergent capabilities and behavior in large
  language models using psychological methods}.{\BBCQ}
\newblock
\APACjournalVolNumPages{arXiv preprint arXiv:2303.13988}{}{}{}.
\PrintBackRefs{\CurrentBib}

\bibitem [\protect \citeauthoryear {%
Han%
, Ransom%
, Perfors%
\BCBL {}\ \BBA {} Kemp%
}{%
Han%
\ \protect \BOthers {.}}{%
{\protect \APACyear {2022}}%
}]{%
hanrpk22}
\APACinsertmetastar {%
hanrpk22}%
\begin{APACrefauthors}%
Han, S\BPBI J.%
, Ransom, K\BPBI J.%
, Perfors, A.%
\BCBL {}\ \BBA {} Kemp, C.%
\end{APACrefauthors}%
\unskip\
\newblock
\APACrefYearMonthDay{2022}{}{}.
\newblock
{\BBOQ}\APACrefatitle {Human-like property induction is a challenge for large
  language models} {Human-like property induction is a challenge for large
  language models}.{\BBCQ}
\newblock
\BIn{} \APACrefbtitle {Proceedings of the Annual Meeting of the Cognitive
  Science Society.} {Proceedings of the annual meeting of the cognitive science
  society.}
\PrintBackRefs{\CurrentBib}

\bibitem [\protect \citeauthoryear {%
Hayes%
\ \BBA {} Heit%
}{%
Hayes%
\ \BBA {} Heit%
}{%
{\protect \APACyear {2018}}%
}]{%
Hayes2018}
\APACinsertmetastar {%
Hayes2018}%
\begin{APACrefauthors}%
Hayes, B\BPBI K.%
\BCBT {}\ \BBA {} Heit, E.%
\end{APACrefauthors}%
\unskip\
\newblock
\APACrefYearMonthDay{2018}{}{}.
\newblock
{\BBOQ}\APACrefatitle {Inductive reasoning 2.0} {Inductive reasoning
  2.0}.{\BBCQ}
\newblock
\APACjournalVolNumPages{Wiley Interdisciplinary Reviews: Cognitive
  Science}{9}{3}{}.
\PrintBackRefs{\CurrentBib}

\bibitem [\protect \citeauthoryear {%
Hayes%
, Navarro%
, Stephens%
, Ransom%
\BCBL {}\ \BBA {} Dilevski%
}{%
Hayes%
\ \protect \BOthers {.}}{%
{\protect \APACyear {2019}}%
}]{%
Hayes19}
\APACinsertmetastar {%
Hayes19}%
\begin{APACrefauthors}%
Hayes, B\BPBI K.%
, Navarro, D\BPBI J.%
, Stephens, R\BPBI G.%
, Ransom, K\BPBI J.%
\BCBL {}\ \BBA {} Dilevski, N.%
\end{APACrefauthors}%
\unskip\
\newblock
\APACrefYearMonthDay{2019}{}{}.
\newblock
{\BBOQ}\APACrefatitle {The diversity effect in inductive reasoning depends on
  sampling assumptions} {The diversity effect in inductive reasoning depends on
  sampling assumptions}.{\BBCQ}
\newblock
\APACjournalVolNumPages{Psychonomic Bulletin \& Review}{26}{}{1043--1050}.
\PrintBackRefs{\CurrentBib}

\bibitem [\protect \citeauthoryear {%
Heit%
}{%
Heit%
}{%
{\protect \APACyear {1998}}%
}]{%
heit98}
\APACinsertmetastar {%
heit98}%
\begin{APACrefauthors}%
Heit, E.%
\end{APACrefauthors}%
\unskip\
\newblock
\APACrefYearMonthDay{1998}{}{}.
\newblock
{\BBOQ}\APACrefatitle {A {B}ayesian analysis of some forms of inductive
  reasoning} {A {B}ayesian analysis of some forms of inductive
  reasoning}.{\BBCQ}
\newblock
\BIn{} M.~Oaksford\ \BBA {} N.~Chater\ (\BEDS), \APACrefbtitle {Rational models
  of cognition} {Rational models of cognition}\ (\BPG~248-274).
\newblock
\APACaddressPublisher{Oxford}{Oxford University Press}.
\PrintBackRefs{\CurrentBib}

\bibitem [\protect \citeauthoryear {%
Heit%
\ \BBA {} Rubinstein%
}{%
Heit%
\ \BBA {} Rubinstein%
}{%
{\protect \APACyear {1994}}%
}]{%
heitr94}
\APACinsertmetastar {%
heitr94}%
\begin{APACrefauthors}%
Heit, E.%
\BCBT {}\ \BBA {} Rubinstein, J.%
\end{APACrefauthors}%
\unskip\
\newblock
\APACrefYearMonthDay{1994}{}{}.
\newblock
{\BBOQ}\APACrefatitle {Similarity and property effects in inductive reasoning}
  {Similarity and property effects in inductive reasoning}.{\BBCQ}
\newblock
\APACjournalVolNumPages{Journal of Experimental Psychology: Learning, Memory
  and Cognition}{20}{2}{411--422}.
\PrintBackRefs{\CurrentBib}

\bibitem [\protect \citeauthoryear {%
Holland%
, Holyoak%
, Nisbett%
\BCBL {}\ \BBA {} Thagard%
}{%
Holland%
\ \protect \BOthers {.}}{%
{\protect \APACyear {1986}}%
}]{%
hollandhnt86}
\APACinsertmetastar {%
hollandhnt86}%
\begin{APACrefauthors}%
Holland, J\BPBI H.%
, Holyoak, K\BPBI J.%
, Nisbett, R\BPBI E.%
\BCBL {}\ \BBA {} Thagard, P\BPBI R.%
\end{APACrefauthors}%
\unskip\
\newblock
\APACrefYear{1986}.
\newblock
\APACrefbtitle {Induction: {P}rocesses of inference, learning and discovery}
  {Induction: {P}rocesses of inference, learning and discovery}.
\newblock
\APACaddressPublisher{Cambridge, MA}{MIT Press}.
\PrintBackRefs{\CurrentBib}

\bibitem [\protect \citeauthoryear {%
Hu%
, Gauthier%
, Qian%
, Wilcox%
\BCBL {}\ \BBA {} Levy%
}{%
Hu%
\ \protect \BOthers {.}}{%
{\protect \APACyear {2020}}%
}]{%
hu20}
\APACinsertmetastar {%
hu20}%
\begin{APACrefauthors}%
Hu, J.%
, Gauthier, J.%
, Qian, P.%
, Wilcox, E.%
\BCBL {}\ \BBA {} Levy, R\BPBI P.%
\end{APACrefauthors}%
\unskip\
\newblock
\APACrefYearMonthDay{2020}{}{}.
\newblock
{\BBOQ}\APACrefatitle {A systematic assessment of syntactic generalization in
  neural language models} {A systematic assessment of syntactic generalization
  in neural language models}.{\BBCQ}
\newblock
\BIn{} \APACrefbtitle {Proceedings of the 58th Annual Meeting of the
  Association for Computational Linguistics} {Proceedings of the 58th annual
  meeting of the association for computational linguistics}\ (\BPG~1725-1744).
\PrintBackRefs{\CurrentBib}

\bibitem [\protect \citeauthoryear {%
Jiang%
\ \protect \BOthers {.}}{%
Jiang%
\ \protect \BOthers {.}}{%
{\protect \APACyear {2023}}%
}]{%
mewl}
\APACinsertmetastar {%
mewl}%
\begin{APACrefauthors}%
Jiang, G.%
, Xu, M.%
, Xin, S.%
, Liang, W.%
, Peng, Y.%
, Zhang, C.%
\BCBL {}\ \BBA {} Zhu, Y.%
\end{APACrefauthors}%
\unskip\
\newblock
\APACrefYearMonthDay{2023}{}{}.
\newblock
{\BBOQ}\APACrefatitle {{MEWL}: Few-shot multimodal word learning with
  referential uncertainty} {{MEWL}: Few-shot multimodal word learning with
  referential uncertainty}.{\BBCQ}
\newblock
\APACjournalVolNumPages{arXiv preprint arXiv:2306.00503}{}{}{}.
\PrintBackRefs{\CurrentBib}

\bibitem [\protect \citeauthoryear {%
Keil%
}{%
Keil%
}{%
{\protect \APACyear {1989}}%
}]{%
keil89}
\APACinsertmetastar {%
keil89}%
\begin{APACrefauthors}%
Keil, F\BPBI C.%
\end{APACrefauthors}%
\unskip\
\newblock
\APACrefYear{1989}.
\newblock
\APACrefbtitle {Concepts, kinds, and cognitive development} {Concepts, kinds,
  and cognitive development}.
\newblock
\APACaddressPublisher{Cambridge, MA}{MIT Press}.
\PrintBackRefs{\CurrentBib}

\bibitem [\protect \citeauthoryear {%
Kemp%
}{%
Kemp%
}{%
{\protect \APACyear {2011}}%
}]{%
kemp11}
\APACinsertmetastar {%
kemp11}%
\begin{APACrefauthors}%
Kemp, C.%
\end{APACrefauthors}%
\unskip\
\newblock
\APACrefYearMonthDay{2011}{}{}.
\newblock
{\BBOQ}\APACrefatitle {Inductive reasoning about chimeric creatures} {Inductive
  reasoning about chimeric creatures}.{\BBCQ}
\newblock
\BIn{} J.~Shawe-Taylor, R.~Zemel, P.~Bartlett, F\BPBI C\BPBI N.~Pereira\BCBL
  {}\ \BBA {} K\BPBI Q.~Weinberger\ (\BEDS), \APACrefbtitle {Advances in
  {N}eural {I}nformation {P}rocessing {S}ystems 24} {Advances in {N}eural
  {I}nformation {P}rocessing {S}ystems 24}\ (\BPGS\ 316--324).
\PrintBackRefs{\CurrentBib}

\bibitem [\protect \citeauthoryear {%
Kemp%
, Chang%
\BCBL {}\ \BBA {} Lombardi%
}{%
Kemp%
\ \protect \BOthers {.}}{%
{\protect \APACyear {2010}}%
}]{%
kempcl10}
\APACinsertmetastar {%
kempcl10}%
\begin{APACrefauthors}%
Kemp, C.%
, Chang, K\BPBI K.%
\BCBL {}\ \BBA {} Lombardi, L.%
\end{APACrefauthors}%
\unskip\
\newblock
\APACrefYearMonthDay{2010}{}{}.
\newblock
{\BBOQ}\APACrefatitle {Category and feature identification} {Category and
  feature identification}.{\BBCQ}
\newblock
\APACjournalVolNumPages{Acta Psychologica}{133}{}{216--233}.
\PrintBackRefs{\CurrentBib}

\bibitem [\protect \citeauthoryear {%
Kemp%
\ \BBA {} Jern%
}{%
Kemp%
\ \BBA {} Jern%
}{%
{\protect \APACyear {2014}}%
}]{%
kempj14}
\APACinsertmetastar {%
kempj14}%
\begin{APACrefauthors}%
Kemp, C.%
\BCBT {}\ \BBA {} Jern, A.%
\end{APACrefauthors}%
\unskip\
\newblock
\APACrefYearMonthDay{2014}{}{}.
\newblock
{\BBOQ}\APACrefatitle {A taxonomy of inductive problems} {A taxonomy of
  inductive problems}.{\BBCQ}
\newblock
\APACjournalVolNumPages{Psychonomic Bulletin \& Review}{21}{}{23--46}.
\PrintBackRefs{\CurrentBib}

\bibitem [\protect \citeauthoryear {%
Kemp%
, Shafto%
\BCBL {}\ \BBA {} Tenenbaum%
}{%
Kemp%
\ \protect \BOthers {.}}{%
{\protect \APACyear {2012}}%
}]{%
kempst12}
\APACinsertmetastar {%
kempst12}%
\begin{APACrefauthors}%
Kemp, C.%
, Shafto, P.%
\BCBL {}\ \BBA {} Tenenbaum, J\BPBI B.%
\end{APACrefauthors}%
\unskip\
\newblock
\APACrefYearMonthDay{2012}{}{}.
\newblock
{\BBOQ}\APACrefatitle {An integrated account of generalization across objects
  and features} {An integrated account of generalization across objects and
  features}.{\BBCQ}
\newblock
\APACjournalVolNumPages{Cognitive Psychology}{1-2}{64}{35--73}.
\PrintBackRefs{\CurrentBib}

\bibitem [\protect \citeauthoryear {%
Kemp%
\ \BBA {} Tenenbaum%
}{%
Kemp%
\ \BBA {} Tenenbaum%
}{%
{\protect \APACyear {2009}}%
}]{%
kempt09}
\APACinsertmetastar {%
kempt09}%
\begin{APACrefauthors}%
Kemp, C.%
\BCBT {}\ \BBA {} Tenenbaum, J\BPBI B.%
\end{APACrefauthors}%
\unskip\
\newblock
\APACrefYearMonthDay{2009}{}{}.
\newblock
{\BBOQ}\APACrefatitle {Structured statistical models of inductive reasoning}
  {Structured statistical models of inductive reasoning}.{\BBCQ}
\newblock
\APACjournalVolNumPages{Psychological Review}{116}{1}{20--58}.
\PrintBackRefs{\CurrentBib}

\bibitem [\protect \citeauthoryear {%
K{\i}c{\i}man%
, Ness%
, Sharma%
\BCBL {}\ \BBA {} Tan%
}{%
K{\i}c{\i}man%
\ \protect \BOthers {.}}{%
{\protect \APACyear {2023}}%
}]{%
kiciman2023causal}
\APACinsertmetastar {%
kiciman2023causal}%
\begin{APACrefauthors}%
K{\i}c{\i}man, E.%
, Ness, R.%
, Sharma, A.%
\BCBL {}\ \BBA {} Tan, C.%
\end{APACrefauthors}%
\unskip\
\newblock
\APACrefYearMonthDay{2023}{}{}.
\newblock
{\BBOQ}\APACrefatitle {Causal reasoning and large language models: Opening a
  new frontier for causality} {Causal reasoning and large language models:
  Opening a new frontier for causality}.{\BBCQ}
\newblock
\APACjournalVolNumPages{arXiv preprint arXiv:2305.00050}{}{}{}.
\PrintBackRefs{\CurrentBib}

\bibitem [\protect \citeauthoryear {%
Lake%
\ \BBA {} Murphy%
}{%
Lake%
\ \BBA {} Murphy%
}{%
{\protect \APACyear {2021}}%
}]{%
lakem21}
\APACinsertmetastar {%
lakem21}%
\begin{APACrefauthors}%
Lake, B\BPBI M.%
\BCBT {}\ \BBA {} Murphy, G\BPBI L.%
\end{APACrefauthors}%
\unskip\
\newblock
\APACrefYearMonthDay{2021}{}{}.
\newblock
{\BBOQ}\APACrefatitle {Word meaning in minds and machines.} {Word meaning in
  minds and machines.}{\BBCQ}
\newblock
\APACjournalVolNumPages{Psychological Review}{}{}{}.
\newblock
\begin{APACrefDOI} \doi{10.1037/rev0000297} \end{APACrefDOI}
\PrintBackRefs{\CurrentBib}

\bibitem [\protect \citeauthoryear {%
Landauer%
\ \BBA {} Dumais%
}{%
Landauer%
\ \BBA {} Dumais%
}{%
{\protect \APACyear {1997}}%
}]{%
landauerd97}
\APACinsertmetastar {%
landauerd97}%
\begin{APACrefauthors}%
Landauer, T\BPBI K.%
\BCBT {}\ \BBA {} Dumais, S\BPBI T.%
\end{APACrefauthors}%
\unskip\
\newblock
\APACrefYearMonthDay{1997}{}{}.
\newblock
{\BBOQ}\APACrefatitle {A solution to {P}lato's problem: {T}he {L}atent
  {S}emantic {A}nalysis theory of acquisition, induction, and representation of
  knowledge} {A solution to {P}lato's problem: {T}he {L}atent {S}emantic
  {A}nalysis theory of acquisition, induction, and representation of
  knowledge}.{\BBCQ}
\newblock
\APACjournalVolNumPages{Psychological Review}{104}{}{211-240}.
\PrintBackRefs{\CurrentBib}

\bibitem [\protect \citeauthoryear {%
Li%
, Nye%
\BCBL {}\ \BBA {} Andreas%
}{%
Li%
\ \protect \BOthers {.}}{%
{\protect \APACyear {2021}}%
}]{%
lina21}
\APACinsertmetastar {%
lina21}%
\begin{APACrefauthors}%
Li, B\BPBI Z.%
, Nye, M.%
\BCBL {}\ \BBA {} Andreas, J.%
\end{APACrefauthors}%
\unskip\
\newblock
\APACrefYearMonthDay{2021}{August}{}.
\newblock
{\BBOQ}\APACrefatitle {Implicit Representations of Meaning in Neural Language
  Models} {Implicit representations of meaning in neural language
  models}.{\BBCQ}
\newblock
\BIn{} \APACrefbtitle {Proceedings of the 59th Annual Meeting of the
  Association for Computational Linguistics and the 11th International Joint
  Conference on Natural Language Processing (Volume 1: Long Papers)}
  {Proceedings of the 59th annual meeting of the association for computational
  linguistics and the 11th international joint conference on natural language
  processing (volume 1: Long papers)}\ (\BPGS\ 1813--1827).
\newblock
\APACaddressPublisher{Online}{Association for Computational Linguistics}.
\newblock
\begin{APACrefURL} \url{https://aclanthology.org/2021.acl-long.143}
  \end{APACrefURL}
\newblock
\begin{APACrefDOI} \doi{10.18653/v1/2021.acl-long.143} \end{APACrefDOI}
\PrintBackRefs{\CurrentBib}

\bibitem [\protect \citeauthoryear {%
Lipkin%
, Wong%
, Grand%
\BCBL {}\ \BBA {} Tenenbaum%
}{%
Lipkin%
\ \protect \BOthers {.}}{%
{\protect \APACyear {2023}}%
}]{%
lipkin2023evaluating}
\APACinsertmetastar {%
lipkin2023evaluating}%
\begin{APACrefauthors}%
Lipkin, B.%
, Wong, L.%
, Grand, G.%
\BCBL {}\ \BBA {} Tenenbaum, J\BPBI B.%
\end{APACrefauthors}%
\unskip\
\newblock
\APACrefYearMonthDay{2023}{}{}.
\newblock
{\BBOQ}\APACrefatitle {Evaluating statistical language models as pragmatic
  reasoners} {Evaluating statistical language models as pragmatic
  reasoners}.{\BBCQ}
\newblock
\APACjournalVolNumPages{arXiv preprint arXiv:2305.01020}{}{}{}.
\PrintBackRefs{\CurrentBib}

\bibitem [\protect \citeauthoryear {%
L\'{o}pez%
, Atran%
, Coley%
, Medin%
\BCBL {}\ \BBA {} Smith%
}{%
L\'{o}pez%
\ \protect \BOthers {.}}{%
{\protect \APACyear {1997}}%
}]{%
lopezacms97}
\APACinsertmetastar {%
lopezacms97}%
\begin{APACrefauthors}%
L\'{o}pez, A.%
, Atran, S.%
, Coley, J\BPBI D.%
, Medin, D.%
\BCBL {}\ \BBA {} Smith, E\BPBI E.%
\end{APACrefauthors}%
\unskip\
\newblock
\APACrefYearMonthDay{1997}{}{}.
\newblock
{\BBOQ}\APACrefatitle {The tree of life: Universal and cultural features of
  folkbiological taxonomies and inductions} {The tree of life: Universal and
  cultural features of folkbiological taxonomies and inductions}.{\BBCQ}
\newblock
\APACjournalVolNumPages{Cognitive Psychology}{32}{3}{251--295}.
\PrintBackRefs{\CurrentBib}

\bibitem [\protect \citeauthoryear {%
L{\'o}pez%
, Gelman%
, Gutheil%
\BCBL {}\ \BBA {} Smith%
}{%
L{\'o}pez%
\ \protect \BOthers {.}}{%
{\protect \APACyear {1992}}%
}]{%
lopezggs92}
\APACinsertmetastar {%
lopezggs92}%
\begin{APACrefauthors}%
L{\'o}pez, A.%
, Gelman, S\BPBI A.%
, Gutheil, G.%
\BCBL {}\ \BBA {} Smith, E\BPBI E.%
\end{APACrefauthors}%
\unskip\
\newblock
\APACrefYearMonthDay{1992}{}{}.
\newblock
{\BBOQ}\APACrefatitle {The development of category-based induction} {The
  development of category-based induction}.{\BBCQ}
\newblock
\APACjournalVolNumPages{Child Development}{63}{5}{1070--1090}.
\PrintBackRefs{\CurrentBib}

\bibitem [\protect \citeauthoryear {%
Magar%
\ \BBA {} Schwartz%
}{%
Magar%
\ \BBA {} Schwartz%
}{%
{\protect \APACyear {2022}}%
}]{%
magar2022data}
\APACinsertmetastar {%
magar2022data}%
\begin{APACrefauthors}%
Magar, I.%
\BCBT {}\ \BBA {} Schwartz, R.%
\end{APACrefauthors}%
\unskip\
\newblock
\APACrefYearMonthDay{2022}{}{}.
\newblock
{\BBOQ}\APACrefatitle {Data contamination: From memorization to exploitation}
  {Data contamination: From memorization to exploitation}.{\BBCQ}
\newblock
\APACjournalVolNumPages{arXiv preprint arXiv:2203.08242}{}{}{}.
\PrintBackRefs{\CurrentBib}

\bibitem [\protect \citeauthoryear {%
Medin%
, Coley%
, Storms%
\BCBL {}\ \BBA {} Hayes%
}{%
Medin%
\ \protect \BOthers {.}}{%
{\protect \APACyear {2003}}%
}]{%
medincsh05}
\APACinsertmetastar {%
medincsh05}%
\begin{APACrefauthors}%
Medin, D\BPBI L.%
, Coley, J\BPBI D.%
, Storms, G.%
\BCBL {}\ \BBA {} Hayes, B.%
\end{APACrefauthors}%
\unskip\
\newblock
\APACrefYearMonthDay{2003}{}{}.
\newblock
{\BBOQ}\APACrefatitle {A relevance theory of induction} {A relevance theory of
  induction}.{\BBCQ}
\newblock
\APACjournalVolNumPages{Psychonomic Bulletin and Review}{10}{}{517--532}.
\PrintBackRefs{\CurrentBib}

\bibitem [\protect \citeauthoryear {%
Misra%
, Ettinger%
\BCBL {}\ \BBA {} {Taylor Rayz}%
}{%
Misra%
\ \protect \BOthers {.}}{%
{\protect \APACyear {2021}}%
}]{%
misraer21}
\APACinsertmetastar {%
misraer21}%
\begin{APACrefauthors}%
Misra, K.%
, Ettinger, A.%
\BCBL {}\ \BBA {} {Taylor Rayz}, J.%
\end{APACrefauthors}%
\unskip\
\newblock
\APACrefYearMonthDay{2021}{}{}.
\newblock
{\BBOQ}\APACrefatitle {Do language models learn typicality judgments from
  text?} {Do language models learn typicality judgments from text?}{\BBCQ}
\newblock
\BIn{} T.~Fitch, C.~Lamm, H.~Leder\BCBL {}\ \BBA {} K.~Te{\ss}mar-Raible\
  (\BEDS), \APACrefbtitle {Proceedings of the 43rd {A}nnual {M}eeting of the
  {C}ognitive {S}cience {S}ociety} {Proceedings of the 43rd {A}nnual {M}eeting
  of the {C}ognitive {S}cience {S}ociety}\ (\BPGS\ 216--222).
\PrintBackRefs{\CurrentBib}

\bibitem [\protect \citeauthoryear {%
Mitchell%
}{%
Mitchell%
}{%
{\protect \APACyear {2023}}%
}]{%
mitchell23}
\APACinsertmetastar {%
mitchell23}%
\begin{APACrefauthors}%
Mitchell, M.%
\end{APACrefauthors}%
\unskip\
\newblock
\APACrefYearMonthDay{2023}{}{}.
\newblock
{\BBOQ}\APACrefatitle {How do we know how smart {AI} systems are?} {How do we
  know how smart {AI} systems are?}{\BBCQ}
\newblock
\APACjournalVolNumPages{Science}{381}{6654}{}.
\PrintBackRefs{\CurrentBib}

\bibitem [\protect \citeauthoryear {%
Murphy%
\ \BBA {} Medin%
}{%
Murphy%
\ \BBA {} Medin%
}{%
{\protect \APACyear {1985}}%
}]{%
murphym85}
\APACinsertmetastar {%
murphym85}%
\begin{APACrefauthors}%
Murphy, G\BPBI L.%
\BCBT {}\ \BBA {} Medin, D\BPBI L.%
\end{APACrefauthors}%
\unskip\
\newblock
\APACrefYearMonthDay{1985}{}{}.
\newblock
{\BBOQ}\APACrefatitle {The role of theories in conceptual coherence} {The role
  of theories in conceptual coherence}.{\BBCQ}
\newblock
\APACjournalVolNumPages{Psychological Review}{92}{}{289-316}.
\PrintBackRefs{\CurrentBib}

\bibitem [\protect \citeauthoryear {%
Olsson%
\ \protect \BOthers {.}}{%
Olsson%
\ \protect \BOthers {.}}{%
{\protect \APACyear {2022}}%
}]{%
olsson22context}
\APACinsertmetastar {%
olsson22context}%
\begin{APACrefauthors}%
Olsson, C.%
, Elhage, N.%
, Nanda, N.%
, Joseph, N.%
, DasSarma, N.%
, Henighan, T.%
\BDBL {}others%
\end{APACrefauthors}%
\unskip\
\newblock
\APACrefYearMonthDay{2022}{}{}.
\newblock
{\BBOQ}\APACrefatitle {In-context learning and induction heads} {In-context
  learning and induction heads}.{\BBCQ}
\newblock
\APACjournalVolNumPages{arXiv preprint arXiv:2209.11895}{}{}{}.
\PrintBackRefs{\CurrentBib}

\bibitem [\protect \citeauthoryear {%
OpenAI%
}{%
OpenAI%
}{%
{\protect \APACyear {2022}}%
}]{%
gpt3embeddings}
\APACinsertmetastar {%
gpt3embeddings}%
\begin{APACrefauthors}%
OpenAI.%
\end{APACrefauthors}%
\unskip\
\newblock
\APACrefYearMonthDay{2022}{}{}.
\newblock
\APACrefbtitle {New and improved embedding model.} {New and improved embedding
  model.}
\newblock
\begin{APACrefURL}
  \url{https://openai.com/blog/new-and-improved-embedding-model}
  \end{APACrefURL}
\newblock
\APACrefnote{Accessed: April 23, 2023}
\PrintBackRefs{\CurrentBib}

\bibitem [\protect \citeauthoryear {%
OpenAI%
}{%
OpenAI%
}{%
{\protect \APACyear {2023}}%
}]{%
gpt4}
\APACinsertmetastar {%
gpt4}%
\begin{APACrefauthors}%
OpenAI.%
\end{APACrefauthors}%
\unskip\
\newblock
\APACrefYearMonthDay{2023}{}{}.
\newblock
\APACrefbtitle {{GPT-4} technical report} {{GPT-4} technical report}\
  \APACbVolEdTR{}{\BTR{}}.
\PrintBackRefs{\CurrentBib}

\bibitem [\protect \citeauthoryear {%
Osherson%
, Smith%
, Wilkie%
, Lopez%
\BCBL {}\ \BBA {} Shafir%
}{%
Osherson%
\ \protect \BOthers {.}}{%
{\protect \APACyear {1990}}%
}]{%
oshersonswls90}
\APACinsertmetastar {%
oshersonswls90}%
\begin{APACrefauthors}%
Osherson, D\BPBI N.%
, Smith, E\BPBI E.%
, Wilkie, O.%
, Lopez, A.%
\BCBL {}\ \BBA {} Shafir, E.%
\end{APACrefauthors}%
\unskip\
\newblock
\APACrefYearMonthDay{1990}{}{}.
\newblock
{\BBOQ}\APACrefatitle {Category-based induction} {Category-based
  induction}.{\BBCQ}
\newblock
\APACjournalVolNumPages{Psychological Review}{97}{2}{185--200}.
\PrintBackRefs{\CurrentBib}

\bibitem [\protect \citeauthoryear {%
Pothos%
\ \BBA {} Wills%
}{%
Pothos%
\ \BBA {} Wills%
}{%
{\protect \APACyear {2011}}%
}]{%
pothosw11}
\APACinsertmetastar {%
pothosw11}%
\begin{APACrefauthors}%
Pothos, E\BPBI M.%
\BCBT {}\ \BBA {} Wills, A\BPBI J.%
\end{APACrefauthors}%
\unskip\
\newblock
\APACrefYear{2011}.
\newblock
\APACrefbtitle {Formal approaches in categorization} {Formal approaches in
  categorization}.
\newblock
\APACaddressPublisher{}{Cambridge University Press}.
\PrintBackRefs{\CurrentBib}

\bibitem [\protect \citeauthoryear {%
Proffitt%
, Coley%
\BCBL {}\ \BBA {} Medin%
}{%
Proffitt%
\ \protect \BOthers {.}}{%
{\protect \APACyear {2000}}%
}]{%
proffittcm00}
\APACinsertmetastar {%
proffittcm00}%
\begin{APACrefauthors}%
Proffitt, J\BPBI B.%
, Coley, J\BPBI D.%
\BCBL {}\ \BBA {} Medin, D\BPBI L.%
\end{APACrefauthors}%
\unskip\
\newblock
\APACrefYearMonthDay{2000}{}{}.
\newblock
{\BBOQ}\APACrefatitle {Expertise and category-based induction.} {Expertise and
  category-based induction.}{\BBCQ}
\newblock
\APACjournalVolNumPages{Journal of Experimental Psychology: Learning, Memory,
  and Cognition}{26}{4}{811--828}.
\PrintBackRefs{\CurrentBib}

\bibitem [\protect \citeauthoryear {%
Rae%
\ \protect \BOthers {.}}{%
Rae%
\ \protect \BOthers {.}}{%
{\protect \APACyear {2021}}%
}]{%
Rae2021gopher}
\APACinsertmetastar {%
Rae2021gopher}%
\begin{APACrefauthors}%
Rae, J\BPBI W.%
, Borgeaud, S.%
, Cai, T.%
, Millican, K.%
, Hoffmann, J.%
, Song, F.%
\BDBL {}Irving, G.%
\end{APACrefauthors}%
\unskip\
\newblock
\APACrefYearMonthDay{2021}{}{}.
\newblock
{\BBOQ}\APACrefatitle {Scaling Language Models: Methods, Analysis \& Insights
  from Training {G}opher} {Scaling language models: Methods, analysis \&
  insights from training {G}opher}.{\BBCQ}
\newblock
\APACjournalVolNumPages{arXiv:2112.11446}{}{}{}.
\PrintBackRefs{\CurrentBib}

\bibitem [\protect \citeauthoryear {%
Ransom%
, Perfors%
\BCBL {}\ \BBA {} Navarro%
}{%
Ransom%
\ \protect \BOthers {.}}{%
{\protect \APACyear {2016}}%
}]{%
ransompn16}
\APACinsertmetastar {%
ransompn16}%
\begin{APACrefauthors}%
Ransom, K\BPBI J.%
, Perfors, A.%
\BCBL {}\ \BBA {} Navarro, D\BPBI J.%
\end{APACrefauthors}%
\unskip\
\newblock
\APACrefYearMonthDay{2016}{}{}.
\newblock
{\BBOQ}\APACrefatitle {Leaping to Conclusions: Why Premise Relevance Affects
  Argument Strength} {Leaping to conclusions: Why premise relevance affects
  argument strength}.{\BBCQ}
\newblock
\APACjournalVolNumPages{Cognitive Science}{40}{7}{1775--1796}.
\PrintBackRefs{\CurrentBib}

\bibitem [\protect \citeauthoryear {%
Rips%
}{%
Rips%
}{%
{\protect \APACyear {1975}}%
}]{%
rips75}
\APACinsertmetastar {%
rips75}%
\begin{APACrefauthors}%
Rips, L\BPBI J.%
\end{APACrefauthors}%
\unskip\
\newblock
\APACrefYearMonthDay{1975}{}{}.
\newblock
{\BBOQ}\APACrefatitle {Inductive judgments about natural categories} {Inductive
  judgments about natural categories}.{\BBCQ}
\newblock
\APACjournalVolNumPages{Journal of Verbal Learning and Verbal
  Behavior}{14}{}{665-681}.
\PrintBackRefs{\CurrentBib}

\bibitem [\protect \citeauthoryear {%
Rips%
}{%
Rips%
}{%
{\protect \APACyear {1989}}%
}]{%
rips89}
\APACinsertmetastar {%
rips89}%
\begin{APACrefauthors}%
Rips, L\BPBI J.%
\end{APACrefauthors}%
\unskip\
\newblock
\APACrefYearMonthDay{1989}{}{}.
\newblock
{\BBOQ}\APACrefatitle {Similarity, typicality, and categorization} {Similarity,
  typicality, and categorization}.{\BBCQ}
\newblock
\BIn{} S.~Vosniadou\ \BBA {} A.~Ortony\ (\BEDS), \APACrefbtitle {Similarity and
  analogical reasoning} {Similarity and analogical reasoning}\ (\BPG~21-59).
\newblock
\APACaddressPublisher{Cambridge}{Cambridge University Press}.
\PrintBackRefs{\CurrentBib}

\bibitem [\protect \citeauthoryear {%
Rogers%
\ \BBA {} McClelland%
}{%
Rogers%
\ \BBA {} McClelland%
}{%
{\protect \APACyear {2004}}%
}]{%
rogersm04}
\APACinsertmetastar {%
rogersm04}%
\begin{APACrefauthors}%
Rogers, T\BPBI T.%
\BCBT {}\ \BBA {} McClelland, J\BPBI L.%
\end{APACrefauthors}%
\unskip\
\newblock
\APACrefYear{2004}.
\newblock
\APACrefbtitle {Semantic cognition: A {P}arallel {D}istributed {P}rocessing
  approach} {Semantic cognition: A {P}arallel {D}istributed {P}rocessing
  approach}.
\newblock
\APACaddressPublisher{Cambridge, MA}{MIT Press}.
\PrintBackRefs{\CurrentBib}

\bibitem [\protect \citeauthoryear {%
Rogers%
\ \BBA {} McClelland%
}{%
Rogers%
\ \BBA {} McClelland%
}{%
{\protect \APACyear {2014}}%
}]{%
rogersm14}
\APACinsertmetastar {%
rogersm14}%
\begin{APACrefauthors}%
Rogers, T\BPBI T.%
\BCBT {}\ \BBA {} McClelland, J\BPBI L.%
\end{APACrefauthors}%
\unskip\
\newblock
\APACrefYearMonthDay{2014}{}{}.
\newblock
{\BBOQ}\APACrefatitle {Parallel distributed processing at 25: Further
  explorations in the microstructure of cognition} {Parallel distributed
  processing at 25: Further explorations in the microstructure of
  cognition}.{\BBCQ}
\newblock
\APACjournalVolNumPages{Cognitive Science}{38}{6}{1024--1077}.
\PrintBackRefs{\CurrentBib}

\bibitem [\protect \citeauthoryear {%
Sap%
, Rashkin%
, Chen%
, LeBras%
\BCBL {}\ \BBA {} Choi%
}{%
Sap%
\ \protect \BOthers {.}}{%
{\protect \APACyear {2019}}%
}]{%
siqa}
\APACinsertmetastar {%
siqa}%
\begin{APACrefauthors}%
Sap, M.%
, Rashkin, H.%
, Chen, D.%
, LeBras, R.%
\BCBL {}\ \BBA {} Choi, Y.%
\end{APACrefauthors}%
\unskip\
\newblock
\APACrefYearMonthDay{2019}{}{}.
\newblock
{\BBOQ}\APACrefatitle {S\textsc{ocial}{IQ}\textsc{A}: Commonsense reasoning
  about social interactions} {S\textsc{ocial}{IQ}\textsc{A}: Commonsense
  reasoning about social interactions}.{\BBCQ}
\newblock
\BIn{} \APACrefbtitle {Proc. of the 2019 {E}{M}{N}{L}{P}-{I}{J}{C}{N}{L}{P}.}
  {Proc. of the 2019 {E}{M}{N}{L}{P}-{I}{J}{C}{N}{L}{P}.}
\PrintBackRefs{\CurrentBib}

\bibitem [\protect \citeauthoryear {%
Shapira%
\ \protect \BOthers {.}}{%
Shapira%
\ \protect \BOthers {.}}{%
{\protect \APACyear {2023}}%
}]{%
shapira2023clever}
\APACinsertmetastar {%
shapira2023clever}%
\begin{APACrefauthors}%
Shapira, N.%
, Levy, M.%
, Alavi, S\BPBI H.%
, Zhou, X.%
, Choi, Y.%
, Goldberg, Y.%
\BDBL {}Shwartz, V.%
\end{APACrefauthors}%
\unskip\
\newblock
\APACrefYearMonthDay{2023}{}{}.
\newblock
{\BBOQ}\APACrefatitle {Clever {Ha}ns or neural theory of mind? {S}tress testing
  social reasoning in large language models} {Clever {Ha}ns or neural theory of
  mind? {S}tress testing social reasoning in large language models}.{\BBCQ}
\newblock
\APACjournalVolNumPages{arXiv preprint arXiv:2305.14763}{}{}{}.
\PrintBackRefs{\CurrentBib}

\bibitem [\protect \citeauthoryear {%
Shepard%
}{%
Shepard%
}{%
{\protect \APACyear {1987}}%
}]{%
shepard87}
\APACinsertmetastar {%
shepard87}%
\begin{APACrefauthors}%
Shepard, R\BPBI N.%
\end{APACrefauthors}%
\unskip\
\newblock
\APACrefYearMonthDay{1987}{}{}.
\newblock
{\BBOQ}\APACrefatitle {Towards a universal law of generalization for
  psychological science} {Towards a universal law of generalization for
  psychological science}.{\BBCQ}
\newblock
\APACjournalVolNumPages{Science}{237}{}{1317-1323}.
\PrintBackRefs{\CurrentBib}

\bibitem [\protect \citeauthoryear {%
Simon%
}{%
Simon%
}{%
{\protect \APACyear {1970}}%
}]{%
simon70}
\APACinsertmetastar {%
simon70}%
\begin{APACrefauthors}%
Simon, H\BPBI A.%
\end{APACrefauthors}%
\unskip\
\newblock
\APACrefYear{1970}.
\newblock
\APACrefbtitle {The Sciences of the Artificial} {The sciences of the
  artificial}.
\newblock
\APACaddressPublisher{}{MIT Press}.
\PrintBackRefs{\CurrentBib}

\bibitem [\protect \citeauthoryear {%
Sloman%
}{%
Sloman%
}{%
{\protect \APACyear {1993}}%
}]{%
sloman93}
\APACinsertmetastar {%
sloman93}%
\begin{APACrefauthors}%
Sloman, S\BPBI A.%
\end{APACrefauthors}%
\unskip\
\newblock
\APACrefYearMonthDay{1993}{}{}.
\newblock
{\BBOQ}\APACrefatitle {Feature-based induction} {Feature-based
  induction}.{\BBCQ}
\newblock
\APACjournalVolNumPages{Cognitive Psychology}{25}{}{231--280}.
\PrintBackRefs{\CurrentBib}

\bibitem [\protect \citeauthoryear {%
Sloman%
\ \BBA {} Lagnado%
}{%
Sloman%
\ \BBA {} Lagnado%
}{%
{\protect \APACyear {2005}}%
}]{%
slomanl05}
\APACinsertmetastar {%
slomanl05}%
\begin{APACrefauthors}%
Sloman, S\BPBI A.%
\BCBT {}\ \BBA {} Lagnado, D\BPBI A.%
\end{APACrefauthors}%
\unskip\
\newblock
\APACrefYearMonthDay{2005}{}{}.
\newblock
{\BBOQ}\APACrefatitle {The problem of induction} {The problem of
  induction}.{\BBCQ}
\newblock
\BIn{} R.~Morrison\ \BBA {} K.~Holyoak\ (\BEDS), \APACrefbtitle {Cambridge
  {H}andbook of {T}hinking and {R}easoning} {Cambridge {H}andbook of {T}hinking
  and {R}easoning}\ (\BPGS\ 95--116).
\newblock
\APACaddressPublisher{New York}{Cambridge University Press}.
\PrintBackRefs{\CurrentBib}

\bibitem [\protect \citeauthoryear {%
Smith%
, Shafir%
\BCBL {}\ \BBA {} Osherson%
}{%
Smith%
\ \protect \BOthers {.}}{%
{\protect \APACyear {1993}}%
}]{%
smithso93}
\APACinsertmetastar {%
smithso93}%
\begin{APACrefauthors}%
Smith, E\BPBI E.%
, Shafir, E.%
\BCBL {}\ \BBA {} Osherson, D.%
\end{APACrefauthors}%
\unskip\
\newblock
\APACrefYearMonthDay{1993}{}{}.
\newblock
{\BBOQ}\APACrefatitle {Similarity, plausibility, and judgments of probability}
  {Similarity, plausibility, and judgments of probability}.{\BBCQ}
\newblock
\APACjournalVolNumPages{Cognition}{49}{}{67--96}.
\PrintBackRefs{\CurrentBib}

\bibitem [\protect \citeauthoryear {%
Srivastava%
\ \protect \BOthers {.}}{%
Srivastava%
\ \protect \BOthers {.}}{%
{\protect \APACyear {2022}}%
}]{%
bigbench}
\APACinsertmetastar {%
bigbench}%
\begin{APACrefauthors}%
Srivastava, A.%
, Rastogi, A.%
, Rao, A.%
, Shoeb, A\BPBI A\BPBI M.%
, Abid, A.%
, Fisch, A.%
\BDBL {}others%
\end{APACrefauthors}%
\unskip\
\newblock
\APACrefYearMonthDay{2022}{}{}.
\newblock
{\BBOQ}\APACrefatitle {Beyond the imitation game: Quantifying and extrapolating
  the capabilities of language models} {Beyond the imitation game: Quantifying
  and extrapolating the capabilities of language models}.{\BBCQ}
\newblock
\APACjournalVolNumPages{arXiv preprint arXiv:2206.04615}{}{}{}.
\PrintBackRefs{\CurrentBib}

\bibitem [\protect \citeauthoryear {%
Storks%
, Gao%
\BCBL {}\ \BBA {} Chai%
}{%
Storks%
\ \protect \BOthers {.}}{%
{\protect \APACyear {2019}}%
}]{%
storks2019}
\APACinsertmetastar {%
storks2019}%
\begin{APACrefauthors}%
Storks, S.%
, Gao, Q.%
\BCBL {}\ \BBA {} Chai, J\BPBI Y.%
\end{APACrefauthors}%
\unskip\
\newblock
\APACrefYearMonthDay{2019}{}{}.
\newblock
{\BBOQ}\APACrefatitle {Recent advances in natural language inference: A survey
  of benchmarks, resources, and approaches} {Recent advances in natural
  language inference: A survey of benchmarks, resources, and
  approaches}.{\BBCQ}
\newblock
\APACjournalVolNumPages{arXiv preprint arXiv:1904.01172}{}{}{}.
\PrintBackRefs{\CurrentBib}

\bibitem [\protect \citeauthoryear {%
Tenenbaum%
, Griffiths%
\BCBL {}\ \BBA {} Kemp%
}{%
Tenenbaum%
\ \protect \BOthers {.}}{%
{\protect \APACyear {2006}}%
}]{%
tenenbaumgk06}
\APACinsertmetastar {%
tenenbaumgk06}%
\begin{APACrefauthors}%
Tenenbaum, J\BPBI B.%
, Griffiths, T\BPBI L.%
\BCBL {}\ \BBA {} Kemp, C.%
\end{APACrefauthors}%
\unskip\
\newblock
\APACrefYearMonthDay{2006}{}{}.
\newblock
{\BBOQ}\APACrefatitle {Theory-based {B}ayesian models of inductive learning and
  reasoning} {Theory-based {B}ayesian models of inductive learning and
  reasoning}.{\BBCQ}
\newblock
\APACjournalVolNumPages{Trends in Cognitive Science}{10}{7}{309--318}.
\PrintBackRefs{\CurrentBib}

\bibitem [\protect \citeauthoryear {%
Todd%
\ \BBA {} Gigerenzer%
}{%
Todd%
\ \BBA {} Gigerenzer%
}{%
{\protect \APACyear {2007}}%
}]{%
toddg07}
\APACinsertmetastar {%
toddg07}%
\begin{APACrefauthors}%
Todd, P\BPBI M.%
\BCBT {}\ \BBA {} Gigerenzer, G.%
\end{APACrefauthors}%
\unskip\
\newblock
\APACrefYearMonthDay{2007}{}{}.
\newblock
{\BBOQ}\APACrefatitle {Environments that make us smart: {E}cological
  rationality} {Environments that make us smart: {E}cological
  rationality}.{\BBCQ}
\newblock
\APACjournalVolNumPages{Current directions in psychological
  science}{16}{3}{167--171}.
\PrintBackRefs{\CurrentBib}

\bibitem [\protect \citeauthoryear {%
Touvron%
\ \protect \BOthers {.}}{%
Touvron%
\ \protect \BOthers {.}}{%
{\protect \APACyear {2023}}%
}]{%
llama23}
\APACinsertmetastar {%
llama23}%
\begin{APACrefauthors}%
Touvron, H.%
, Lavril, T.%
, Izacard, G.%
, Martinet, X.%
, Lachaux, M\BHBI A.%
, Lacroix, T.%
\BDBL {}others%
\end{APACrefauthors}%
\unskip\
\newblock
\APACrefYearMonthDay{2023}{}{}.
\newblock
{\BBOQ}\APACrefatitle {Llama: Open and efficient foundation language models}
  {Llama: Open and efficient foundation language models}.{\BBCQ}
\newblock
\APACjournalVolNumPages{arXiv preprint arXiv:2302.13971}{}{}{}.
\PrintBackRefs{\CurrentBib}

\bibitem [\protect \citeauthoryear {%
Ullman%
}{%
Ullman%
}{%
{\protect \APACyear {2023}}%
}]{%
ullman2023}
\APACinsertmetastar {%
ullman2023}%
\begin{APACrefauthors}%
Ullman, T.%
\end{APACrefauthors}%
\unskip\
\newblock
\APACrefYearMonthDay{2023}{}{}.
\newblock
{\BBOQ}\APACrefatitle {Large language models fail on trivial alterations to
  theory-of-mind tasks} {Large language models fail on trivial alterations to
  theory-of-mind tasks}.{\BBCQ}
\newblock
\APACjournalVolNumPages{arXiv preprint arXiv:2302.08399}{}{}{}.
\PrintBackRefs{\CurrentBib}

\bibitem [\protect \citeauthoryear {%
Voita%
, Talbot%
, Moiseev%
, Sennrich%
\BCBL {}\ \BBA {} Titov%
}{%
Voita%
\ \protect \BOthers {.}}{%
{\protect \APACyear {2019}}%
}]{%
voita19}
\APACinsertmetastar {%
voita19}%
\begin{APACrefauthors}%
Voita, E.%
, Talbot, D.%
, Moiseev, F.%
, Sennrich, R.%
\BCBL {}\ \BBA {} Titov, I.%
\end{APACrefauthors}%
\unskip\
\newblock
\APACrefYearMonthDay{2019}{{\APACmonth{07}}}{}.
\newblock
{\BBOQ}\APACrefatitle {Analyzing Multi-Head Self-Attention: Specialized Heads
  Do the Heavy Lifting, the Rest Can Be Pruned} {Analyzing multi-head
  self-attention: Specialized heads do the heavy lifting, the rest can be
  pruned}.{\BBCQ}
\newblock
\BIn{} \APACrefbtitle {Proceedings of the 57th Annual Meeting of the
  Association for Computational Linguistics} {Proceedings of the 57th annual
  meeting of the association for computational linguistics}\ (\BPGS\
  5797--5808).
\newblock
\APACaddressPublisher{Florence, Italy}{Association for Computational
  Linguistics}.
\newblock
\begin{APACrefURL} \url{https://aclanthology.org/P19-1580} \end{APACrefURL}
\newblock
\begin{APACrefDOI} \doi{10.18653/v1/P19-1580} \end{APACrefDOI}
\PrintBackRefs{\CurrentBib}

\bibitem [\protect \citeauthoryear {%
Voorspoels%
, Navarro%
, Perfors%
, Ransom%
\BCBL {}\ \BBA {} Storms%
}{%
Voorspoels%
\ \protect \BOthers {.}}{%
{\protect \APACyear {2015}}%
}]{%
voorspoelsetal15}
\APACinsertmetastar {%
voorspoelsetal15}%
\begin{APACrefauthors}%
Voorspoels, W.%
, Navarro, D.%
, Perfors, A.%
, Ransom, K.%
\BCBL {}\ \BBA {} Storms, G.%
\end{APACrefauthors}%
\unskip\
\newblock
\APACrefYearMonthDay{2015}{}{}.
\newblock
{\BBOQ}\APACrefatitle {How do people learn from negative evidence?
  {N}on-monotonic generalizations and sampling assumptions in inductive
  reasoning} {How do people learn from negative evidence? {N}on-monotonic
  generalizations and sampling assumptions in inductive reasoning}.{\BBCQ}
\newblock
\APACjournalVolNumPages{Cognitive Psychology}{81}{}{1--25}.
\PrintBackRefs{\CurrentBib}

\bibitem [\protect \citeauthoryear {%
Vosniadou%
\ \BBA {} Ortony%
}{%
Vosniadou%
\ \BBA {} Ortony%
}{%
{\protect \APACyear {1989}}%
}]{%
vosniadouo89}
\APACinsertmetastar {%
vosniadouo89}%
\begin{APACrefauthors}%
Vosniadou, S.%
\BCBT {}\ \BBA {} Ortony, A.%
\end{APACrefauthors}%
\ (\BEDS).
\unskip\
\newblock
\APACrefYear{1989}.
\newblock
\APACrefbtitle {Similarity and analogical reasoning} {Similarity and analogical
  reasoning}.
\newblock
\APACaddressPublisher{}{Cambridge University Press}.
\PrintBackRefs{\CurrentBib}

\bibitem [\protect \citeauthoryear {%
Webb%
, Holyoak%
\BCBL {}\ \BBA {} Lu%
}{%
Webb%
\ \protect \BOthers {.}}{%
{\protect \APACyear {2022}}%
}]{%
webbhl22}
\APACinsertmetastar {%
webbhl22}%
\begin{APACrefauthors}%
Webb, T.%
, Holyoak, K\BPBI J.%
\BCBL {}\ \BBA {} Lu, H.%
\end{APACrefauthors}%
\unskip\
\newblock
\APACrefYearMonthDay{2022}{}{}.
\newblock
{\BBOQ}\APACrefatitle {Emergent Analogical Reasoning in Large Language Models}
  {Emergent analogical reasoning in large language models}.{\BBCQ}
\newblock
\APACjournalVolNumPages{arXiv preprint arXiv:2212.09196}{}{}{}.
\PrintBackRefs{\CurrentBib}

\bibitem [\protect \citeauthoryear {%
Zhang%
, She%
, Gerstenberg%
\BCBL {}\ \BBA {} Rose%
}{%
Zhang%
\ \protect \BOthers {.}}{%
{\protect \APACyear {2023}}%
}]{%
zhang2023}
\APACinsertmetastar {%
zhang2023}%
\begin{APACrefauthors}%
Zhang, S.%
, She, S.%
, Gerstenberg, T.%
\BCBL {}\ \BBA {} Rose, D.%
\end{APACrefauthors}%
\unskip\
\newblock
\APACrefYearMonthDay{2023}{}{}.
\newblock
{\BBOQ}\APACrefatitle {You are what you’re for: Essentialist categorization
  in large language models} {You are what you’re for: Essentialist
  categorization in large language models}.{\BBCQ}
\newblock

\PrintBackRefs{\CurrentBib}

\end{thebibliography}

\appendix
\pagebreak
\section{Prompt variations from Experiment 1}
\label{sec:exp1prompts}

Each of the possible prompts for GPT-4 was constructed by combining a system message with variations of four components. These four components are the \textbf{Context}, \textbf{Arguments}, \textbf{Question}, and \textbf{Options}. Each variant for each component is shown below.

\begin{table}[h!]
    \centering
    \begin{tabular}{|p{0.05\linewidth}| p{0.16\linewidth} | p{0.74\linewidth}|}
    \multicolumn{3}{c}{\textbf{System:} general, task-agnostic instructions} \\
    \hline
    \small \textit{ID} & \textit{Description} & \textit{Format} \\
    \hline
    S1 & Default (blank) & \\
    \hline
    S2 & Be specific & \small{\texttt{You are an intelligent language model that responds exactly as people do. The questions you'll see don't have right or wrong answers, and you are willing to use your best judgment and commit to a concrete, specific response even in cases where you can't be sure that you are correct.
    }} \\
    \hline
    S3 & Domain expert & \small{\texttt{You are an expert on X and the types of real world properties that they have. The questions you'll see don't have right or wrong answers, and you are willing to use your best judgment and commit to a concrete, specific response even in cases where you can't be sure that you are correct.
    }}\\ 
    & & \hspace{5mm}(X = \small{\texttt{living things}} or \small{\texttt{objects}}, depending on the domain)\\
    \hline
    \end{tabular}
    \vspace{-2mm}
    \label{tbl:promptsystem}
\end{table}

\begin{table}[h!]
    \centering
    \begin{tabular}{|p{0.05\linewidth}| p{0.16\linewidth} | p{0.74\linewidth}|}
    \multicolumn{3}{c}{\textbf{Context:} explanation of the task and general task info} \\
    \hline
    \small \textit{ID} & \textit{Description} & \textit{Format} \\
    \hline
    C1 & Same as human experiment & \small{\texttt{We are interested in how people evaluate arguments. On each trial there will be two arguments labeled `A' and `B.' Each will contain one, two, or three statements separated from a claim by a line. Assume that the statements above the line are facts, and choose the argument whose facts provide a better reason for believing the claim. These are subjective judgments; there are no right or wrong answers.}} \\
    \hline
    C2 & Same as C1, but used GPT-4 to paraphrase & \small{\texttt{We aim to understand how individuals assess arguments. In every trial, you will encounter two arguments marked as `A' and `B.' Each argument may have one, two, or three statements, which are followed by a claim and separated by a line. Treat the statements above the line as factual and select the argument with facts that better support the claim. Keep in mind that these evaluations are subjective, and there are no correct or incorrect answers.}} \\
    \hline
    C3 & Sparse context & \small{\texttt{On each trial there will be two arguments labeled `A' and `B.' Each will contain one, two, or three statements separated from a claim by a line. Assume that the statements above the line are facts, and choose the argument whose facts provide a better reason for believing the claim.}} \\
    \hline
    C4 & No context & \\
    \hline
    \end{tabular}
    \vspace{-2mm}
    \label{tbl:promptcontext}
\end{table}

\begin{table}[h!]
    \centering
    \begin{tabular}{|p{0.05\linewidth}| p{0.16\linewidth} | p{0.74\linewidth}|}
\multicolumn{3}{c}{\vspace{1mm}\textbf{Arguments:} structure of the arguments being rated} \\
    \hline
    \textit{ID} & \textit{Description} & \textit{Format} \\
    \hline
A1 & Same as human & \small{\texttt{Argument A: Fact - Dogs have property P.}}\\
& experiment & \hspace{21mm} \small{\texttt{Claim - All mammals have property P.}}\\
& & \small{\texttt{Argument B: Fact - Hedgehogs have property P.}}\\
& & \hspace{21mm} \small{\texttt{Claim - All mammals have property P.}}\\
\hline
A2 & Similar to A1, & \small{\texttt{Argument A: Fact - Dogs possess property P.}}\\
& but minor & \hspace{21mm} \small{\texttt{Claim - All mammals possess property P.}}\\
& rewording & \small{\texttt{Argument B: Fact - Hedgehogs possess property P.}}\\
& & \hspace{21mm} \small{\texttt{Claim - All mammals possess property P.}}\\
        \hline
A3 & More natural & \small{\texttt{Argument A: Based on the fact that dogs have property P, we claim}} \\
& & \hspace{21mm} \small{\texttt{that all mammals have property P.}} \\
        & & \small{\texttt{Argument B: Based on the fact that hedgehogs have property P, we}} \\
& & \hspace{21mm} \small{\texttt{claim that all mammals have property P.}} \\
        \hline
    \end{tabular}
    \vspace{-2mm}
    \label{tbl:parsearguments}
\end{table}

\begin{table}[h!]
    \centering
    \begin{tabular}{|p{0.05\linewidth}| p{0.16\linewidth} | p{0.74\linewidth}|}
\multicolumn{3}{c}{\vspace{1mm}\textbf{Question:} how we ask GPT-4 to make a judgment} \\
    \hline
    \textit{ID} & \textit{Description} & \textit{Format} \\
    \hline
Q1 & Same as human experiment & \small{\texttt{Question: Assuming all the facts given are true, which argument makes a stronger case for the claim?}}\\
\hline
Q2 & Simplest  & \small{\texttt{Question: Which argument makes a stronger case for the claim?}}\\
\hline
Q3 & Prompts \hspace{5mm}reasoning & \small{\texttt{Question: Assuming all the facts given are true, which argument makes a stronger case for the claim? To get the best answer, first write down your reasoning. Then, based on this,}}\\
\hline
Q4 & Prompts \hspace{5mm}reasoning for properties & \small{\texttt{Question: Assuming all the facts given are true, which argument makes a stronger case for the claim? To get the best answer, first write down your reasoning about which real world properties `property P' might be referring to. Then, based on this,}}\\
\hline
    \end{tabular}
    \vspace{-2mm}
    \label{tbl:parsequestion}
\end{table}

\begin{table}[h!]
    \centering 
    \begin{tabular}{|p{0.05\linewidth}| p{0.16\linewidth} | p{0.74\linewidth}|}
\multicolumn{3}{c}{\vspace{1mm}\textbf{Options:} how we specify the format of the answer} \\
    \hline
    \textit{ID} & \textit{Description} & \textit{Format} \\
    \hline
O1 & Same as human experiment & \small{\texttt{Indicate the strength of your preference by providing one of the following options:}}\\
& & \small{\texttt{A - Argument A is much stronger}}\\
& & \small{\texttt{B - Argument A is stronger}}\\
& & \small{\texttt{C - Argument A is slightly stronger}}\\
& & \small{\texttt{D - Argument B is slightly stronger}}\\
& & \small{\texttt{E - Argument B is stronger}}\\
& & \small{\texttt{F - Argument B is much stronger}}\\
\hline
O2 & Scale from 0 to 100 & \small{\texttt{Indicate the strength of your preference by providing a number between 0 and 100, where 0 corresponds to argument A being much stronger and 100 corresponds to argument B being much stronger.}}\\
\hline
    \end{tabular}
    \label{tbl:promptoptions}
\end{table}

\clearpage
\section{Prompt variations from Experiment 2}
\label{sec:exp2prompts}

As in Experiment 1, each of the possible prompts for GPT-4 was constructed by combining a system message with variations of four components. The system message was the same as before and the variants for each of the other components are shown below.

\begin{table}[h!]
    \centering
    \begin{tabular}{|p{0.05\linewidth}| p{0.16\linewidth} | p{0.74\linewidth}|}
    \multicolumn{3}{c}{\textbf{Context:} explanation of the task and general task info} \\
    \hline
    \small \textit{ID} & \textit{Description} & \textit{Format} \\
    \hline
    C1 & Same as human experiment & \small{\texttt{We're going to show you a series of claims relating to living things and the properties they share. Rather than mention any specific property (e.g. "Hyenas have sesamoid bones") we'll refer to an unspecified property (e.g. "Hyenas have property P"). Each claim may be true or false, and to help you decide which, we'll provide you with facts about whether or not other living things have the same property (e.g. "Lions have property P", and "Zebras have property P").}} \\
    \hline
    C2 & Sparse context & \small{\texttt{We're going to show you a series of claims relating to living things and the properties they share. Rather than mention any specific property we'll refer to an unspecified property. Each claim may be true or false, and to help you decide which, we'll provide you with facts about whether or not other living things have the same property.}} \\
    \hline
    C3 & No context & \\
    \hline
    \end{tabular}
    \vspace{-2mm}
    \label{tbl:promptcontext2}
\end{table}

\begin{table}[h!]
    \centering
    \begin{tabular}{|p{0.05\linewidth}| p{0.16\linewidth} | p{0.74\linewidth}|}
\multicolumn{3}{c}{\vspace{1mm}\textbf{Arguments:} structure of the arguments being rated} \\
    \hline
    \textit{ID} & \textit{Description} & \textit{Format} \\
    \hline
A1 & Same as human & \small{\texttt{Argument A: Fact - Dogs have property P.}}\\
& experiment & \hspace{21mm} \small{\texttt{Claim - All mammals have property P.}}\\
\hline
A2 & Minor reword & \small{\texttt{Argument A: Fact - Dogs possess property P.}}\\
& of A1 & \hspace{21mm} \small{\texttt{Claim - All mammals possess property P.}}\\
        \hline
A3 & More natural & \small{\texttt{Argument A: Based on the fact that dogs have property P, we claim}} \\
& & \hspace{21mm} \small{\texttt{that all mammals have property P.}} \\
        \hline
    \end{tabular}
    \vspace{-2mm}
    \label{tbl:promptarguments2}
\end{table}

\begin{table}[h!]
    \centering
    \begin{tabular}{|p{0.05\linewidth}| p{0.16\linewidth} | p{0.74\linewidth}|}
\multicolumn{3}{c}{\vspace{1mm}\textbf{Question:} how we ask GPT-4 to make a judgment} \\
    \hline
    \textit{ID} & \textit{Description} & \textit{Format} \\
    \hline
Q1 & Same as human experiment & \small{\texttt{Question: Given the facts presented, how likely is it that the claim is true?}}\\
\hline
Q2 & Simplest  & \small{\texttt{Question: How likely is it that the claim is true?}}\\
\hline
Q3 & Prompts \hspace{5mm}reasoning & \small{\texttt{Question: Given the facts presented, how likely is it that the claim is true? To get the best answer, first write down your reasoning. Then, based on this,}}\\
\hline
Q4 & Prompts \hspace{5mm}reasoning for properties & \small{\texttt{Question: Given the facts presented, how likely is it that the claim is true? To get the best answer, first write down your reasoning about which real world properties ‘property P’ might be referring to. Then, based on this,}}\\
\hline
    \end{tabular}
    \vspace{-2mm}
    \label{tbl:promptquestion2}
\end{table}

\begin{table}[h!]
    \centering 
    \begin{tabular}{|p{0.05\linewidth}| p{0.16\linewidth} | p{0.74\linewidth}|}
\multicolumn{3}{c}{\vspace{1mm}\textbf{Options:} how we specify the format of the answer} \\
    \hline
    \textit{ID} & \textit{Description} & \textit{Format} \\
    \hline
O1 & Same as human experiment & \small{\texttt{Indicate your answer by providing a number between 0 and 100, where 0 means that the claim is very unlikely and 100 means that the claim is very likely.}}\\
    \hline
O2 & Six choice & \small{\texttt{Indicate your answer by providing one of the following options:}}\\
& options & \small{\texttt{A - The claim is very unlikely}}\\
& & \small{\texttt{B - The claim is moderately unlikely}}\\
& & \small{\texttt{C - The claim is slightly unlikely}}\\
& & \small{\texttt{D - The claim is slightly likely}}\\
& & \small{\texttt{E - The claim is moderately likely}}\\
& & \small{\texttt{F - The claim is very likely}}\\
\hline
    \end{tabular}
    \label{tbl:promptoptions2}
\end{table}

\begin{table}[h!]
    \centering 
    \begin{tabular}{|p{0.05\linewidth}| p{0.16\linewidth} | p{0.74\linewidth}|}
\multicolumn{3}{c}{\vspace{1mm}\textbf{Trials:} inserted practice trials after context and before main trials} \\
    \hline
    \textit{ID} & \textit{Description} & \textit{Format} \\
    \hline
T1 & Same as human experiment & \small{\texttt{This section contains a series of claims that include only one supporting fact. Before we start, we'll give you two examples as practice.}}\\
& & \\
& & \small{\texttt{Argument A: Fact - Papayas have property P.}}\\
& & \hspace{21mm} \small{\texttt{Claim - All fruits have property P.}}\\
& & \\
& & \small{\texttt{Question: Given the facts presented, how likely is it that the claim is true? Indicate your answer by providing a number between 0 and 100, where 0 means that the claim is very unlikely and 100 means that the claim is very likely.}}\\
& & \\
& & \small{\textit{\texttt{[GPT-4 completion assistant message]}}}\\
& & \\
& & \small{\texttt{Argument A: Fact - Apples have property P.}}\\
& & \hspace{21mm} \small{\texttt{Claim - All fruits have property P.}}\\
& & \\
& & \small{\texttt{Question: Given the facts presented, how likely is it that the claim is true? Indicate your answer by providing a number between 0 and 100, where 0 means that the claim is very unlikely and 100 means that the claim is very likely.}}\\
& & \\
& & \small{\textit{\texttt{[GPT-4 completion assistant message]}}}\\
    \hline
T2 & No trials & \\
\hline
    \end{tabular}
    \label{tbl:prompttrials}
\end{table}

\clearpage
\section{Property Induction Phenomena from Experiment 1}
\label{sec:exp1phenomena}

Table~\ref{tab:exp1responses} shows responses for all of the argument pairs listed in Table~\ref{tbl:phenomena}. The prompts used the same format shown in Table~\ref{tbl:expt1bestprompt}, and Table~\ref{tab:exp1responses} shows the first result obtained for each prompt (we did not query GPT-3.5 and GPT-4 multiple times and select among the responses).

Because there is a single prompt for each phenomenon and Table~\ref{tab:exp1responses} includes only a single response to each prompt, the results in this table and in the appendices that follow are little more than anecdotal. Even so they suggest that GPT-4 often explains its choices in ways that seem sensible and invoke some of the key notions (e.g.\ similarity (row 7), typicality (row 2) and diversity (row 6) considered in the psychological literature on property induction. The explanations offered by GPT-4 are also usually more convincing than those provided by GPT-3.5.

Because non-monotonicity is the most important phenomenon not consistently captured by GPT-4, the responses for the non-monotonicity pairs are of particular interest. GPT-4's response in row 9 suggests that additional positive examples provide additional support for the conclusion, which is an appropriate explanation of \emph{monotonicity}, but does not capture how people tend to think about \emph{non-monotonicity}. GPT-4's response in row 8 actually shows a non-monotonicity effect, but the explanation provided is unconvincing --- according to that explanation both arguments in the pair should be equally strong. 

\singlespacing
\begin{longtable}{p{.24\textwidth}  p{.35\textwidth}  p{.35\textwidth} }
\textbf{Phenomenon} & \textbf{GPT-4} & \textbf{GPT-3.5} \\
\hline
1) \phen{Similarity} \newline \newline
\textit{Argument A:} \newline 
Robins\newline
Bluejays\newline
\vspace{2mm} $\rightarrow$ Sparrows\newline \newline
\textit{Argument B:} \newline
Robins\newline
Bluejays\newline
\vspace{2mm} $\rightarrow$ Geese\newline
 &
\small{\texttt{Robins, Bluejays, and Sparrows are all small birds that belong to the passerine order. Geese, on the other hand, are larger birds that belong to the Anseriformes order. Since the facts provided in both arguments are the same, we can assume that property P might be more likely to be shared among birds that are more closely related. 
Based on this reasoning, I would choose option B - Argument A is moderately stronger.}}
&
\small{\texttt{Answer: B - Argument A is moderately stronger.\newline
Reasoning: Argument A provides two facts that support the claim that sparrows have property P, which is more evidence than Argument B provides for the claim that geese have property P. Therefore, Argument A makes a moderately stronger case for the claim.}}\\
\hline
2) \phen{Typicality} \newline  \newline
\textit{Argument A:} \newline 
Robins\newline
\vspace{2mm} $\rightarrow$ All birds\newline \newline
\textit{Argument B:} \newline
Penguins\newline
\vspace{2mm} $\rightarrow$ All birds\newline
&
\small{\texttt{Robins are more representative of the typical bird, as they can fly and have many features that are commonly associated with birds. Penguins, on the other hand, are flightless and have some unique features that set them apart from other birds. Therefore, if a property is found in robins, it is more likely to be found in other birds as well, compared to if the property is found in penguins.
Based on this reasoning, I would choose option B - Argument A is moderately stronger.}} &
\small{\texttt{C - Argument A is slightly stronger.\newline
The fact that robins have property P provides some evidence that all birds have property P, but the fact that penguins have property P provides stronger evidence that all birds have property P. Therefore, Argument B makes a slightly stronger case for the claim than Argument A.}} \\
\hline
3) \phen{Specificity} \newline  \newline
\textit{Argument A:} \newline 
Bluejays\newline
Falcons\newline
\vspace{2mm} $\rightarrow$ All birds\newline \newline
\textit{Argument B:} \newline
Bluejays\newline
Falcons\newline
\vspace{2mm} $\rightarrow$ All animals\newline
&
\small{\texttt{Both arguments are based on the same facts, which are that Bluejays and Falcons have property P. Bluejays and Falcons are both types of birds, so it is more reasonable to generalize the property P to all birds rather than all animals. Generalizing to all animals would require more diverse examples of animals having property P.
Based on this reasoning, I would choose option B - Argument A is moderately stronger.}}&
\small{\texttt{Answer: B - Argument A is moderately stronger.\newline
Reasoning: Argument A makes a stronger case for the claim because it is more specific and limited in scope. It states that all birds have property P, which is supported by the facts given. Argument B states that all animals have property P, which is not supported by the facts given since only birds have been mentioned. Therefore, Argument A is moderately stronger.}}\\
\hline
4) \phen{Monotonicity \newline (General)} \newline  \newline
\textit{Argument A:} \newline 
Sparrows\newline
Eagles\newline
Hawks\newline
\vspace{2mm} $\rightarrow$ All birds\newline \newline
\textit{Argument B:} \newline
Sparrows\newline
Eagles\newline
\vspace{2mm} $\rightarrow$ All birds
& \small{\texttt{Argument A provides three examples of birds with property P, while Argument B provides only two examples. Having more examples of birds with property P strengthens the case for the claim that all birds have property P.
Based on this reasoning, I would choose option B - Argument A is moderately stronger.}} &
\small{\texttt{Answer: B - Argument A is moderately stronger.\newline
Reasoning: Argument A provides more evidence for the claim that all birds have property P, as it includes three facts that support the claim. Argument B only includes two facts, which is not as strong of a case for the claim. Therefore, Argument A is moderately stronger than Argument B.}} \\
\hline
5) \phen{Monotonicity \newline (Specific)} \newline  \newline
\textit{Argument A:} \newline 
Pigs\newline
Wolves\newline
Foxes\newline
\vspace{2mm} $\rightarrow$ Gorillas\newline \newline
\textit{Argument B:} \newline
Pigs\newline
Wolves\newline
\vspace{2mm} $\rightarrow$ Gorillas
& \small{\texttt{Argument A provides more evidence for the claim by including an additional animal (foxes) that has property P. This additional evidence strengthens the argument by showing that property P is more widespread among animals, making it more likely that gorillas might also have property P.
Based on this reasoning, I would choose option B - Argument A is moderately stronger.}} &
\small{\texttt{Answer: B - Argument A is moderately stronger.\newline
Reasoning: Argument A provides more evidence for the claim that gorillas have property P, as it includes three facts that all support the claim. Argument B only includes two facts, so it provides less evidence for the claim. Therefore, Argument A is moderately stronger than Argument B.}} \\
\hline
6) \phen{Diversity \newline (General)} \newline  \newline
\textit{Argument A:} \newline 
Hippos\newline
Hamsters\newline
\vspace{2mm} $\rightarrow$ All mammals\newline \newline
\textit{Argument B:} \newline
Hippos\newline
Rhinos\newline
\vspace{2mm} $\rightarrow$ All mammals
& \small{\texttt{Argument A provides evidence that both a large aquatic mammal (hippo) and a small terrestrial mammal (hamster) have property P. Argument B provides evidence that two large, semi-aquatic mammals (hippo and rhino) have property P. Since Argument A includes a wider range of mammals in terms of size and habitat, it provides a slightly better case for the claim that all mammals have property P. 
Answer: C - Argument A is slightly stronger}} &
\small{\texttt{Answer: B - Argument A is moderately stronger. \newline
Reasoning: Argument A provides two examples of mammals that have property P, while Argument B provides only one. This suggests that Argument A provides a better case for the claim that all mammals have property P.}
}\\
\hline
7) \phen{Diversity \newline (Specific)} \newline  \newline
\textit{Argument A:} \newline 
Lions\newline
Giraffes\newline
\vspace{2mm} $\rightarrow$ Rabbits\newline \newline
\textit{Argument B:} \newline
Lions\newline
Tigers\newline
\vspace{2mm} $\rightarrow$ Rabbits\newline
& \small{\texttt{In Argument A, we have two animals with property P - lions and giraffes. Lions are carnivorous mammals, while giraffes are herbivorous mammals. In Argument B, we have lions and tigers with property P, both of which are carnivorous mammals. Since rabbits are herbivorous mammals, they share a more similar trait with giraffes than with lions or tigers. Therefore, the presence of an herbivorous mammal (giraffe) in Argument A makes it slightly more likely that rabbits might also have property P.
Answer: C - Argument A is slightly stronger}} &
\small{\texttt{Answer: B - Argument A is moderately stronger.\newline
Reasoning: Argument A provides two facts that are both relevant to the claim, which makes it a moderately stronger argument than Argument B, which only provides one relevant fact.}} \\
\hline
8) \phen{Non-monotonicity \newline (General)} \newline  \newline
\textit{Argument A:} \newline 
Crows\newline
Peacocks\newline
\vspace{2mm} $\rightarrow$ All birds\newline \newline
\textit{Argument B:} \newline
Crows\newline
Peacocks\newline
Rabbits\newline
\vspace{2mm} $\rightarrow$ All birds\newline
& \small{\texttt{Both arguments A and B have the same claim that all birds have property P. Argument A provides two examples of birds (crows and peacocks) having property P, while Argument B provides the same two examples of birds plus an additional example of a non-bird (rabbits) having property P. The inclusion of rabbits in Argument B does not strengthen the claim about birds, as rabbits are not birds. Therefore, the additional fact in Argument B does not contribute to the claim about birds.
Based on this reasoning, I would choose option C - Argument A is slightly stronger. This is because both arguments have the same bird examples, but Argument B includes an irrelevant fact about rabbits that does not contribute to the claim about birds.}} &
\small{\texttt{Answer: E - Argument B is moderately stronger.\newline
Reasoning: Argument B provides more evidence for the claim that all birds have property P by providing an additional fact that rabbits have property P. This additional fact strengthens the argument that all birds have property P, making Argument B moderately stronger than Argument A}} \\
\hline
9) \phen{Non-monotonicity \newline (Specific)} \newline  \newline
\textit{Argument A:} \newline 
Flies\newline
\vspace{2mm} $\rightarrow$ Bees\newline \newline
\textit{Argument B:} \newline
Flies\newline
Orangutans\newline
\vspace{2mm} $\rightarrow$ Bees\newline
& \small{\texttt{Argument A only provides one example of an animal (flies) having property P. Argument B, on the other hand, provides two examples of animals (flies and orangutans) having property P. Although neither flies nor orangutans are closely related to bees, the fact that two different animals have property P makes it more likely that bees might also have property P.
Based on this reasoning, my preference is:
E - Argument B is moderately stronger}} &
\small{\texttt{Answer: B - Argument A is moderately stronger.\newline
Reasoning: Argument A provides one fact that supports the claim that bees have property P, while Argument B provides two facts that support the claim. Therefore, Argument B makes a stronger case for the claim that bees have property P.}}\\
\hline
10) \phen{Asymmetry} \newline  \newline
\textit{Argument A:} \newline 
Mice\newline
\vspace{2mm} $\rightarrow$ Bats\newline \newline
\textit{Argument B:} \newline
Bats\newline
\vspace{2mm} $\rightarrow$ Mice\newline
& \small{\texttt{Mice and bats are both mammals, but they belong to different orders (Rodentia and Chiroptera, respectively). While they share some common features, they also have many differences. In this case, it is not clear what property P is, so it is difficult to determine how likely it is that one species having property P would imply the other species also has it.
However, bats are more closely related to other mammals like primates and carnivores than they are to rodents like mice. This means that if a property is found in bats, it might be more likely to be found in other mammals, including mice. On the other hand, if a property is found in mice, it might be less likely to be found in more distantly related mammals like bats.
Based on this reasoning, I would choose option E - Argument B is moderately stronger.}} &
\small{\texttt{Answer: F - Argument B is much stronger.\newline 
The fact given in Argument B directly supports the claim, while the fact given in Argument A does not. Therefore, Argument B makes a much stronger case for the claim than Argument A.}}\\
\hline
11) \phen{Inclusion Fallacy} \newline  \newline
\textit{Argument A:} \newline 
Robins\newline
\vspace{2mm} $\rightarrow$ All birds\newline \newline
\textit{Argument B:} \newline
Robins\newline
\vspace{2mm} $\rightarrow$ Ostriches\newline
& \small{\texttt{Argument A makes a general claim about all birds, while Argument B makes a specific claim about ostriches. Since robins and ostriches are both birds, knowing that robins have property P provides some evidence that ostriches might also have property P. However, it does not provide strong evidence that all birds have property P, as there are many different types of birds with varying properties.
Based on this reasoning, I would choose option D - Argument B is slightly stronger.}} &
\small{\texttt{Answer: B - Argument A is moderately stronger.\newline
Reasoning: Argument A makes a stronger case for the claim because it states that all birds have property P, which is supported by the fact that robins have property P. Argument B states that ostriches have property P, but does not provide any evidence to support this claim. Therefore, Argument A is moderately stronger.}} \\
\hline
\label{tab:exp1responses}\\
\caption{\small{Responses for argument pairs illustrating the eleven inductive phenomena in Experiment 1. In all cases argument A is stronger according to data reported by \cite{oshersonswls90}.}} 
\end{longtable}
\doublespacing

\section{Results for Other OpenAI Models}
\label{sec:otheropenaimodels}

We began this project when GPT-3 (\texttt{davinci}) was the best performing OpenAI model available, and updated our work as subsequent versions of GPT-3, GPT-3.5 and GPT-4 were released. Our primary analysis focuses on GPT-3.5 (\texttt{text-davinci-003}) and GPT-4 because they are the best performing completion and chat based OpenAI models respectively, but here we provide results for all models that we analysed.

For Experiment 1, Table~\ref{tbl:expt1stats_appendix} provides a quantitative evaluation of GPT-3.5's chat variant (\texttt{gpt-3.5-turbo-0613}) alongside the other models that we report in Table~\ref{tbl:expt1stats} for the argument pair rating task. Although we ran this analysis on other GPT-3 and GPT-3.5 variants including \texttt{davinci}, \texttt{text-davinci-001} and \texttt{text-davinci-002}, we found that these models largely failed to respond appropriately to our prompt, which may not be surprising because this prompt was optimised for GPT-4.

Like GPT-4, GPT-3.5's chat variant captures the phenomena of specificity and monotonicity, and it also leans towards predicting the opposite argument from humans for non-monotonicity.  GPT-3.5's chat variant, however, fails to capture similarity and typicality as robustly as does GPT-4, and also displays more sensitivity towards the inclusion fallacy.

\begin{figure}[t!]
\begin{center}
\includegraphics[width=\textwidth]{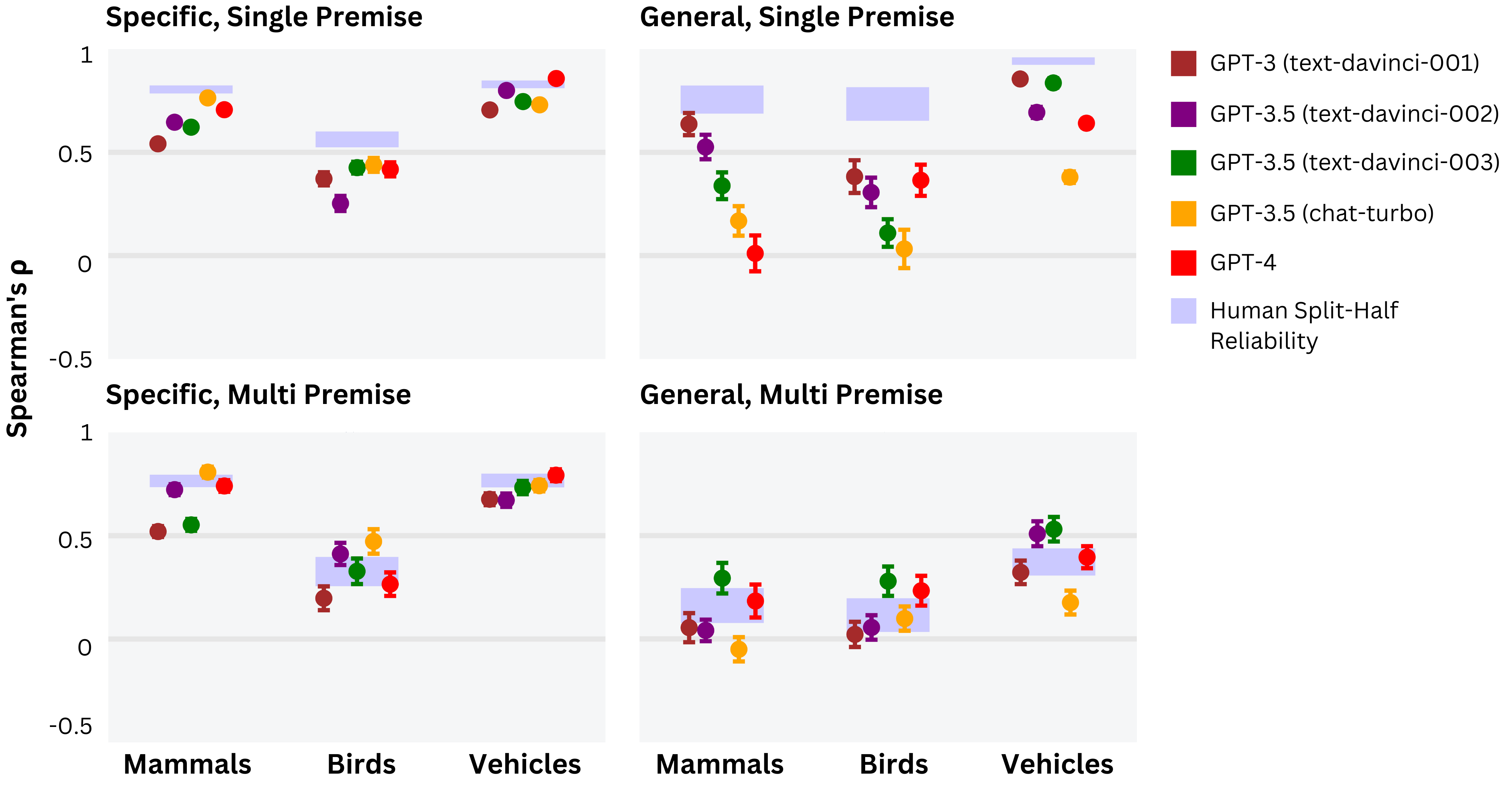}
\end{center}
\vspace{-1mm}
\caption{Correlations (Spearman's $\rho$) between human argument rankings and rankings of GPT-3 (text-davinci-001), GPT-3.5 (text-davinci-002, text-davinci-003 and chat-turbo), and GPT-4. } 
\label{fig:argrating_appendix}
\vspace{-2mm}
\end{figure}

For Experiment 2, Figure~\ref{fig:argrating_appendix} displays model correlations with humans for our rating task across five models: \texttt{text-davinci-001}, \texttt{text-davinci-002}, \texttt{text-davinci-003}, \texttt{gpt-3.5-turbo} and GPT-4. Like our analysis for Experiment 1, we also attempted to gather results for \texttt{davinci}, but found that our GPT-4 optimised prompt did not work on models that lacked instruction tuning. 

In general, we found that the differences between the GPT models are relatively small. The most striking difference is the degradation that can be observed across the models for single premise general arguments. We believe that this finding may reflect recent alignment efforts to prevent models from making broad generalisations (`Claim - \textit{All} X have property P') based on user prompts that contain limited information.

\begin{table}[t!]
\centering
\caption{\small{Quantitative evaluation of GPT-3.5 (text-davinci-003), GPT-3.5 (chat-turbo), GPT-4 and Humans on the 11 phenomena across all three domains.}}
\vspace{1mm}
\begin{tabular}{|p{0.2\linewidth}| p{0.15\linewidth} | p{0.12\linewidth} p{0.12\linewidth} p{0.12\linewidth} p{0.12\linewidth} |}
\hline
\textbf{Phenomenon} & \textbf{Domain} & \gptthreefivethree{\textbf{GPT-3.5 (text-davinci-003)}} & \gptthreefivechat{\textbf{GPT-3.5 (chat-turbo)}} & \gptfour{\textbf{GPT-4}} & \human{\textbf{Humans}}\\ \hline
Similarity & Mammals & 0.17 & 0.17 & \gptfour{\textbf{$<$0.001 *}} & \human{\textbf{$<$0.001 *}} \\
  & Birds & 0.69 & 0.4 & \gptfour{\textbf{$<$0.001 *}} & \human{\textbf{$<$0.001 *}} \\
  & Vehicles & 0.11 & \gptthreefivechat{\textbf{$<$0.001 *}} & \gptfour{\textbf{$<$0.001 *}} & \human{\textbf{$<$0.001 *}} \\
\hline
Typicality & Mammals & 0.54 & 0.29 & \gptfour{\textbf{$<$0.02 *}} & \human{\textbf{$<$0.001 *}} \\
  & Birds & 0.54 & 0.15 & \gptfour{\textbf{$<$0.001 *}} & \human{\textbf{$<$0.001 *}} \\
  & Vehicles & 1.0 & \gptthreefivechat{\textbf{$<$0.02 *}} & \gptfour{\textbf{$<$0.001 *}} & \human{\textbf{$<$0.001 *}} \\
\hline
Specificity & Mammals & 0.84 & \gptthreefivechat{\textbf{$<$0.001 *}} & \gptfour{\textbf{$<$0.001 *}} & \human{\textbf{$<$0.001 *}} \\
  & Birds & 0.84 & \gptthreefivechat{\textbf{$<$0.001 *}} & \gptfour{\textbf{$<$0.001 *}} & \human{\textbf{$<$0.001 *}} \\
  & Vehicles & 0.54 & \gptthreefivechat{\textbf{$<$0.001 *}} & \gptfour{\textbf{$<$0.001 *}} & \human{\textbf{$<$0.001 *}} \\
\hline
Monotonicity & Mammals & \gptthreefivethree{\textbf{$<$0.02 *}} & \gptthreefivechat{\textbf{$<$0.001 *}} & \gptfour{\textbf{$<$0.001 *}} & \human{\textbf{$<$0.001 *}} \\
(General) & Birds & 0.54 & \gptthreefivechat{\textbf{$<$0.001 *}} & \gptfour{\textbf{$<$0.001 *}} & \human{\textbf{$<$0.001 *}} \\
  & Vehicles & \gptthreefivethree{\textbf{$<$0.001 *}} & \gptthreefivechat{\textbf{$<$0.001 *}} & \gptfour{\textbf{$<$0.001 *}} & \human{\textbf{$<$0.001 *}} \\
\hline
Monotonicity & Mammals & 0.31 & \gptthreefivechat{\textbf{$<$0.001 *}} & \gptfour{\textbf{$<$0.001 *}} & 0.06 \\
(Specific) & Birds & 0.84 & \gptthreefivechat{\textbf{$<$0.001 *}} & \gptfour{\textbf{$<$0.001 *}} & \human{\textbf{$<$0.001 *}} \\
  & Vehicles & 0.84 & \gptthreefivechat{\textbf{$<$0.001 *}} & \gptfour{\textbf{$<$0.001 *}} & 0.29 \\
\hline
Diversity & Mammals & 0.84 & 0.31 & \gptfour{\textbf{$<$0.001 *}} & 0.06 \\
(General) & Birds & 0.54 & 0.15 & \gptfour{\textbf{$<$0.001 *}} & 1.0 \\
  & Vehicles & 1.0 & 0.84 & 0.06 & \human{\textbf{$<$0.03 $\circ$}} \\
\hline
Diversity & Mammals & 1.0 & 1.0 & 0.15 & \human{\textbf{$<$0.01 $\circ$}} \\
(Specific) & Birds & 0.54 & 0.68 & 0.84 & 0.68 \\
  & Vehicles & 0.31 & 0.4 & 0.15 & 0.68 \\
\hline
Nonmonotonicity & Mammals & \gptthreefivethree{\textbf{$<$0.001 $\circ$}} & 1.0 & \gptfour{\textbf{$<$0.02 *}} & \human{\textbf{$<$0.001 *}} \\
(General) & Birds & 0.31 & 0.06 & \gptfour{\textbf{$<$0.001 *}} & \human{\textbf{$<$0.001 *}} \\
  & Vehicles & 0.84 & \gptthreefivechat{\textbf{$<$0.02 $\circ$}} & 0.15 & \human{\textbf{$<$0.01 *}} \\
\hline
Nonmonotonicity & Mammals & 1.0 & \gptthreefivechat{\textbf{$<$0.001 $\circ$}} & \gptfour{\textbf{$<$0.001 $\circ$}} & 0.15 \\
(Specific) & Birds & 0.15 & \gptthreefivechat{\textbf{$<$0.001 $\circ$}} & \gptfour{\textbf{$<$0.001 $\circ$}} & \human{\textbf{$<$0.001 *}} \\
  & Vehicles & 1.0 & \gptthreefivechat{\textbf{$<$0.001 $\circ$}} & \gptfour{\textbf{$<$0.001 $\circ$}} & \human{\textbf{$<$0.01 *}} \\
\hline
Asymmetry & Mammals & 0.06 & 0.82 & 0.68 & 0.4 \\
  & Birds & 0.06 & 0.84 & 1.0 & \human{\textbf{$<$0.001 *}} \\
  & Vehicles & 0.54 & 0.09 & 0.82 & 0.05 \\
\hline
Inclusion & Mammals & 0.84 & \gptthreefivechat{\textbf{$<$0.001 *}} & 0.06 & 0.06 \\
Fallacy & Birds & 0.84 & \gptthreefivechat{\textbf{$<$0.02 *}} & \gptfour{\textbf{$<$0.001 $\circ$}} & 1.0 \\
  & Vehicles & 0.54 & 0.06 & 0.54 & \human{\textbf{$<$0.001 *}} \\
\hline
\end{tabular}
\label{tbl:expt1stats_appendix}
\vspace{-3mm}
\end{table}

\section{Other Property Induction Phenomena}
\label{sec:otherpiphenomena}

Table~\ref{tab:otherpiresponses} illustrates a range of property induction phenomena from the literature that were not explored in our experiments but can be systematically explored in future work. As in Appendix~\ref{sec:inductivephenomena}, the prompts matched the format shown in Table~\ref{tbl:expt1bestprompt}, and Table~\ref{tab:otherpiresponses} shows the first model response we obtained for each argument pair.

The first four rows of the table illustrate phenomena described by \cite{medincsh05} that involve inferences about blank properties. GPT-4 provides relatively compelling responses to the first two argument-pairs, but the explanation given for the third pair is unconvincing (birds and mammals are both vertebrates so it is not clear that birds are more closely related to mammals than to insects). The fourth pair illustrates that non-monotonicity can arise for reasons that differ from those captured by our non-monotonicity stimuli in Experiment 1. In that experiment non-monotonicity was achieved by adding a premise drawn from a different superordinate category, but the fourth row shows that non-monotonicity can also arise when all premises share a very salient feature (e.g.\ the feature of being bears) which makes it less likely that ``Property P'' extends outside the set of items with that feature. GPT-4's response to the fourth pair suggests that it may struggle with non-monotonicity effects in general, not just with the variety of non-monotonicity explored in Experiment 1.

The final six rows in Table~\ref{tab:otherpiresponses} all involve arguments with non-blank predicates. They include four cases (rows 5,6,8 and 10) in which the specific predicate used overturns inferences that would be expected on the basis of overall similarity between the premise and conclusion categories. For example, german shepherds are more similar to dobermanns than to poodles, but learning that poodles ``bite through wire'' provides more evidence that german shepherds have the same property than learning that dobermanns bite through wire. Similarly, mosquitoes are more similar to ladybugs than to vampire bats, but learning that vampire bats ``travel at speeds of twice their body length per second'' after eating provides more evidence that mosquitoes have the same property than learning that ladybugs have the property. Across all four cases, GPT-4 gives responses that are consistent with overall similarity between premise and conclusion categories rather than the specific properties used in the arguments, suggesting that the model provides a relatively poor account of inferences about non-blank properties. In contrast, people appear to rely on similarity only when reasoning about properties (e.g.\ rows 7 and 9) that align well with overall similarity relationships. 

\singlespacing
\begin{longtable}{ p{.30\textwidth}  p{.35\textwidth}  p{.35\textwidth} }
\textbf{Phenomenon} & \textbf{GPT-4} & \textbf{GPT-3.5} \\
1) \phen{Causal asymmetry} \newline
\citep{medincsh05} \newline  
\texttt{
Argument A: \newline
F - Carrots have property P.\newline
C - Rabbits have property P.\newline
Argument B:\newline
F - Rabbits have property P.\newline
C - Carrots have property P.\newline
}
  &
\texttt{Carrots are a type of vegetable, while rabbits are animals. It is more likely that a property found in a vegetable could be present in an animal that consumes it, rather than a property found in an animal being present in the vegetable it consumes. This is because animals can obtain certain properties from their diet, while vegetables do not obtain properties from the animals that consume them.\newline
Based on this reasoning, I would choose option B - Argument A is moderately stronger
}&
\texttt{Answer: E - Argument B is moderately stronger.\newline
Reasoning: Argument B provides direct evidence that rabbits have property P, while Argument A does not provide any direct evidence that rabbits have property P. Therefore, Argument B makes a stronger case for the claim that carrots have property P
} \\
2) Causal violations of similarity \newline
\citep{medincsh05} \newline
\texttt{
Argument A: \newline
F - Bananas have property P.\newline
C - Monkeys have property P.\newline
Argument B:\newline
F - Mice have property P.\newline
C - Monkeys have property P.\newline
}
   &
\texttt{Monkeys are more closely related to bananas in terms of their diet and habitat, as they are known to consume bananas and live in regions where bananas grow. Mice, on the other hand, have a more distant relationship with monkeys in terms of their diet and habitat. Therefore, it is more likely that a property found in bananas could also be found in monkeys, as opposed to a property found in mice.\newline 
Based on this reasoning, I would choose option B - Argument A is moderately stronger.}
&
\texttt{Answer: B - Argument A is moderately stronger.\newline  
Reasoning: Argument A provides a direct connection between the fact and the claim, as bananas are a type of fruit that monkeys eat. Argument B provides an indirect connection, as mice are not typically associated with monkeys. Therefore, Argument A provides a moderately stronger case for the claim that monkeys have property P.}\\
3) \phen{Causal nondiversity} \newline
\citep{medincsh05} \newline
\texttt{
Argument A: \newline
F - Fleas have property P.\newline
F - Butterflies have property P.\newline
C - Sparrows have property P.\newline
Argument B:\newline
F - Fleas have property P.\newline
F - Dogs have property P.\newline
C - Sparrows have property P.\newline
}
   &
\texttt{Fleas and butterflies are both insects, while dogs are mammals. Sparrows are birds, which are more closely related to insects than mammals in terms of taxonomy. Therefore, the fact that both fleas and butterflies have property P might be more relevant to the claim that sparrows have property P than the fact that fleas and dogs have property P.\newline 
Based on this reasoning, I would choose option B - Argument A is moderately stronger.}&
\texttt{Answer: B - Argument A is moderately stronger.\newline
Reasoning: Argument A provides two facts that are both relevant to the claim, which makes it a moderately stronger argument than Argument B, which only provides one relevant fact.}\\
4) \phen{Non-monotonicity via property reinforcement} \newline
\citep{medincsh05} \newline
\texttt{
Argument A: \newline
F - Brown bears have property P.\newline
C - Buffalos have property P.\newline
Argument B:\newline
F - Brown bears have property P.\newline
F - Polar bears have property P.\newline
F - Grizzly bears have property P.\newline
C - Buffalos have property P.\newline
}
   &
\texttt{Argument A provides only one fact, that brown bears have property P. Argument B provides three facts, that brown bears, polar bears, and grizzly bears all have property P. Although none of these animals are closely related to buffalos, Argument B provides more evidence of property P being present in a wider range of animals, which could potentially increase the likelihood that buffalos also have property P.\newline
Based on this reasoning, I would choose option D - Argument B is slightly stronger
} &
\texttt{Answer: E - Argument B is moderately stronger.\newline
Reasoning: Argument B provides more evidence for the claim that buffalos have property P by providing three facts about other animals that have property P. This makes the argument stronger than Argument A, which only provides one fact.}\\   

5) \phen{Property effects} \newline
\citep{smithso93} \newline
\texttt{
Argument A: \newline
F - Poodles can bite through wire.\newline
C - German shepherds can bite through wire.\newline
Argument B:\newline
F - Dobermanns can bite through wire.\newline
C - German shepherds can bite through wire.\newline
}
   & 
\texttt{Both Poodles and Dobermanns are different breeds of dogs, just like German Shepherds. While Poodles are known for their intelligence and agility, Dobermanns are known for their strength and guarding abilities. German Shepherds are also known for their strength, intelligence, and guarding abilities. Since Dobermanns and German Shepherds share more similar characteristics in terms of strength and function, it is more likely that if Dobermanns can bite through wire, German Shepherds can as well.\newline
Preference: E - Argument B is moderately stronger}  &
\texttt{Answer: E - Argument B is moderately stronger.\newline
Reasoning: Argument B provides a more direct connection between the fact and the claim, as it states that a Dobermann, which is a type of German shepherd, can bite through wire. This is a more specific example than the fact provided in Argument A, which only states that poodles can bite through wire. Therefore, Argument B makes a moderately stronger case for the claim that German shepherds can bite through wire.} \\

6) \phen{Property effects} \newline
\citep{smithso93} \newline
\texttt{
Argument A: \newline
F - House cats have skins that are more resistant to penetration than most synthetic fibers.\newline
C - Hippos have skins that are more resistant to penetration than most synthetic fibers.\newline
Argument B:\newline
F - Elephants have skins that are more resistant to penetration than most synthetic fibers.\newline
C - Hippos have skins that are more resistant to penetration than most synthetic fibers.\newline
}

   & 
\texttt{Both house cats and elephants have skin that is more resistant to penetration than most synthetic fibers. However, elephants are more similar to hippos in terms of size, habitat, and overall biology than house cats are. Therefore, it is more likely that the skin properties of elephants would be a better indicator of the skin properties of hippos.\newline
Based on this reasoning, I would choose option F - Argument B is much stronger.}   &
\texttt{Answer: E - Argument B is moderately stronger.\newline
Reasoning: Argument B provides a more direct comparison to the claim, as it states that elephants have skin that is more resistant to penetration than most synthetic fibers. This is a more direct comparison to the claim that hippos have skin that is more resistant to penetration than most synthetic fibers. Argument A only states that house cats have skin that is more resistant to penetration than most synthetic fibers, which is not as direct of a comparison. Therefore, Argument B makes a moderately stronger case for the claim.}\\  
7) \phen{Property effects} \newline
\citep{heitr94} \newline
\texttt{
Argument A: \newline
F - Chickens have livers with two chambers that act as one.\newline
C - Hawks have livers with two chambers that act as one.\newline
Argument B:\newline
F - Tigers have livers with two chambers that act as one.\newline
C - Hawks have livers with two chambers that act as one.\newline
}
   &
\texttt{Chickens and hawks are both birds, while tigers are mammals. Since the claim is about hawks, it is more likely that the liver structure of another bird (chicken) would be similar to that of a hawk than the liver structure of a mammal (tiger).\newline
Based on this reasoning, I would choose option B - Argument A is moderately stronger.} &
\texttt{Answer: B - Argument A is moderately stronger.\newline
Reasoning: Argument A provides a direct comparison between chickens and hawks, which are both birds, and suggests that hawks likely have livers with two chambers that act as one. Argument B provides a comparison between tigers and hawks, which are not closely related, and does not provide as strong of evidence that hawks have livers with two chambers that act as one.}\\
8) \phen{Property effects} \newline
\citep{heitr94} \newline
\texttt{
Argument A: \newline
F - Tigers usually gather large amounts of food at once.\newline
C - Hawks usually gather large amounts of food at once.\newline
Argument B:\newline
F - Chickens usually gather large amounts of food at once.\newline
C - Hawks usually gather large amounts of food at once.\newline
}

   &
\texttt{Both tigers and chickens are not closely related to hawks, as tigers are mammals and chickens are birds but not birds of prey. However, chickens might be slightly more related to hawks than tigers, as they are both birds. Therefore, the fact that chickens gather large amounts of food at once might be a slightly better indicator of hawks' behavior than the fact that tigers do the same.\newline
Based on this reasoning, my preference is:
D - Argument B is slightly stronger} &
\texttt{Answer: B - Argument A is moderately stronger.\newline
Reasoning: Argument A provides a more relevant fact than Argument B, as tigers and hawks are both predatory animals that hunt for food, while chickens are not. Therefore, Argument A provides a better reason for believing the claim that hawks usually gather large amounts of food at once.}\\
9) \phen{Property effects} \newline
\citep{heitr94} \newline
\texttt{
Argument A: \newline
F - Ladybugs have some cells in their respiratory systems that require carbon dioxide to function.\newline
C - Mosquitoes have some cells in their respiratory systems that require carbon dioxide to function.\newline
Argument B:\newline
F - Vampire bats have some cells in their respiratory systems that require carbon dioxide to function.\newline
C - Mosquitoes have some cells in their respiratory systems that require carbon dioxide to function.\newline
}

&
\texttt{Ladybugs and mosquitoes are both insects, while vampire bats are mammals. Since insects are more closely related to each other than they are to mammals, it is more likely that they share similar respiratory system properties.\newline
Based on this reasoning, I would choose option B - Argument A is moderately stronger.
} & \texttt{Answer: E - Argument B is moderately stronger.\newline
Reasoning: Argument B provides a more specific example of an animal with cells in its respiratory system that require carbon dioxide to function, which makes it a stronger argument for the claim that mosquitoes have some cells in their respiratory systems that require carbon dioxide to function.} \\
   
10) \phen{Property effects} \newline
\citep{heitr94} \newline
\texttt{
Argument A: \newline
F - After eating, vampire bats travel at speeds of twice their body length per second.\newline
C - After eating, mosquitoes travel at speeds of twice their body length per second.\newline
Argument B:\newline
F - After eating, ladybugs travel at speeds of twice their body length per second.\newline
C - After eating, mosquitoes travels at speeds of twice their body length per second.\newline
}
&
\texttt{Vampire bats, ladybugs, and mosquitoes are all different types of animals with distinct characteristics. Vampire bats are mammals, ladybugs are insects, and mosquitoes are also insects. Since ladybugs and mosquitoes are both insects, they share more similarities in terms of their body structure and movement patterns compared to vampire bats. Therefore, the fact about ladybugs might provide a better basis for making a claim about mosquitoes.\newline
Based on this reasoning, I would choose option E - Argument B is moderately stronger.}
& \texttt{Answer: B - Argument A is moderately stronger.\newline
Reasoning: Argument A provides a fact about vampire bats, which are closely related to mosquitoes, and thus provides some evidence that the claim about mosquitoes may be true. Argument B provides a fact about ladybugs, which are not closely related to mosquitoes, and thus provides less evidence that the claim about mosquitoes may be true. Therefore, Argument A provides a moderately stronger case for the claim.} \\
\label{tab:otherpiresponses}\\
\caption{Phenomena from the property induction literature that go beyond those tested in our analyses. All argument pairs are either identical to or modeled on pairs for which average human ratings indicate that argument A is stronger than argument B.} 
\end{longtable}
\doublespacing

\section{Other Inductive Phenomena}
\label{sec:otherphenomena}

Property induction is sometimes considered separately from other inductive problems such as categorization and generalization, but all of these problems are so closely related that it makes little sense to separate them when developing theories of induction~\citep{kempj14}. Table~\ref{tab:otherresponses} includes model responses to an illustrative set of problems from the broader literature on inductive reasoning.

In all cases, the system message supplied to GPT-4 was
\vspace{-2mm}
\begin{displayquote}
{\small
\texttt{You will be asked a series of questions that don't have right or wrong answers. You are willing to use your best judgment and commit to a concrete, specific response even in cases where you can't be sure that you are correct.}}
\end{displayquote}

The user message included one of the exact questions shown in Table~\ref{tab:otherresponses} followed by {\small
\texttt{Please explain your answer carefully.}} As for all other experiments, GPT-3.5 was given a single prompt that included both the system and user messages just described.

The first two rows in the table consider arguments which involve a single category (Animal A) and multiple properties (e.g.\ ``has scales'', ``can fly'' and ``has wings'' in row 1). They therefore differ from the arguments used in our experiments, all of which involved a single property and multiple categories. In the most general case an argument may involve multiple categories and properties, and these arguments have also been studied in the literature~\citep{kempst12}. GPT-4 gives a fairly good response to the first question, but in row 2 it makes the questionable claim that there are ``many reptiles with scales and two legs that do not have wings.''

Rows 3 and 4 consider two examples from the work of \cite{collinsm89} on plausible reasoning. Most people have never encountered these questions previously, but can nevertheless construct a chain of reasoning that arrives at a plausible answer. In contrast, these questions are presumably discussed explicitly in the corpora used when training GPT-3.5 and GPT-4, which makes them less than ideal as a test of reasoning in LLMs. 

Rows 5 through 8 consider examples from the literature on categorization. In some settings it may be useful to distinguish category labels (e.g.\ ``is a wug'') from properties (e.g.\ ``is wuggish'') but the problem of projecting a property from several items to a novel item seems deeply related to the problem of projecting a category label. Ultimately, then, it seems best to consider these problems together.

Row 5 is a case that requires reasoners to go beyond similarity (the 3 inch object is more similar to a pizza than a quarter) to infer that the object is likely to be a pizza. GPT-3.5 fails the test, but GPT-4 suggests that it may be an ``oversized novelty quarter'' and therefore identifies a plausible way in which it could actually be considered a quarter.

Row 6 is inspired by an example in which a person who falls into a pool is classified as ``intoxicated'' even though falling into a pool is not typically associated with the concept of intoxication. Both GPT-4 and GPT-3.5 give good responses by pointing out that alcohol may have been involved.

Rows 7 and 8 consider questions from a classic line of work that explores how inferences about category membership can go beyond surface appearances. In Row 7, GPT-4 gives an excellent response and argues that the creature is still a raccoon even though it looks identical to a skunk. In Row 8, however, GPT-4 suggests that an analogous transformation applied to a coffeepot can indeed transform this object into a bird feeder. A more extensive analysis of examples like these is provided by \cite{zhang2023}.

Although GPT-4 provides good responses to most of the questions in Table~\ref{tab:otherresponses}, these questions are even more likely than those in Tables~\ref{tab:exp1responses} and Tables~\ref{tab:otherpiresponses} to have appeared in its training data in some form. Finding appropriate ways to test the underlying phenomena is therefore a major challenge for future work on inductive reasoning in large language models.

\singlespacing
\begin{longtable}{ p{.30\textwidth}  p{.35\textwidth}  p{.35\textwidth} }
\textbf{Phenomenon} & \textbf{GPT-4} & \textbf{GPT-3.5} \\
1) \phen{Feature inference} \newline
\citep{kemp11} \newline  
\texttt{
Animal A has scales and can fly. How likely is it that Animal A has wings?
}
  &
\texttt{It is likely that Animal A has wings, as the ability to fly is typically associated with the presence of wings. However, the presence of scales is unusual for a flying animal, as most animals with scales (such as reptiles) do not have wings. This suggests that Animal A might be an atypical or fictional creature. Nonetheless, given the information provided, I would estimate that there is a 70\% chance that Animal A has wings, considering that wings are a common adaptation for flight.}
  & 
\texttt{It is very likely that Animal A has wings, as most animals with scales and the ability to fly have wings. However, there are some exceptions, such as some species of fish that can fly, so it is not certain that Animal A has wings.} \\
2) \phen{Feature inference} \newline
\citep{kemp11} \newline
\texttt{
Animal A has scales and two legs. How likely is it that Animal A has wings?
}
  &
\texttt{It is moderately likely that Animal A has wings. The presence of scales suggests that Animal A could be a reptile, and the two legs could indicate that it is a bipedal creature. Some reptiles, like certain species of lizards, have wing-like structures that allow them to glide or fly short distances. However, it is not guaranteed that Animal A has wings, as there are also many reptiles with scales and two legs that do not have wings. Without more information, it is difficult to determine the exact likelihood, but based on the given characteristics, I would estimate that there is a moderate chance that Animal A has wings.} &
\texttt{It is difficult to say how likely it is that Animal A has wings without knowing more information about the animal. If Animal A is a reptile, then it is unlikely that it has wings, as most reptiles do not have wings. However, if Animal A is a bird, then it is more likely that it has wings, as most birds have wings. Ultimately, without more information, it is impossible to say how likely it is that Animal A has wings.} \\
3) \phen{Plausible reasoning} \newline
\citep{collinsm89} \newline
\texttt{
Do you think they might grow rice in Florida?
}
  &
\texttt{Yes, I think they might grow rice in Florida. Florida has a warm, subtropical climate with plenty of rainfall, which is suitable for growing rice. Additionally, the state has large areas of flat land and abundant water resources, which are essential for rice cultivation. However, rice production in Florida might not be as extensive as in other major rice-growing states like California, Arkansas, or Texas. The primary crops in Florida are citrus fruits, sugarcane, and vegetables, but it is still possible that rice is grown on a smaller scale or for local consumption.} &
\texttt{It is possible to grow rice in Florida, but it is not common. Rice requires a lot of water and a long growing season, both of which are not typically found in Florida. Additionally, the soil in Florida is not ideal for growing rice. While it is possible to grow rice in Florida, it is not likely to be successful.} \\
\phen{Plausible reasoning} \newline
4) \citep{collinsm89} \newline
\texttt{
Can a goose quack?
}
  &
\texttt{A goose does not quack like a duck. Geese make a variety of vocalizations, including honking, hissing, and cackling. The primary sound a goose makes is a honk, which is different from a duck's quack. The honk is a louder and more assertive sound, often used by geese to communicate with each other or to warn off potential threats. While both ducks and geese are waterfowl and share some similarities, their vocalizations are distinct from one another.} & 
\texttt{No, a goose cannot quack. Geese make a honking sound, while ducks make a quacking sound.
}\\
5) \phen{Theory-based categorization} \newline
\citep{rips89} \newline
\texttt{
I have a circle-shaped object that is 3 inches in diameter. Do you think it's more likely to be a pizza or a quarter?
}
  &
\texttt{I think it's more likely to be a quarter. A 3-inch diameter is quite small for a pizza, as most personal-sized pizzas are at least 6 inches in diameter. On the other hand, a quarter is approximately 0.955 inches in diameter, which is closer to 3 inches than a typical pizza size. While a 3-inch diameter is still larger than a standard quarter, it's more plausible that it could be an oversized novelty quarter or a similarly sized coin, rather than an unusually small pizza.} & 
\texttt{Based on the size of the object, it is more likely to be a quarter than a pizza. A quarter is approximately 1.2 inches in diameter, so the object is more than twice the size of a quarter. A pizza, on the other hand, is typically much larger than 3 inches in diameter.} \\
6) \phen{Theory-based categorization} \newline
\citep{murphym85} \newline
\texttt{
You are at a party and you learn that one of the guests fell into the pool. Why do you think this happened?
}
  &
\texttt{I think the guest fell into the pool due to a combination of factors, such as being distracted, possibly consuming alcohol, and being too close to the pool's edge. It's common for people to be engaged in conversations or other activities at a party, which can lead to a lack of awareness of their surroundings. Additionally, if the guest had been drinking, their coordination and balance might have been impaired, making it more likely for them to accidentally fall into the pool. Finally, standing or walking too close to the pool's edge increases the risk of slipping or losing balance and falling in.}&
\texttt{It is difficult to say why the guest fell into the pool without knowing more information. It is possible that the guest was not paying attention and accidentally stepped into the pool, or that they were pushed or slipped. It is also possible that the guest was intoxicated and lost their balance.} \\
7) \phen{Theory-based categorization} \newline
\citep{keil89} \newline
\texttt{
The doctors took a raccoon and shaved away some of its fur. They dyed what was left all black. Then they bleached a single stripe all white down the center of its back. Then, with surgery, they put in its body a sac of super smelly odor, just like a skunk has. When they were done, the animal looked just like a skunk. After the operation was this a skunk or a raccoon?
}
  &
\texttt{After the operation, the animal is still a raccoon. Despite the changes made to its appearance and the addition of a smelly odor sac, its genetic makeup and species remain that of a raccoon. The modifications made by the doctors only altered its external appearance and gave it a skunk-like feature, but these changes do not transform the raccoon into a different species. The animal's identity is determined by its genetic makeup and biological characteristics, not just its appearance.} &
\texttt{This is a difficult question to answer definitively because it depends on how you define a skunk and a raccoon. From a physical standpoint, the animal looks like a skunk because it has the same physical features. However, genetically, it is still a raccoon because it has not been altered in any way to change its species. Therefore, it is both a skunk and a raccoon.} \\
8) \phen{Theory-based categorization} \newline
\citep{keil89} \newline
\texttt{
The doctors took a coffeepot. They sawed off the handle, sealed the top, took off the top knob, sealed closed the spout, and sawed it off. They also sawed off the base and attached a flat piece of metal. They attached a little stick, cut a window in it, and filled the metal container with bird food. After the operation was this a coffeepot or a bird feeder? 
}
  &
\texttt{After the operation, this object can be considered a bird feeder. The reason for this classification is that the original function and features of the coffeepot have been significantly altered and repurposed to serve as a container for bird food. The handle, top knob, and spout have been removed and sealed, and a window has been cut to allow birds access to the food. The addition of a flat piece of metal and a stick further supports its new function as a bird feeder. While it may have originated as a coffeepot, its current form and purpose align more closely with that of a bird feeder.}&\texttt{No, this is not a coffeepot anymore. It has been modified so much that it is now a bird feeder. The handle has been sawed off, the top knob has been removed, the spout has been sealed closed, the base has been sawed off, and a flat piece of metal has been attached. A stick has been attached and a window has been cut in it, and it has been filled with bird food. All of these modifications have changed the object from a coffeepot to a bird feeder.} \\  
\label{tab:otherresponses}\\
\caption{Phenomena from the broader literature on inductive reasoning.} 
\end{longtable}
\doublespacing

\end{document}